\documentclass[11pt]{article}

\usepackage[preprint]{acl}

\usepackage{times}
\usepackage{latexsym}

\usepackage[T1]{fontenc}

\usepackage[utf8]{inputenc}

\usepackage{microtype}

\usepackage{inconsolata}

\usepackage{graphicx}
\usepackage{arydshln}
\usepackage{colortbl}

\usepackage{kotex}

\usepackage{makecell}
\usepackage{amssymb}
\usepackage{tcolorbox}
\usepackage{verbatim}
\tcbuselibrary{breakable}
\usepackage{listings}
\usepackage{multirow}
\usepackage{tabularx}
\usepackage{amsmath}
\usepackage{booktabs}
\usepackage{xspace}
\usepackage{subcaption}
\usepackage{fvextra}

\newcommand{\mymethod}{\textsc{VFStance}\xspace}
\newcommand{\mymethodfull}{\textbf{V}isual \textbf{F}raming for \textbf{Stance} Detection\xspace}

\newcolumntype{C}{>{\centering\arraybackslash}X}

\usepackage{xcolor}
\definecolor{promptblue}{RGB}{0,92,175}

\usepackage{bm}

\usepackage{svg}
\usepackage{float}

\title{Visual Framing for News Stance Detection via Image Generation}

\author{Dahyun Lee \\
  Soongsil University\\
  \small{\texttt{hyundai@soongsil.ac.kr}} \\\And
  Jiyoung Han \\
  KAIST\\
  \small{\texttt{jiyoung.han@kaist.ac.kr}} \\\And
  Kunwoo Park\\
  Soongsil University\\
  \small{\texttt{kunwoo.park@ssu.ac.kr}} \\}

\begin{document}
\maketitle
\begin{abstract}
Article-level news stance detection aims to identify the perspective of news articles toward social issues. Despite advances in stance detection and its importance for trustworthy media environments, news articles pose distinct challenges because their stances are often implicit, subtly conveyed through journalistic framing, and embedded in long, structurally complex texts. To address these challenges, we introduce \mymethod, which leverages visual framing to make implicit stance cues more explicit via image generation. In evaluation experiments, we demonstrate the effectiveness of \mymethod over existing methods and the contribution of visual framing to its performance. Finally, a controlled user study ($N=200$) in a snippet-based news consumption setting further demonstrates that \mymethod can make stance signals visually salient and highlights its potential use beyond automated stance detection.
\end{abstract}

\section{Introduction}
News articles can present different perspectives on the same issue through selection and emphasis of particular interpretive frames~\cite{gentzkow-etal-2010-drives}. Automatically identifying these perspectives is important for analyzing media bias~\cite{hamborg-etal-2019-automated}, understanding public opinion~\cite{chong-etal-2007-framing}, and supporting informed news consumption~\cite{park-etal-2009-newscube}. Stance detection, the task of identifying an author's expressed attitude toward a specific target from text, has been extensively studied in social media domains~\cite{mohammad-etal-2016-semeval}. More recently, the task has been extended to news articles, where the goal is to classify whether a full article's position toward a target issue is supportive, neutral, or oppositional~\cite{mascarell-etal-2021-stance}.

\begin{figure}[t]
\centering
  \includegraphics[width=\columnwidth]{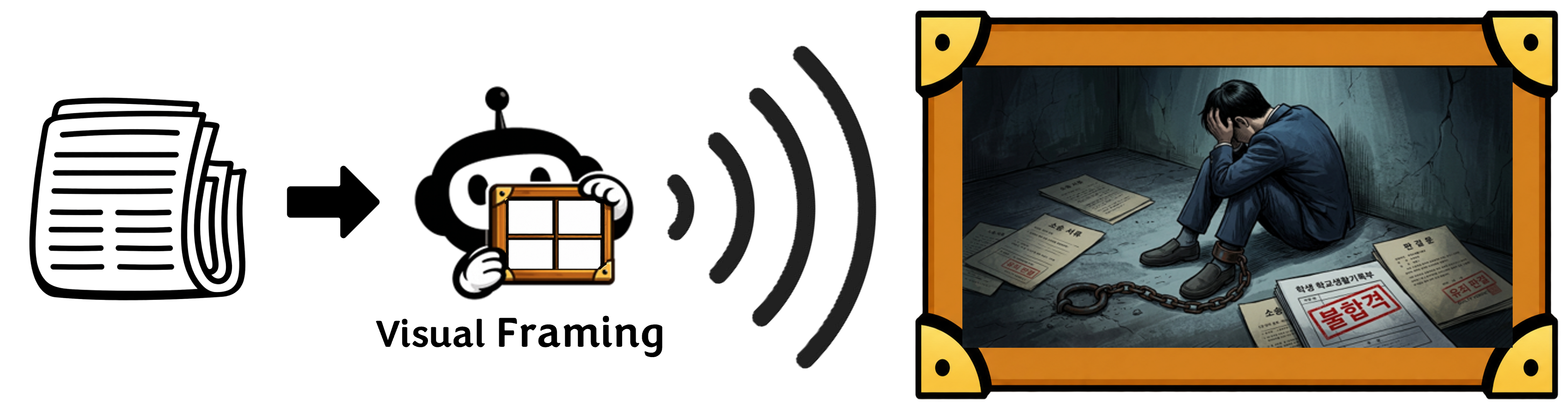}
  \caption{Key idea behind \mymethod: making implicit stance cues in a lengthy, rhetorically complex news article explicit through image generation grounded in visual framing.}  
  \label{fig:figure1}
\end{figure}

Despite advances in NLP and large language models (LLMs), article-level news stance detection remains challenging for two key reasons. First, professional journalistic norms generally favor detached and fact-oriented reporting~\cite{kovach-etal-2021-elements}; thus, a news article's stance toward a target is often implicit and subtly expressed. Second, news articles are long and structurally complex. Stance detection methods developed primarily for short texts therefore do not readily transfer to article-level news stance detection. Even human readers may find it difficult to infer stance from subtle framing cues distributed across lengthy, rhetorically elaborate news articles. 

These challenges motivate the use of visual information, which can convey attitudinal cues more immediately and intuitively than text~\cite{mehrabian-etal-1981-silent, paivio-etal-1986-mental}. Visual representations may therefore provide a complementary means of making  implicit and dispersed stance cues in news articles more accessible. Publisher-provided news images (i.e., the original images published alongside the articles), however, are not always available and, even when present, may not closely align with an article's stance, as press photographs are typically produced under documentary norms that may constrain overt evaluative signaling~\cite{schwartz-etal-1992-photojournalism}. This motivates our first research question: \emph{Can image generation technology enable more effective news stance detection?}

Recent advances in text-to-image (T2I) generation and its growing adoption across diverse domains suggest a new possibility: generated images may serve as intermediate visual representations that make stance cues already present in news text more explicit to downstream models~\cite{zhang-etal-2025-exploring-artificial}. Generating such representations, however, remains challenging because the cues most relevant to article-level news stance detection are themselves subtle, implicit, and distributed throughout the text. Na\"ively generating an image from a news article may therefore fail to capture the framing choices most indicative of its stance. This leads to our second research question: \emph{How can we generate images that make stance cues more explicit?}

\begin{figure*}[t]
  \centering
  \resizebox{\textwidth}{!}{%
    \includegraphics[width=\columnwidth]{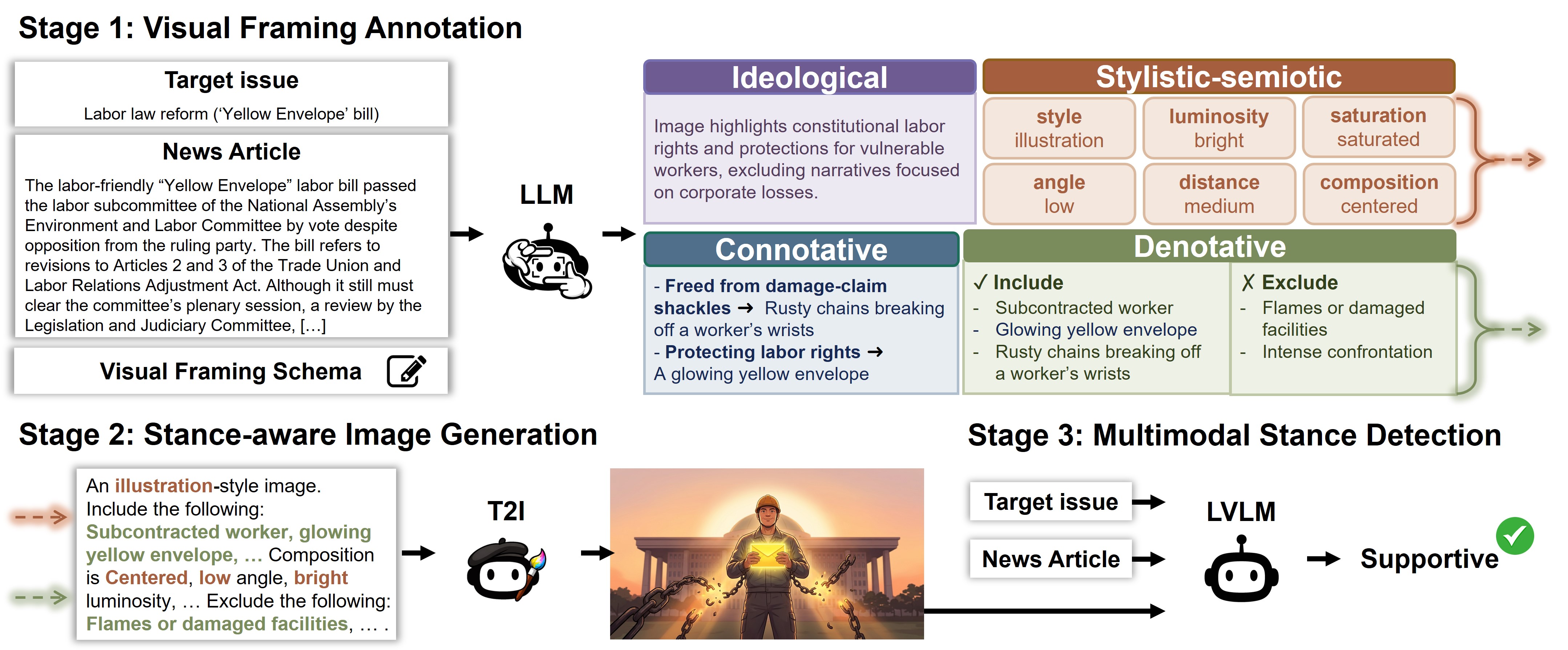}%
  }
  \caption{Overview of \mymethod: a multi-stage, modular framework grounded in visual framing.}
  \label{fig:figure2}
\end{figure*}

To address these questions, we draw on framing theory from communication and media studies, particularly research on visual framing in news~\cite{messaris-etal-2001-role, rodriguez-etal-2011-levels}. We hypothesize that image generation grounded in visual framing can transform implicit stance cues in a news article into visually salient stance signals, thereby enabling more effective stance detection. \mymethod (\mymethodfull) is a multi-stage, modular framework designed to test this hypothesis (Figure~\ref{fig:figure1}). As illustrated in Figure~\ref{fig:figure2}, \mymethod first uses an LLM to derive visual framing specifications from the article, which are then used to prompt a T2I model to generate an image that makes the corresponding stance cues more explicit. Finally, an instruction-following large vision-language model (LVLM) predicts the article-level stance jointly from the article and the generated image. 

We evaluate \mymethod on two article-level stance detection datasets in Korean and German and compare its performance with existing methods. The results show that \mymethod outperforms existing approaches across both datasets. Ablation experiments further demonstrate the contributions of visual framing and image generation to its performance. Finally, we conduct a controlled user study ($N=200$) in a snippet-based news consumption setting to examine whether the generated images also help human readers identify article stance. Participants exposed to images generated by \mymethod identify article stance more accurately than those exposed to alternative image conditions. Taken together, these findings suggest that visual framing and image generation can make otherwise implicit stance cues more readily discernible to both computational models and human readers.

The contributions of this study are summarized as follows:
\begin{itemize}
    \item We propose \mymethod, a multi-stage framework grounded in visual framing that transforms implicit stance cues in news articles into more explicit visual representations for article-level stance detection.
    \item We evaluate \mymethod on two article-level stance detection datasets and demonstrate its effectiveness across languages, while ablation experiments establish the contributions of both visual framing and image generation.
    \item We conduct a controlled user study ($N=200$) in a snippet-based news consumption setting, showing that images generated by \mymethod help readers identify article stance more accurately and suggesting potential applications beyond automated stance detection.
\end{itemize}

\section{Related Work}
\paragraph{Article-Level News Stance Detection}
Much of the stance detection literature has examined short-form social media text, particularly tweets. Recent approaches have included fine-tuning pre-trained language models~\cite{liang-etal-2022-zero}, leveraging LLM reasoning~\cite{gatto-etal-2023-chain, zhang-etal-2024-llm} and employing in-context learning~\cite{zhu-etal-2023-can, cruickshank-ng-2023-prompting}. Multimodal stance detection methods that jointly utilize text and images have also emerged, including target-aware multimodal prompt tuning~\cite{liang-etal-2024-multi}, cross-modal fusion~\cite{weinzierl-harabagiu-2023-identification}, multimodal alignment~\cite{zhang-etal-2025-mad}, and image generation~\cite{zhang-etal-2025-exploring-artificial}. Recent studies have further explored zero-shot detection with LVLMs~\cite{weinzierl-harabagiu-2024-tree, alshenaifi-etal-2026-beyond}. 

News stance detection remains relatively underexplored. Existing work has largely centered on headline- or sentence-level stance in contexts such as fake news detection or rumor verification~\cite{ferreira-2016-emergent, pomerleau-rao-2017-fnc, hanselowski-etal-2018-retrospective, conforti-etal-2020-stander}, while studies of article-level stance toward social issues remain scarce~\cite{mascarell-etal-2021-stance,lee-etal-2025-journalism}. Building on recent advances in multimodal stance detection~\cite{liang-etal-2024-multi, zhang-etal-2025-exploring-artificial}, we address this gap by generating visual representations that make implicit stance signals in news articles more explicit and leveraging them for article-level stance detection.

\paragraph{Visual Framing Analysis}
Framing involves selecting and emphasizing certain aspects of reality to promote particular interpretations~\cite{entman-etal-1993-framing}, and visual framing concerns how such interpretive emphasis is conveyed through visual elements~\cite{messaris-etal-2001-role}. Empirical studies in communication have operationalized frames through manual content analysis using predefined categories~\cite{coleman-etal-2010-framing}. In the computational domain, framing analysis has emerged as an active research area, as documented by recent surveys~\cite{ali-etal-2022-survey, otmakhova-etal-2024-media, vallejo-etal-2024-connecting}. Earlier computational approaches primarily analyzed the textual modality using annotated corpora and supervised methods. Widely used resources include the Media Frames Corpus with fifteen generic policy frames~\cite{card-etal-2015-media}, the Gun Violence Frame Corpus~\cite{liu-etal-2019-detecting}, and the multilingual SemEval-2023 benchmark covering nine languages~\cite{piskorski-etal-2023-semeval}. More recent research has explored multimodal approaches, such as fusion-based approaches using paired headlines and lead images~\cite{tourni-etal-2021-detecting}. Studies have also leveraged LLMs and LVLMs for frame analysis of news text and imagery~\cite{arora-etal-2025-multimodal,lu-etal-2026-evaluating, eldamanhoury-etal-2026-visual}. Our study extends this line of research by using visual framing to guide image generation for article-level stance detection.

\section{Problem and Dataset}

\subsection{Target Problem}
For a news article $A$ that covers an issue $T$, the goal of article-level stance detection is to determine the stance of $A$ toward $T$ as \emph{supportive}, \emph{neutral}, or \emph{oppositional} using a detection model $M(\cdot)$. We investigate whether specifying visual framing elements and generating images based on these specifications can lead to more effective stance detection.

\subsection{Datasets}
We use two article-level news stance detection datasets: one accompanied by news images and one without images. The first is \textsf{K-News-Stance-MM}, a novel multimodal extension of the existing article-level stance detection dataset~\cite{lee-etal-2025-journalism}, comprising 1,816 Korean news articles with accompanying news images. As the first dataset to provide article-level stance labels together with news images, it serves as the primary testbed for this study. The second is \textsf{CheeSE}~\cite{mascarell-etal-2021-stance}, which contains 1,762 German news articles. We use this dataset to examine the applicability of our method for predicting article-level stance in a text-only setting. Dataset sizes and label distributions are provided in Table~\ref{tab:dataset_statistics}, and detailed dataset information is provided in Appendix~\ref{app:sec:dataset}.

For broader accessibility and supplementary analysis, we also provide LLM-translated multilingual extensions of \textsf{K-News-Stance-MM} in English, Chinese, Indonesian, and Arabic. The target languages were selected to span different resource levels~\cite{joshi-etal-2020-state}.

\section{Proposed Method}
\label{sec:method}

We introduce \mymethod (\mymethodfull), a multi-stage, modular framework for article-level stance detection. The central idea is to transform implicit textual stance cues into more explicit framing signals that can be rendered as visually salient elements by a T2I model, thereby enabling an instruction-following LVLM to detect article-level stance through multimodal reasoning.

News articles often express their stance toward a target issue implicitly, consistent with professional journalistic norms favoring detached and fact-oriented reporting~\cite{kovach-etal-2021-elements}. Rather than expressing stance through explicit evaluative claims, articles may convey it through framing choices, such as selective emphasis on particular actors, causes, consequences, or interpretations of an issue~\cite{entman-etal-1993-framing}. These cues are often subtle and distributed across long documents, making them difficult to detect from text alone.
 
Incorporating publisher images is a promising direction for improving stance detection performance (Appendix~\ref{app:sec:analysis}). However, relying on conventional news images presents two limitations. First, not all news articles include an image. Second, even when images are available, they may provide limited stance-relevant information because publisher news images often serve documentary and contextual functions rather than explicitly foregrounding an article's interpretive stance~\cite{reuters-2008-handbook, associatedpress-2024-news-values}. Image generation using a T2I model offers a possible solution, but na\"ively prompting the model with a full article or its summary may fail to capture stance-relevant framing cues effectively. The model must first identify which subtle and dispersed framing cues are most indicative of the article's stance before rendering them visually (see Table~\ref{tab:app:image_comparison} for examples of na\"ive generation using a straightforward prompt).

To address these challenges, \mymethod draws on visual framing~\cite{rodriguez-etal-2011-levels} to make subtle stance cues more explicit for article-level stance detection. As illustrated in Figure~\ref{fig:figure2}, \mymethod is a multi-stage, modular framework built on an LLM, a T2I model, and an LVLM. The framework first uses an LLM to derive a structured visual framing specification (\textbf{Stage 1}). A T2I model then generates a news image based on this specification (\textbf{Stage 2}), making the underlying stance cues more visually evident. Finally, an LVLM uses the article and the generated image to predict article-level stance (\textbf{Stage 3}). Stage 2 can also be skipped, in which case the visual framing specification is provided directly to Stage 3 in textual form; we refer to this variant as \textsc{\mymethod (Text)}.

\begin{table*}[t]
\centering
\resizebox{\textwidth}{!}{%
\begin{tabular}{l l c c c c c}
\toprule
\textbf{Category} & \textbf{Method} 
& \textbf{ACC} & \textbf{mF1} 
& \textbf{F1$_{\text{Supportive}}$} 
& \textbf{F1$_{\text{Neutral}}$} 
& \textbf{F1$_{\text{Oppositional}}$} \\

\midrule
\multirow{3}{*}{\mymethod} 
& Gemini-3-flash & \textbf{0.746 $\pm$ 0.002} & \textbf{0.747 $\pm$ 0.002} & \textbf{0.78 $\pm$ 0.003} & \textbf{0.649 $\pm$ 0.002} & \textbf{0.813 $\pm$ 0.003} \\
& Claude-4.6-sonnet & 0.694 $\pm$ 0.002 & 0.696 $\pm$ 0.002 & 0.732 $\pm$ 0.002 & 0.556 $\pm$ 0.002 & 0.8 $\pm$ 0.001 \\
& GPT-5.4-mini      & 0.671 $\pm$ 0.003 & 0.675 $\pm$ 0.003 & 0.69 $\pm$ 0.004 & 0.562 $\pm$ 0.003 & 0.774  $\pm$ 0.004 \\

\midrule
\multirow{7}{*}{Multimodal}
& Gemini-3-flash     & \textbf{0.719 $\pm$ 0.001} & \textbf{0.726 $\pm$ 0.001} & \textbf{0.73 $\pm$ 0.002} & \textbf{0.659 $\pm$ 0.001} & \textbf{0.788 $\pm$ 0.001} \\ 
& Claude-4.6-sonnet  & 0.669 $\pm$ 0.002 & 0.674 $\pm$ 0.002 & 0.691 $\pm$ 0.005 & 0.555 $\pm$ 0.002 & 0.778 $\pm$ 0.002 \\ 
& GPT-5.4-mini       & 0.658 $\pm$ 0.002 & 0.664 $\pm$ 0.002 & 0.652 $\pm$ 0.001 & 0.569 $\pm$ 0.004 & 0.769 $\pm$ 0.002 \\ 
& RoBERTa+ViT        & 0.619 $\pm$ 0.033 & 0.62  $\pm$ 0.036 & 0.633 $\pm$ 0.022 & 0.597 $\pm$ 0.015 & 0.63  $\pm$ 0.075 \\
& CLIP               & 0.364 $\pm$ 0.021 & 0.345 $\pm$ 0.023 & 0.25  $\pm$ 0.071 & 0.364 $\pm$ 0.039 & 0.421 $\pm$ 0.032 \\
& TMPT               & 0.347 $\pm$ 0.019 & 0.335 $\pm$ 0.013 & 0.272 $\pm$ 0.054 & 0.366 $\pm$ 0.045 & 0.368 $\pm$ 0.047 \\
& T-MAD              & 0.332 $\pm$ 0.017 & 0.306 $\pm$ 0.023 & 0.306 $\pm$ 0.033 & 0.403 $\pm$ 0.044 & 0.208 $\pm$ 0.08 \\
\midrule
\multirow{7}{*}{Textual}
& Gemini-3-flash    & \textbf{0.712 $\pm$ 0.002} & \textbf{0.719 $\pm$ 0.002} & \textbf{0.711 $\pm$ 0.004} & \textbf{0.663 $\pm$ 0.002} & \textbf{0.781 $\pm$ 0.004} \\
& GPT-5.4-mini      & 0.653 $\pm$ 0.003 & 0.66  $\pm$ 0.003 & 0.653 $\pm$ 0.005 & 0.563 $\pm$ 0.004 & 0.765 $\pm$ 0.005 \\
& Claude-4.6-sonnet & 0.636 $\pm$ 0.002 & 0.629 $\pm$ 0.002 & 0.659 $\pm$ 0.002 & 0.451 $\pm$ 0.003 & 0.776 $\pm$ 0.003 \\
& PT-HCL            & 0.621 $\pm$ 0.007 & 0.621 $\pm$ 0.005 & 0.638 $\pm$ 0.02  & 0.604 $\pm$ 0.021 & 0.62  $\pm$ 0.015 \\
& LKI-BART          & 0.618 $\pm$ 0.021 & 0.614 $\pm$ 0.027 & 0.596 $\pm$ 0.059 & 0.578 $\pm$ 0.065 & 0.669 $\pm$ 0.025 \\
& RoBERTa           & 0.604 $\pm$ 0.038 & 0.602 $\pm$ 0.04  & 0.61 $\pm$ 0.063  & 0.604 $\pm$ 0.024 & 0.594 $\pm$ 0.038 \\
& CoT Embeddings    & 0.583 $\pm$ 0.07  & 0.569 $\pm$ 0.088 & 0.608 $\pm$ 0.069 & 0.504 $\pm$ 0.196 & 0.593 $\pm$ 0.132 \\

\midrule
\multirow{6}{*}{Visual}
& Gemini-3-flash     & \textbf{0.46 $\pm$ 0.004} & \textbf{0.456 $\pm$ 0.005} & 0.418 $\pm$ 0.006 & \textbf{0.48 $\pm$ 0.005} & \textbf{0.471 $\pm$ 0.005} \\
& Claude-4.6-sonnet  & 0.435 $\pm$ 0.002 & 0.433 $\pm$ 0.002 & \textbf{0.425 $\pm$ 0.004} & 0.453 $\pm$ 0.003 & 0.42 $\pm$ 0.004 \\
& GPT-5.4-mini       & 0.38 $\pm$ 0.002 & 0.35 $\pm$ 0.002 & 0.346 $\pm$ 0.004 & 0.457 $\pm$ 0.002 & 0.246 $\pm$ 0.003 \\

& ResNet             & 0.345 $\pm$ 0.007 & 0.32 $ \pm$ 0.018 & 0.218 $\pm$ 0.064 & 0.316 $\pm$ 0.042 & 0.427 $\pm$ 0.018 \\
& SwinT              & 0.337 $\pm$ 0.018 & 0.31  $\pm$ 0.026 & 0.206 $\pm$ 0.05 & 0.39   $\pm$ 0.03  & 0.334 $\pm$ 0.104 \\
& ViT                & 0.317 $\pm$ 0.014 & 0.305 $\pm$ 0.018 & 0.259 $\pm$ 0.07 & 0.363  $\pm$ 0.037 & 0.292 $\pm$ 0.025 \\
\bottomrule
\end{tabular}%
}
\caption{Article-level stance detection performance on the test split of \textsf{K-News-Stance-MM}, measured by accuracy (ACC) and macro F1 (mF1). Categories and models are sorted by accuracy, with the best-performing model in each category highlighted in bold. Models in the \mymethod category differ only in the LVLM used as the Stage 3 detector. Methods in the multimodal and visual categories use publisher-provided news images.}
\label{tab:main_results}
\end{table*}

\paragraph{Stage 1: Visual Framing Annotation}

We first employ an LLM to specify visual framing features. The specification captures both the image content---that is, which actors, objects, or scenes should be included or excluded---and visual presentation, including style, composition, angle, distance, saturation, and luminosity. To construct this specification, we adapt Rodriguez and Dimitrova's (\citeyear{rodriguez-etal-2011-levels}) model of visual framing as an annotation schema for image generation. Because the original model does not define variables specifically for image generation, we use prior visual framing literature~\cite{kress-etal-1996-reading, hall-etal-1966-hidden, barthes-etal-1977-rhetoric, messaris-etal-2001-role} to define annotatable features suitable for text-to-image generation. The theoretical grounding and operationalization of these features are described in Appendix~\ref{app:sec:method:mymethod}.

The resulting schema consists of four levels, as illustrated in Stage 1 of Figure~\ref{fig:figure2}. At the \emph{ideological} level, the LLM identifies whose perspective or interests the image should serve. At the \emph{connotative} level, it specifies the interpretive associations that the image should evoke beyond its literal depiction. For instance, when an article frames inter-ministerial cooperation as effective coordination, the image may include symbolic elements such as interlocking gears to convey integration and collective action. At the \emph{stylistic-semiotic} level, the LLM assigns values to six features: style, composition, angle, distance, saturation, and luminosity. Together, these features shape the image's visual presentation and tone. At the \emph{denotative} level, the model determines which subjects, objects, or scenes should be included or excluded.

Given a news article $A$ and its target issue $T$ as input, an LLM annotates the ten features across the four levels and returns the visual framing specification in JSON format. Stage 1 derives this specification solely from these inputs, without predicting the article's stance or accessing the gold label. Figure~\ref{fig:visual_framing_system_prompt} shows the full prompt, and Figure~\ref{fig:visual_framing_output_ko} provides an example output.

\paragraph{Stage 2: Stance-Aware Image Generation}
Given the visual framing specification from Stage 1, we generate a news image using a T2I model. Among the ten features across four levels, we use eight features from the stylistic-semiotic and denotative levels. These levels provide explicit, visually renderable specifications, whereas the ideological and connotative levels capture more abstract interpretive information. Image $I$ is generated using the template-based prompt shown in Figure~\ref{fig:text-to-image_example_en}.

\paragraph{Stage 3: Multimodal Stance Detection}
The final step is to predict the stance of article $A$ toward a target issue $T$. The detector model $M$ receives both the article text and the stance-aware image $I$ generated in Stage 2. We use an LVLM as $M$, with the prompt shown in Figure~\ref{fig:multimodal_prompt}. \mymethod aims to improve stance detection by combining the article text, which contains implicit stance cues, with a generated image designed to externalize stance-relevant framing signals. For \textsc{\mymethod (Text)}, the visual framing specification from Stage 1 is provided to the detector instead of $I$. 

\paragraph{Model Configurations}
We use Gemini-3-flash as the LLM in Stage 1 and Gemini-3.1-flash-image (also known as Nano Banana 2) as the T2I model in Stage 2. The latter was selected for its multilingual support and image generation quality. For Stage 3, we employ Gemini-3-flash as the LVLM because it achieved the best stance detection performance among the three proprietary models evaluated in this study---Gemini-3-flash, GPT-5.4-mini, and Claude-4.6-sonnet. A comparison with open-model alternatives is provided in Appendix~\ref{app:sec:openweight}.

\section{Evaluation Results}
\label{sec:evaluation}
We present evaluation results on article-level stance detection, measured by accuracy (ACC) and macro F1 (mF1). We report the average performance over five runs, along with standard errors. The Mann--Whitney U test was used to assess the statistical significance of the differences. Detailed experimental settings are provided in Appendix~\ref{app:sec:experimental_setups}. 

\paragraph{Comparison with Existing Methods} Table~\ref{tab:main_results} presents the article-level stance detection results on the \textsf{K-News-Stance-MM} test split, comparing \mymethod with the baseline methods. We evaluated eleven fine-tuned baselines grouped into textual, visual, and multimodal methods, all of which were proposed in previous studies. These models were fine-tuned on samples from the training split. Further details are provided in Appendix \ref{app:sec:method}.

\emph{Textual} baselines include \textbf{RoBERTa}~\cite{liu-etal-2019-roberta}, a fine-tuned classifier based on a masked language model; \textbf{CoT Embeddings}~\cite{gatto-etal-2023-chain}, which trains RoBERTa on chain-of-thought reasoning traces generated by an LLM; \textbf{LKI-BART}~\cite{zhang-etal-2024-llm}, which injects text--target relational knowledge extracted by an LLM into a BART model; and \textbf{PT-HCL}~\cite{liang-etal-2022-zero}, which separates target-invariant and target-specific features via contrastive learning. 

\emph{Visual} baselines include \textbf{ResNet}~\cite{he-etal-2016-deep}, a convolutional neural network; \textbf{ViT}~\cite{dosovitskiy-etal-2021-image}, a vision transformer; and \textbf{SwinT}~\cite{liu-etal-2021-swin}, a hierarchical vision transformer with shifted window self-attention. 

\emph{Multimodal} baselines include \textbf{RoBERTa+ViT}, which concatenates textual and visual [CLS] representations; \textbf{CLIP}~\cite{radford-etal-2021-learning}, which concatenates text and image embeddings from pretrained CLIP encoders; \textbf{TMPT}~\cite{liang-etal-2024-multi}, a target-aware multimodal prompt-tuning method; and \textbf{T-MAD}~\cite{zhang-etal-2025-mad}, a target-driven multimodal alignment method. Additionally, we evaluated the three proprietary LVLMs used as Stage 3 backbones in the proposed method using a straightforward prompting strategy. 

Three key observations emerge from Table~\ref{tab:main_results}. First, the three LVLMs exhibited strong performance across the three baseline categories, outperforming all fine-tuned methods (\textit{p}$<$0.01). Among them, Gemini-3-flash consistently achieved the best performance across categories (\textit{p}$<$0.01), supporting its use as the backbone for \mymethod. Second, \mymethod with Gemini-3-flash as the backbone achieved the best overall performance, with an accuracy of 0.746 and a macro F1 score of 0.747. The proposed method outperformed all baselines (\textit{p}$<$0.01), including both LVLM-based and fine-tuned methods. This result provides empirical support for the effectiveness of the proposed framework for article-level stance detection. Third, in terms of class-wise prediction performance, \mymethod with Gemini-3-flash achieved higher F1 scores for the supportive and oppositional labels than the multimodal baseline, increasing F1$_{\text{Supportive}}$ from 0.73 to 0.78 (\textit{p}$<$0.01) and F1$_{\text{Oppositional}}$ from 0.788 to 0.813 (\textit{p}$<$0.01). By contrast, F1$_{\text{Neutral}}$ slightly decreased from 0.659 to 0.649 (\textit{p}$<$0.01). These findings suggest that the proposed method is particularly effective at making directional stance cues more explicit.

\begin{table}[t]
\centering
\small
\resizebox{\columnwidth}{!}{%
\begin{tabular}{l c c}
\toprule
\textbf{Method} & \textbf{ACC} & \textbf{mF1} \\
\midrule
\mymethod              & \textbf{0.746 $\pm$ 0.002} & \textbf{0.747 $\pm$ 0.002} \\
Direct T2I  & 0.721  $\pm$ 0.001 &    0.723  $\pm$ 0.001  \\
EAIG4SD                & 0.72  $\pm$ 0.002 &     0.727 $\pm$ 0.002  \\
Meta-Prompting  & 0.712  $\pm$ 0.004 &    0.709  $\pm$ 0.004  \\
\bottomrule
\end{tabular}%
}
\caption{Ablation results on the impact of visual framing in image generation on stance detection performance.}
\label{tab:image_gen_comparison}
\end{table}

\begin{table}[t]
\centering
\resizebox{\columnwidth}{!}{%
\begin{tabular}{lcc}
\toprule
\textbf{Method} & \textbf{ACC} & \textbf{mF1} \\
\midrule
\mymethod        & \textbf{0.746 $\pm$ 0.002} & \textbf{0.747 $\pm$ 0.002} \\
\mymethod \textsc{(Text)}   & 0.73 $\pm$ 0.003 &     0.725 $\pm$ 0.003 \\
1-Step Prompting    & 0.688 $\pm$ 0.007 &     0.677 $\pm$ 0.008 \\
\bottomrule
\end{tabular}%
}
\caption{Ablation results on the impact of image generation on stance detection performance.}
\label{tab:modality_ablation}
\end{table}

\paragraph{Ablation: Visual Framing in Image Generation}
To investigate the contribution of visual framing to image generation, we compared \mymethod with variants that use alternative image generation methods in Stage 2; the results are presented in Table~\ref{tab:image_gen_comparison}. \emph{Meta-Prompting} instructs an LLM to generate a prompt for a T2I model by specifying the task goal without using a visual framing schema. \emph{Direct T2I} passes the news text directly to the T2I model to generate stance-aware images for prediction. \emph{EAIG4SD} is an image generation framework proposed for stance detection on tweets~\cite{zhang-etal-2025-exploring-artificial} and, to our knowledge, is the only prior method that uses image generation for stance detection. All three methods achieved lower accuracy and macro F1 scores than \mymethod, with gaps of at least 0.025 in accuracy and 0.02 in macro F1 (\textit{p}$<$0.01). These results provide empirical support for the role of visual framing to image generation and, consequently, to the performance gains achieved by \mymethod.

\paragraph{Ablation: Use of Image Generation}
Given the effectiveness of visual framing in image generation identified above, we next investigated whether image generation itself is necessary. We compare \mymethod with two alternative methods that do not use image generation but remain grounded in visual framing; their performance is reported in Table~\ref{tab:modality_ablation}. \textsc{\mymethod~(Text)} uses the visual framing specification from Stage 1 as input context for the LVLM in Stage 3 while skipping image generation in Stage 2. \emph{1-Step Prompting} instructs an LVLM, Gemini-3-flash, to perform visual framing annotation and stance prediction in a single step. 

The results show that \mymethod achieves statistically significant improvements over both alternatives without image generation (\textit{p}$<$0.01), indicating that image generation provides an additional performance gain beyond the contribution of visual framing identified in Table~\ref{tab:image_gen_comparison}. Another notable finding is the strong performance of \textsc{\mymethod~(Text)}, which outperforms all baseline methods in Table~\ref{tab:main_results}, as well as \emph{1-Step Prompting}. These results provide empirical evidence that visual framing contributes substantially even when represented only as a textual specification rather than rendered as an image. Considering the cost--accuracy trade-off between the two variants (Table~\ref{tab:app:comparison_tradeoff_perf_inf-cost}), \mymethod and \textsc{\mymethod~(Text)} may serve different practical needs: users may prefer \textsc{\mymethod~(Text)} when computational cost is the primary concern and thus omit image generation, whereas \mymethod is preferable when maximizing detection accuracy is the priority.

\begin{table}[t]
\centering
\resizebox{\columnwidth}{!}{%
\begin{tabular}{cccccc}
\toprule
\textbf{D} & \textbf{S} & \textbf{C} & \textbf{I} & \textbf{ACC} & \textbf{mF1} \\
\midrule
\checkmark & \checkmark &            &            & \textbf{0.746 $\pm$ 0.002} & \textbf{0.747 $\pm$ 0.002} \\
\midrule
\checkmark &            &            &            & 0.732 $\pm$ 0.002 & 0.736 $\pm$ 0.002 \\

\checkmark & \checkmark & \checkmark &            & 0.725 $\pm$ 0.002 & 0.728 $\pm$ 0.002 \\
\checkmark & \checkmark & \checkmark & \checkmark & 0.722  $\pm$ 0.003 & 0.723 $\pm$ 0.003 \\
\bottomrule
\end{tabular}%
}
\caption{Stance detection performance according to the visual framing levels adopted in Stage 2 (D: denotative, S: stylistic-semiotic, C: connotative, and I: ideological).}
\label{tab:level_ablation}
\end{table}

\begin{table}[t]
\centering
\resizebox{\columnwidth}{!}{
\begin{tabular}{cccccc}
\toprule
\textbf{D} & \textbf{S} & \textbf{C} & \textbf{I} & \textbf{ACC} & \textbf{mF1} \\
\midrule
\checkmark & \checkmark & \checkmark & \checkmark & \textbf{0.746 $\pm$ 0.002} & \textbf{0.747 $\pm$ 0.002} \\
\checkmark & \checkmark &            &            & 0.739 $\pm$ 0.003 & 0.737 $\pm$ 0.003 \\
\bottomrule
\end{tabular}
}
\caption{Stance detection performance according to the visual framing levels used in Stage 1 (D: denotative, S: stylistic-semiotic, C: connotative, and I: ideological).}
\label{tab:annotated_levels_ablation}
\end{table}

\paragraph{Ablation: Visual Framing Level Selection}
We present ablation studies to support our choice of the denotative and stylistic-semiotic levels in Stage~2, selected from the four levels produced by the LLM-based annotation in Stage~1. In each comparison, only the features from the selected levels are used in the image-generation prompt for Stage~2, while all four levels are annotated identically in Stage~1. Table~\ref{tab:level_ablation} presents the level-wise ablation results, showing that the two levels adopted in \mymethod are critical for stance detection performance, whereas the connotative and ideological levels are ineffective and even reduce accuracy and macro F1 when included. This finding suggests that the abstract nature of the two excluded levels makes them difficult to render visually in a way that makes stance signals more explicit, consistent with prior findings on the difficulty of generating abstract concepts~\cite{liao-etal-2024-tiac}.

Given the effectiveness of these two levels, we further examined whether annotating the visual framing specification across all four levels in Stage 1 is necessary. Specifically, we measured the performance of \mymethod when only the stylistic-semiotic and denotative levels were included in the Stage 1 annotation schema. Table~\ref{tab:annotated_levels_ablation} shows that additionally annotating the connotative and ideological features improves the effectiveness of the resulting annotations for the eight denotative and stylistic-semiotic features, yielding a 0.01 increase in macro F1. According to the hierarchical structure of \citet{rodriguez-etal-2011-levels}'s model, the connotative and ideological levels provide interpretive context for lower visual elements, which may partially explain this performance gain. 

\begin{table}[t]
\centering
\resizebox{\columnwidth}{!}{%
\begin{tabular}{l c c}
\toprule
\textbf{Method} & \textbf{ACC} & \textbf{mF1} \\

\midrule
\multicolumn{3}{c}{\mymethod} \\
\midrule
Gemini-3-flash & \textbf{0.618 $\pm$ 0.002} & \textbf{0.62 $\pm$ 0.002} \\
Claude-4.6-sonnet  & 0.6 $\pm$ 0.003 & 0.595 $\pm$ 0.003 \\
GPT-5.4-mini       & 0.598 $\pm$ 0.003 & 0.59 $\pm$ 0.003 \\

\midrule
\multicolumn{3}{c}{Textual} \\
\midrule
Gemini-3-flash     & 0.605 $\pm$ 0.001 & 0.58  $\pm$ 0.001 \\
GPT-5.4-mini       & 0.584 $\pm$ 0.003 & 0.578 $\pm$ 0.003 \\
Claude-4.6-sonnet  & 0.577 $\pm$ 0.003 & 0.566 $\pm$ 0.003 \\
RoBERTa          & 0.526 $\pm$ 0.014 & 0.442 $\pm$ 0.034 \\
PT-HCL           & 0.517 $\pm$ 0.021 & 0.39 $\pm$ 0.048 \\
LKI-BART         & 0.466 $\pm$ 0.01 & 0.34 $\pm$ 0.011 \\
CoT Embeddings   & 0.424 $\pm$ 0.01 & 0.198 $\pm$ 0.003 \\

\bottomrule
\end{tabular}%
}
\caption{Stance detection performance on \textsf{CheeSE}, a German dataset without original news images.}
\label{tab:cheese_vlm_results}
\end{table}

\paragraph{Effectiveness Across Languages}
To assess whether \mymethod is effective across datasets and languages, we evaluated it on \textsf{CheeSE}~\cite{mascarell-etal-2021-stance}, a German article-level news stance detection dataset. Since \textsf{CheeSE} does not provide original news images, \mymethod was compared against seven textual baseline methods: three LVLM-based and four fine-tuned methods. As shown in Table~\ref{tab:cheese_vlm_results}, \mymethod achieved higher accuracy and macro F1 scores across multiple LVLM backbones. Gemini-3-flash again performed best, achieving an accuracy of 0.618 and a macro F1 score of 0.62, outperforming all baseline methods by a substantial margin (\textit{p}$<$0.01). These results, together with the Korean-language findings above, suggest that \mymethod can be effective across languages. To further support this finding, we provide supplementary results on translated versions of \textsf{K-News-Stance-MM} in four languages in Appendix~\ref{app:sec:morelanguages}.

\section{User Study}
\label{sec:case_study}

We conduct a controlled user study to examine whether images generated by \mymethod help news readers identify article stance. Specifically, we investigate whether participants can accurately discern article stance in a snippet-based news consumption setting where only limited textual information is available, reflecting common patterns of news consumption in online information environments and social feeds~\cite{gabielkov-etal-2016-social}. We recruited 200 native Korean speakers through PMI Research \& Consulting (PMI)\footnote{\url{https://pmirnc.com/}}, with the sample balanced by gender and age. 

For the user study, we selected nine articles covering three issues, with one supportive, one neutral, and one oppositional article per issue. These articles were published after June 2024, outside the period covered by \textsf{K-News-Stance-MM}. Each experimental snippet consisted of the article headline, the first two to three sentences of its lead paragraph, and a visual treatment determined by the experimental condition. For each article, we created four presentation versions with identical text: (1) \textbf{Text-only}, with no accompanying image; (2) \textbf{Original}, with the publisher's original image; (3) \textbf{Na\"ive}, with an image generated using a straightforward prompt without a visual framing specification; and (4) \textbf{Proposed}, with an image generated by \mymethod. 

Each participant viewed all nine articles in randomized order, with each article randomly assigned to one of the four presentation conditions. This design yielded approximately 50 observations per condition for each article and approximately 450 observations per condition overall. After viewing each snippet, participants classified the article's stance toward the target issue as supportive, neutral, or oppositional through a web survey interface, a screenshot of which is provided in Figure~\ref{fig:case_study_interface}.

\begin{figure}[t]
\centering
  \includegraphics[width=\columnwidth]{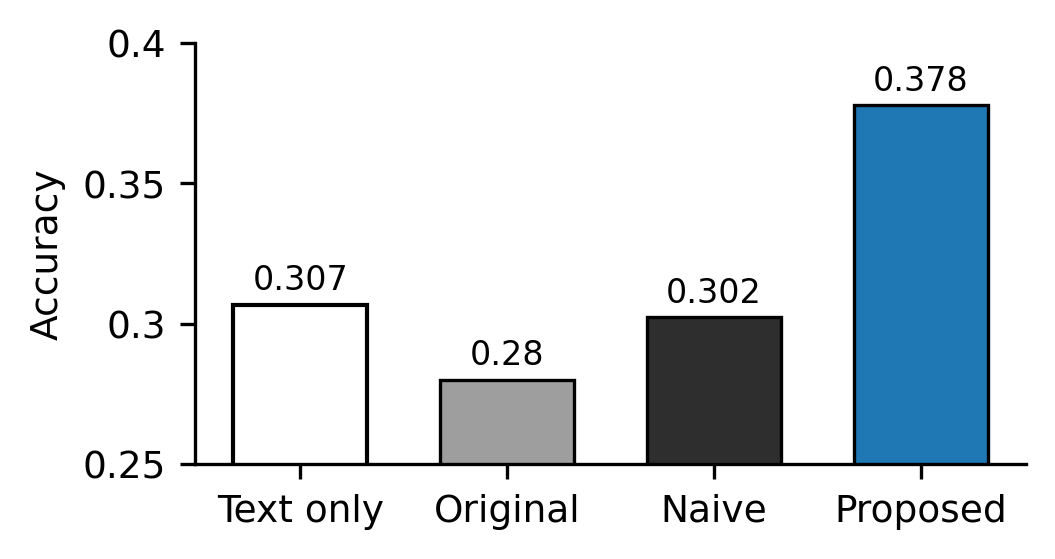}
  \caption{Stance identification accuracy across presentation conditions in a controlled user study under snippet-based news consumption.}  \label{fig:case_study}
\end{figure}

Figure~\ref{fig:case_study} reports stance identification accuracy across conditions. \mymethod achieved the highest accuracy of 0.378, exceeding the text-only, original, and na\"ive conditions by 0.071, 0.098, and 0.076, respectively. A mixed-effects logistic regression, with presentation condition as a fixed effect and random intercepts for participants and articles, further showed that all three comparison conditions had significantly lower odds of correct stance identification than \mymethod: text-only ($\mathrm{OR}=0.708$, $p=0.001$), original ($\mathrm{OR}=0.620$, $p<0.0001$), and na\"ive ($\mathrm{OR}=0.697$, $p=0.0006$). 

Despite modest overall accuracy, these findings indicate that \mymethod-generated images facilitate stance identification under abbreviated news exposure. Their advantage over text-only, publisher-provided and na\"ively generated images further highlights their potential utility beyond automated stance detection.

\section{Conclusion}

This study applies framing theory~\cite{entman-etal-1993-framing} and its visual extension~\cite{rodriguez-etal-2011-levels} to article-level news stance detection. Despite advances in NLP and LLMs, the task remains challenging because stance cues in long, structurally complex news articles are often implicit and dispersed throughout the text. \mymethod is a multi-stage, modular stance detection framework in which an LLM produces a visual framing specification, a T2I model generates a stance-aware image, and an LVLM predicts article-level stance using the article and the generated image. 

Evaluation results demonstrate that \mymethod outperforms existing stance detection methods and that both visual framing and image generation contribute to its performance. A controlled user study in a snippet-based news consumption setting further shows that images generated by \mymethod improve stance identification, shedding light on potential applications beyond automated stance detection. More broadly, the findings suggest that visual framing can function as an intermediate representational layer, transforming dispersed textual stance cues into signals that are more readily accessible to both computational models and human readers.

Taken together, these findings point to the potential of generative AI grounded in visual framing to make media perspectives more transparent, thereby supporting the identification of news bias and contributing to more pluralistic media environments. Future work could extend this approach to other areas of AI and NLP that involve implicit framing or evaluative signals, such as argument mining, bias analysis, and model bias auditing. The data, code, and prompts are available at \url{https://github.com/ssu-humane/VFStance}. 

\section*{Limitations}

\paragraph{Computational Costs} \mymethod employs three models across its corresponding stages. Encouragingly, as discussed in Section~\ref{sec:evaluation}, \textsc{\mymethod (Text)} offers a computationally efficient alternative by omitting image generation while still outperforming all baseline methods. This further demonstrates the effectiveness of the visual framing schema used in the proposed method. Considering the cost--accuracy trade-off (Appendix~\ref{app:sec:tradeoff}), \textsc{\mymethod (Text)} could be used when computational cost is the primary concern, whereas \mymethod remains the strongest choice when maximizing detection accuracy is the priority.

\paragraph{Multilingual Evaluation} The primary testbed, \textsf{K-News-Stance-MM}, is a Korean corpus, which limits the scope of multilingual evaluation in this study. This choice was necessary because \textsf{K-News-Stance-MM} is the first and only dataset to provide publisher images for article-level stance detection, which are required to compare \mymethod with visual and multimodal baselines. To provide additional evidence of multilingual effectiveness, we conducted experiments on \textsf{CheeSE}, a German text-only dataset, and on LLM-translated versions of \textsf{K-News-Stance-MM} in four languages (Appendix~\ref{app:sec:morelanguages}). Future studies could construct article-level stance detection datasets with accompanying news images by adapting the guidelines provided by \citet{lee-etal-2025-journalism}.

\paragraph{Model Selection} Proprietary models were selected as primary backbones for \mymethod because of their stronger language understanding, reasoning, and generation capabilities. We provide supplementary results for open-weight models in Appendix~\ref{app:sec:openweight}, where they achieved lower performance. Because the framework is modular and model-agnostic, future work could evaluate a broader range of models and incorporate additional training to further improve detection accuracy.

\section*{Ethical Considerations}
This study was approved by the Institutional Review Board at Soongsil University (SSU-202604-HR-804-1). 

\paragraph{Copyright and Privacy Issues}
\textsf{K-News-Stance-MM} extends an existing dataset~\cite{lee-etal-2025-journalism} built from news articles distributed through the Naver News platform. To respect the intellectual property rights of the original news publishers, \textsf{K-News-Stance-MM} is released under gated access with a custom Data Use Agreement: prospective users must submit an access request describing their research purpose and agree to terms that restrict use to non-commercial academic research and prohibit redistribution. The news data include the names of public figures, but private individuals are anonymized when present; thus, the dataset does not contain any personally identifiable information about private individuals, as manually verified for all samples in \textsf{K-News-Stance-MM}. \textsf{CheeSE}~\cite{mascarell-etal-2021-stance} is a publicly released benchmark and is used under its original release terms.

\paragraph{User Study Participants}
We recruited 200 Korean native speakers residing in South Korea through a survey panel provider. The task took approximately 10 minutes, with compensation of about USD~3.6, which exceeded the Korean statutory minimum hourly wage. All participants provided informed consent and could withdraw at any time. The study collected no sensitive personal information as defined by the Korean Personal Information Protection Act. Two potential risks were disclosed in advance: minimal-risk discomfort from politically contested content and the indirect inference of attitudes from stance judgments. Responses contained only randomly assigned IDs; therefore, individual records could not be selectively deleted after submission, a limitation that was also disclosed before consent.

\paragraph{Risks Associated with Image Generation}
Images generated by \mymethod could be repurposed for reader-facing applications, as demonstrated in the user study in Section~\ref{sec:case_study}. However, such use requires careful consideration because T2I models may reproduce or amplify social biases. Before its use, generated images should be clearly labeled as synthetic and reviewed to ensure compliance with applicable defamation, personality-rights, and synthetic-media regulations.

\paragraph{AI Assistant Use}
We used AI-assisted language editing tools, primarily ChatGPT, exclusively for grammar checking and improving readability.

\section*{Acknowledgements}
This research was supported by the Institute of Information \& Communications Technology Planning \& Evaluation (IITP), funded by the Korea government (MSIT) (IITP-2026-RS-2022-00156360, IITP-2026-RS-2024-00430997, and IITP-2026-RS-2020-II201602), and by the National Research Foundation of Korea (NRF), funded by the Korea government (MSIT) (RS-2023-00252535). KP and JH are the corresponding authors.

\bibliography{acl_latex}

@inproceedings{liang-etal-2022-zero,
  title = {Zero-Shot Stance Detection via Contrastive Learning},
  author = {Liang, Bin and Chen, Zixiao and Gui, Lin and He, Yulan and Yang, Min and Xu, Ruifeng},
  booktitle = {Proceedings of the ACM Web Conference 2022},
  pages = {2738--2747},
  year = {2022},
  publisher = {Association for Computing Machinery},
  doi = {10.1145/3485447.3511994},
  url = {https://doi.org/10.1145/3485447.3511994}
}

@inproceedings{gatto-etal-2023-chain,
  title = {Chain-of-Thought Embeddings for Stance Detection on Social Media},
  author = {Gatto, Joseph and Sharif, Omar and Preum, Sarah},
  booktitle = {Findings of the Association for Computational Linguistics: EMNLP 2023},
  month = dec,
  year = {2023},
  address = {Singapore},
  publisher = {Association for Computational Linguistics},
  url = {https://aclanthology.org/2023.findings-emnlp.273/},
  doi = {10.18653/v1/2023.findings-emnlp.273},
  pages = {4154--4161}
}

@inproceedings{zhang-etal-2024-llm,
  title = {{LLM}-Driven Knowledge Injection Advances Zero-Shot and Cross-Target Stance Detection},
  author = {Zhang, Zhao and Li, Yiming and Zhang, Jin and Xu, Hui},
  booktitle = {Proceedings of the 2024 Conference of the North American Chapter of the Association for Computational Linguistics: Human Language Technologies (Volume 2: Short Papers)},
  month = jun,
  year = {2024},
  address = {Mexico City, Mexico},
  publisher = {Association for Computational Linguistics},
  url = {https://aclanthology.org/2024.naacl-short.32/},
  doi = {10.18653/v1/2024.naacl-short.32},
  pages = {371--378}
}

@article{zhu-etal-2023-can,
  title = {Can {ChatGPT} Reproduce Human-Generated Labels? A Study of Social Computing Tasks},
  author = {Zhu, Yiming and Zhang, Peixian and Haq, Ehsan-Ul and Hui, Pan and Tyson, Gareth},
  journal = {arXiv preprint arXiv:2304.10145},
  year = {2023},
  url = {https://arxiv.org/abs/2304.10145}
}

@article{cruickshank-ng-2023-prompting,
author = {Cruickshank, Iain and Ng, Lynnette},
year = {2026},
month = {04},
pages = {1-25},
title = {Prompting and Fine-Tuning Open Source Large Language Models for Stance Classification},
volume = {17},
journal = {ACM Transactions on Intelligent Systems and Technology},
doi = {10.1145/3725816}
}

@inproceedings{ferreira-2016-emergent,
  title={Emergent: a novel data-set for stance classification},
  author={Ferreira, William and Vlachos, Andreas},
  booktitle={Proceedings of the 2016 conference of the North American chapter of the association for computational linguistics: Human language technologies},
  pages={1163--1168},
  year={2016}
}

@misc{pomerleau-rao-2017-fnc,
  title = {Fake News Challenge Stage 1 ({FNC-I}): Stance Detection},
  author = {Pomerleau, Dean and Rao, Delip},
  year = {2017},
  howpublished = {\url{http://www.fakenewschallenge.org}}
}

@inproceedings{hanselowski-etal-2018-retrospective,
  title = {A Retrospective Analysis of the Fake News Challenge Stance-Detection Task},
  author = {Hanselowski, Andreas and
    PVS, Avinesh and
    Schiller, Benjamin and
    Caspelherr, Felix and
    Chaudhuri, Debanjan and
    Meyer, Christian M. and
    Gurevych, Iryna},
  booktitle = {Proceedings of the 27th International Conference on Computational Linguistics},
  pages = {1859--1874},
  year = {2018},
  address = {Santa Fe, New Mexico, USA},
  publisher = {Association for Computational Linguistics},
  url = {https://aclanthology.org/C18-1158/}
}

@inproceedings{conforti-etal-2020-stander,
  title = {{STANDER}: An Expert-Annotated Dataset for News Stance Detection and Evidence Retrieval},
  author = {Conforti, Costanza and
    Berndt, Jakob and
    Pilehvar, Mohammad Taher and
    Giannitsarou, Chryssi and
    Toxvaerd, Flavio and
    Collier, Nigel},
  booktitle = {Findings of the Association for Computational Linguistics: {EMNLP} 2020},
  pages = {4086--4101},
  year = {2020},
  address = {Online},
  publisher = {Association for Computational Linguistics},
  doi = {10.18653/v1/2020.findings-emnlp.365},
  url = {https://aclanthology.org/2020.findings-emnlp.365/}
}

@inproceedings{mascarell-etal-2021-stance,
    title = "Stance Detection in {G}erman News Articles",
    author = "Mascarell, Laura  and
      Ruzsics, Tatyana  and
      Schneebeli, Christian  and
      Schlattner, Philippe  and
      Campanella, Luca  and
      Klingler, Severin  and
      Kadar, Cristina",
    editor = "Aly, Rami  and
      Christodoulopoulos, Christos  and
      Cocarascu, Oana  and
      Guo, Zhijiang  and
      Mittal, Arpit  and
      Schlichtkrull, Michael  and
      Thorne, James  and
      Vlachos, Andreas",
    booktitle = "Proceedings of the Fourth Workshop on Fact Extraction and VERification (FEVER)",
    month = nov,
    year = "2021",
    address = "Dominican Republic",
    publisher = "Association for Computational Linguistics",
    url = "https://aclanthology.org/2021.fever-1.8/",
    doi = "10.18653/v1/2021.fever-1.8",
    pages = "66--77",
}

@inproceedings{lee-etal-2025-journalism,
  title = {Journalism-Guided Agentic In-context Learning for News Stance Detection},
  author = {Lee, Dahyun and
    Choi, Jonghyeon and
    Han, Jiyoung and
    Park, Kunwoo},
  booktitle = {Proceedings of the 2025 Conference on Empirical Methods in Natural Language Processing},
  pages = {15393--15416},
  year = {2025},
  address = {Suzhou, China},
  publisher = {Association for Computational Linguistics},
  doi = {10.18653/v1/2025.emnlp-main.778},
  url = {https://aclanthology.org/2025.emnlp-main.778/}
}

@inproceedings{weinzierl-harabagiu-2023-identification,
    title = "Identification of Multimodal Stance Towards Frames of Communication",
    author = "Weinzierl, Maxwell  and
      Harabagiu, Sanda",
    editor = "Bouamor, Houda  and
      Pino, Juan  and
      Bali, Kalika",
    booktitle = "Proceedings of the 2023 Conference on Empirical Methods in Natural Language Processing",
    month = dec,
    year = "2023",
    address = "Singapore",
    publisher = "Association for Computational Linguistics",
    url = "https://aclanthology.org/2023.emnlp-main.776/",
    doi = "10.18653/v1/2023.emnlp-main.776",
    pages = "12597--12609"
}

@inproceedings{liang-etal-2024-multi,
  title = {Multi-modal Stance Detection: New Datasets and Model},
  author = {Liang, Bin and Li, Ang and Zhao, Jingqian and Gui, Lin and Yang, Min and Yu, Yue and Wong, Kam-Fai and Xu, Ruifeng},
  editor = {Ku, Lun-Wei and Martins, Andre and Srikumar, Vivek},
  booktitle = {Findings of the Association for Computational Linguistics: ACL 2024},
  month = aug,
  year = {2024},
  address = {Bangkok, Thailand},
  publisher = {Association for Computational Linguistics},
  url = {https://aclanthology.org/2024.findings-acl.736/},
  doi = {10.18653/v1/2024.findings-acl.736},
  pages = {12373--12387}
}

@inproceedings{zhang-etal-2025-mad,
    title = "{T}-{MAD}: Target-driven Multimodal Alignment for Stance Detection",
    author = "Zhang, ZhaoDan  and
      Zhang, Jin  and
      Cheng, Xueqi  and
      Xu, Hui",
    editor = "Christodoulopoulos, Christos  and
      Chakraborty, Tanmoy  and
      Rose, Carolyn  and
      Peng, Violet",
    booktitle = "Proceedings of the 2025 Conference on Empirical Methods in Natural Language Processing",
    month = nov,
    year = "2025",
    address = "Suzhou, China",
    publisher = "Association for Computational Linguistics",
    url = "https://aclanthology.org/2025.emnlp-main.30/",
    doi = "10.18653/v1/2025.emnlp-main.30",
    pages = "580--595",
    ISBN = "979-8-89176-332-6"
}

@inproceedings{zhang-etal-2025-exploring-artificial,
    title = "Exploring Artificial Image Generation for Stance Detection",
    author = "Zhang, Zhengkang  and
      Wang, Zhongqing  and
      Zhou, Guodong",
    editor = "Christodoulopoulos, Christos  and
      Chakraborty, Tanmoy  and
      Rose, Carolyn  and
      Peng, Violet",
    booktitle = "Proceedings of the 2025 Conference on Empirical Methods in Natural Language Processing",
    month = nov,
    year = "2025",
    address = "Suzhou, China",
    publisher = "Association for Computational Linguistics",
    url = "https://aclanthology.org/2025.emnlp-main.1004/",
    doi = "10.18653/v1/2025.emnlp-main.1004",
    pages = "19846--19861",
    ISBN = "979-8-89176-332-6"
}

@article{alshenaifi-etal-2026-beyond,
  title = {Beyond text: Multimodal stance detection in Arabic tweets},
  author = {AlShenaifi, Nouf and Alangari, Nourah},
  journal = {Machine Learning with Applications},
  volume = {23},
  pages = {100823},
  year = {2026},
  issn = {2666-8270},
  doi = {10.1016/j.mlwa.2025.100823},
  url = {https://www.sciencedirect.com/science/article/pii/S2666827025002063}
}

@inproceedings{weinzierl-harabagiu-2024-tree,
    title = "Tree-of-Counterfactual Prompting for Zero-Shot Stance Detection",
    author = "Weinzierl, Maxwell  and
      Harabagiu, Sanda",
    editor = "Ku, Lun-Wei  and
      Martins, Andre  and
      Srikumar, Vivek",
    booktitle = "Proceedings of the 62nd Annual Meeting of the Association for Computational Linguistics (Volume 1: Long Papers)",
    month = aug,
    year = "2024",
    address = "Bangkok, Thailand",
    publisher = "Association for Computational Linguistics",
    url = "https://aclanthology.org/2024.acl-long.49/",
    doi = "10.18653/v1/2024.acl-long.49",
    pages = "861--880"
}

@inproceedings{tourni-etal-2021-detecting,
  title = {Detecting Frames in News Headlines and Lead Images in {U}.{S}. Gun Violence Coverage},
  author = {Tourni, Isidora and Guo, Lei and Daryanto, Taufiq Husada and Zhafransyah, Fabian and Halim, Edward Edberg and Jalal, Mona and Chen, Boqi and Lai, Sha and Hu, Hengchang and Betke, Margrit and Ishwar, Prakash and Wijaya, Derry Tanti},
  booktitle = {Findings of the Association for Computational Linguistics: EMNLP 2021},
  year = {2021},
  month = nov,
  address = {Punta Cana, Dominican Republic},
  publisher = {Association for Computational Linguistics},
  url = {https://aclanthology.org/2021.findings-emnlp.339/},
  doi = {10.18653/v1/2021.findings-emnlp.339},
  pages = {4037--4050}
}

@inproceedings{arora-etal-2025-multimodal,
  title = {Multi-Modal Framing Analysis of News},
  author = {Arora, Arnav and Yadav, Srishti and Antoniak, Maria and Belongie, Serge and Augenstein, Isabelle},
  booktitle = {Proceedings of the 2025 Conference on Empirical Methods in Natural Language Processing},
  pages = {31531--31553},
  year = {2025},
  month = nov,
  address = {Suzhou, China},
  publisher = {Association for Computational Linguistics},
  doi = {10.18653/v1/2025.emnlp-main.1606},
  url = {https://aclanthology.org/2025.emnlp-main.1606/}
}

@inproceedings{lu-etal-2026-evaluating,
author = {Lu, Linqi and Wan, Zihan and Kwon, Hyerin and Kim, Sang Jung and Kang, Jiwon and Abbas, Laila and Liu, Jiawei and Mcleod, Douglas},
year = {2026},
month = {01},
pages = {},
title = {Evaluating Large Vision-Language Models for Visual Framing Analysis in News Imagery: A Theory-Driven Benchmark},
doi = {10.24251/HICSS.2026.325}
}

@article{rodriguez-etal-2011-levels,
  title={The levels of visual framing},
  author={Rodriguez, Lulu and Dimitrova, Daniela V},
  journal={Journal of visual literacy},
  volume={30},
  number={1},
  pages={48--65},
  year={2011},
  publisher={Taylor \& Francis}
}

@book{kress-etal-1996-reading,
  author    = {Kress, Gunther and van Leeuwen, Theo},
  title     = {Reading Images: The Grammar of Visual Design},
  year      = {1996},
  publisher = {Routledge},
  address   = {London and New York}
}

@article{entman-etal-1993-framing,
  author  = {Entman, Robert M.},
  title   = {Framing: Toward Clarification of a Fractured Paradigm},
  journal = {Journal of Communication},
  volume  = {43},
  number  = {4},
  pages   = {51--58},
  year    = {1993},
  doi     = {10.1111/j.1460-2466.1993.tb01304.x}
}

@incollection{messaris-etal-2001-role,
  author    = {Messaris, Paul and Abraham, Linus},
  title     = {The Role of Images in Framing News Stories},
  booktitle = {Framing Public Life: Perspectives on Media and Our Understanding of the Social World},
  editor    = {Reese, Stephen D. and Gandy, Oscar H. and Grant, August E.},
  publisher = {Lawrence Erlbaum Associates},
  address   = {Mahwah, NJ},
  pages     = {215--226},
  year      = {2001},
  doi       = {10.4324/9781410605689-22}
}

@inproceedings{he-etal-2016-deep,
  title = {Deep Residual Learning for Image Recognition},
  author = {He, Kaiming and Zhang, Xiangyu and Ren, Shaoqing and Sun, Jian},
  booktitle = {Proceedings of the IEEE Conference on Computer Vision and Pattern Recognition},
  pages = {770--778},
  year = {2016}
}

@inproceedings{dosovitskiy-etal-2021-image,
  author       = {Alexey Dosovitskiy and
                  Lucas Beyer and
                  Alexander Kolesnikov and
                  Dirk Weissenborn and
                  Xiaohua Zhai and
                  Thomas Unterthiner and
                  Mostafa Dehghani and
                  Matthias Minderer and
                  Georg Heigold and
                  Sylvain Gelly and
                  Jakob Uszkoreit and
                  Neil Houlsby},
  title        = {An Image is Worth 16x16 Words: Transformers for Image Recognition
                  at Scale},
  booktitle    = {9th International Conference on Learning Representations, {ICLR} 2021,
                  Virtual Event, Austria, May 3-7, 2021},
  publisher    = {OpenReview.net},
  year         = {2021},
  url          = {https://openreview.net/forum?id=YicbFdNTTy},
  bibsource    = {dblp computer science bibliography, https://dblp.org}
}

@inproceedings{liu-etal-2021-swin,
  title = {Swin Transformer: Hierarchical Vision Transformer Using Shifted Windows},
  author = {Liu, Ze and Lin, Yutong and Cao, Yue and Hu, Han and Wei, Yixuan and Zhang, Zheng and Lin, Stephen and Guo, Baining},
  booktitle = {Proceedings of the IEEE/CVF International Conference on Computer Vision},
  pages = {10012--10022},
  year = {2021}
}

@inproceedings{mohammad-etal-2016-semeval,
    title = "{S}em{E}val-2016 Task 6: Detecting Stance in Tweets",
    author = "Mohammad, Saif  and
      Kiritchenko, Svetlana  and
      Sobhani, Parinaz  and
      Zhu, Xiaodan  and
      Cherry, Colin",
    editor = "Bethard, Steven  and
      Carpuat, Marine  and
      Cer, Daniel  and
      Jurgens, David  and
      Nakov, Preslav  and
      Zesch, Torsten",
    booktitle = "Proceedings of the 10th International Workshop on Semantic Evaluation ({S}em{E}val-2016)",
    month = jun,
    year = "2016",
    address = "San Diego, California",
    publisher = "Association for Computational Linguistics",
    url = "https://aclanthology.org/S16-1003/",
    doi = "10.18653/v1/S16-1003",
    pages = "31--41"
}

@incollection{barthes-etal-1977-rhetoric,
  author    = {Barthes, Roland},
  title     = {Rhetoric of the Image},
  booktitle = {Image, Music, Text},
  editor    = {Heath, Stephen},
  pages     = {32--51},
  publisher = {Hill and Wang},
  address   = {New York},
  year      = {1977}
}

@book{hall-etal-1966-hidden,
  author    = {Hall, Edward T.},
  title     = {The Hidden Dimension},
  publisher = {Doubleday},
  address   = {Garden City, NY},
  year      = {1966}
}

@inproceedings{ali-etal-2022-survey,
  title = {A Survey of Computational Framing Analysis Approaches},
  author = {Ali, Mohammad and Hassan, Naeemul},
  editor = {Goldberg, Yoav and Kozareva, Zornitsa and Zhang, Yue},
  booktitle = {Proceedings of the 2022 Conference on Empirical Methods in Natural Language Processing},
  month = dec,
  year = {2022},
  address = {Abu Dhabi, United Arab Emirates},
  publisher = {Association for Computational Linguistics},
  url = {https://aclanthology.org/2022.emnlp-main.633/},
  doi = {10.18653/v1/2022.emnlp-main.633},
  pages = {9335--9348}
}

@inproceedings{otmakhova-etal-2024-media,
  title = {Media Framing: A Typology and Survey of Computational Approaches Across Disciplines},
  author = {Otmakhova, Yulia and Khanehzar, Shima and Frermann, Lea},
  editor = {Ku, Lun-Wei and Martins, Andre and Srikumar, Vivek},
  booktitle = {Proceedings of the 62nd Annual Meeting of the Association for Computational Linguistics (Volume 1: Long Papers)},
  month = aug,
  year = {2024},
  address = {Bangkok, Thailand},
  publisher = {Association for Computational Linguistics},
  url = {https://aclanthology.org/2024.acl-long.822/},
  doi = {10.18653/v1/2024.acl-long.822},
  pages = {15407--15428}
}

@inproceedings{vallejo-etal-2024-connecting,
  title = {Connecting the Dots in News Analysis: Bridging the Cross-Disciplinary Disparities in Media Bias and Framing},
  author = {Vallejo, Gisela and Baldwin, Timothy and Frermann, Lea},
  editor = {Card, Dallas and Field, Anjalie and Hovy, Dirk and Keith, Katherine},
  booktitle = {Proceedings of the Sixth Workshop on Natural Language Processing and Computational Social Science (NLP+CSS 2024)},
  month = jun,
  year = {2024},
  address = {Mexico City, Mexico},
  publisher = {Association for Computational Linguistics},
  url = {https://aclanthology.org/2024.nlpcss-1.2/},
  doi = {10.18653/v1/2024.nlpcss-1.2},
  pages = {16--31}
}

@inproceedings{card-etal-2015-media,
  title = {The Media Frames Corpus: Annotations of Frames Across Issues},
  author = {Card, Dallas and Boydstun, Amber E. and Gross, Justin H. and Resnik, Philip and Smith, Noah A.},
  editor = {Zong, Chengqing and Strube, Michael},
  booktitle = {Proceedings of the 53rd Annual Meeting of the Association for Computational Linguistics and the 7th International Joint Conference on Natural Language Processing (Volume 2: Short Papers)},
  month = jul,
  year = {2015},
  address = {Beijing, China},
  publisher = {Association for Computational Linguistics},
  url = {https://aclanthology.org/P15-2072/},
  doi = {10.3115/v1/P15-2072},
  pages = {438--444}
}

@inproceedings{liu-etal-2019-detecting,
  title = {Detecting Frames in News Headlines and Its Application to Analyzing News Framing Trends Surrounding {U}.{S}. Gun Violence},
  author = {Liu, Siyi and Guo, Lei and Mays, Kate and Betke, Margrit and Wijaya, Derry Tanti},
  editor = {Bansal, Mohit and Villavicencio, Aline},
  booktitle = {Proceedings of the 23rd Conference on Computational Natural Language Learning (CoNLL)},
  month = nov,
  year = {2019},
  address = {Hong Kong, China},
  publisher = {Association for Computational Linguistics},
  url = {https://aclanthology.org/K19-1047/},
  doi = {10.18653/v1/K19-1047},
  pages = {504--514}
}

@inproceedings{piskorski-etal-2023-semeval,
  title = {{S}em{E}val-2023 Task 3: Detecting the Category, the Framing, and the Persuasion Techniques in Online News in a Multi-lingual Setup},
  author = {Piskorski, Jakub and Stefanovitch, Nicolas and Da San Martino, Giovanni and Nakov, Preslav},
  editor = {Ojha, Atul Kr. and Do{\u{g}}ru{\"o}z, A. Seza and Da San Martino, Giovanni and Tayyar Madabushi, Harish and Kumar, Ritesh and Sartori, Elisa},
  booktitle = {Proceedings of the 17th International Workshop on Semantic Evaluation (SemEval-2023)},
  month = jul,
  year = {2023},
  address = {Toronto, Canada},
  publisher = {Association for Computational Linguistics},
  url = {https://aclanthology.org/2023.semeval-1.317/},
  doi = {10.18653/v1/2023.semeval-1.317},
  pages = {2343--2361}
}

@article{liu-etal-2019-roberta,
  title={RoBERTa: A Robustly Optimized BERT Pretraining Approach},
  author={Liu, Yinhan and Ott, Myle and Goyal, Naman and Du, Jingfei and Joshi, Mandar and Chen, Danqi and Levy, Omer and Lewis, Mike and Zettlemoyer, Luke and Stoyanov, Veselin},
  journal={arXiv preprint arXiv:1907.11692},
  year={2019},
  url={https://arxiv.org/abs/1907.11692}
}

@InProceedings{radford-etal-2021-learning,
  title = {Learning Transferable Visual Models From Natural Language Supervision},
  author = {Radford, Alec and Kim, Jong Wook and Hallacy, Chris and Ramesh, Aditya and Goh, Gabriel and Agarwal, Sandhini and Sastry, Girish and Askell, Amanda and Mishkin, Pamela and Clark, Jack and Krueger, Gretchen and Sutskever, Ilya},
  booktitle = {Proceedings of the 38th International Conference on Machine Learning},
  pages = {8748--8763},
  year = {2021},
  editor = {Meila, Marina and Zhang, Tong},
  volume = {139},
  series = {Proceedings of Machine Learning Research},
  month = {18--24 Jul},
  publisher = {PMLR},
  url = {https://proceedings.mlr.press/v139/radford21a.html},
}

@book{kovach-etal-2021-elements,
  title={The elements of journalism, revised and updated 4th edition: What newspeople should know and the public should expect},
  author={Kovach, Bill and Rosenstiel, Tom},
  year={2021},
  publisher={Crown}
}

@misc{associatedpress-2024-news-values,
  title        = {AP News Values and Principles},
  author       = {{Associated Press}},
  year         = {2024},
  howpublished = {\url{https://www.ap.org/wp-content/uploads/2024/02/ap-news-values-and-principles-1.pdf}},
  note         = {Accessed: 2026-05-18}
}

@book{reuters-2008-handbook,
  author    = {{Reuters}},
  title     = {Reuters Handbook of Journalism},
  year      = {2008},
  publisher = {Thomson Reuters}
}

@book{mehrabian-etal-1981-silent,
  author    = {Mehrabian, Albert},
  title     = {Silent Messages: Implicit Communication of Emotions and Attitudes},
  edition   = {2nd},
  year      = {1981},
  publisher = {Wadsworth},
  address   = {Belmont, CA}
}

@book{paivio-etal-1986-mental,
  author    = {Paivio, Allan},
  title     = {Mental Representations: A Dual Coding Approach},
  year      = {1986},
  publisher = {Oxford University Press},
  address   = {New York}
}

@article{eldamanhoury-etal-2026-visual,
    title     = {Visual Framing in the {AI} Era: Lessons from Manual Approaches for Computational Methods},
    author    = {El Damanhoury, Kareem and Winkler, Carol and Lokmanoglu, Ayse D. and Chen Glanz, Keyu Alexander},
    journal   = {Computational Communication Research},
    volume    = {8},
    number    = {1},
    pages     = {1--41},
    year      = {2026},
    doi       = {10.5117/CCR2026.1.2.ELDA}
}

@inproceedings{liao-etal-2024-tiac,
  title     = {Text-to-Image Generation for Abstract Concepts},
  author    = {Liao, Jiayi and Chen, Xu and Fu, Qiang and Du, Lun and He, Xiangnan and Wang, Xiang and Han, Shi and Zhang, Dongmei},
  booktitle = {Proceedings of the AAAI Conference on Artificial Intelligence},
  year      = {2024},
  volume    = {38},
  url       = {https://ojs.aaai.org/index.php/AAAI/article/view/28122}
}

@article{park-etal-2021-klue,
  title={KLUE: Korean Language Understanding Evaluation},
  author={Park, Sungjoon and Moon, Jihyung and Kim, Sungdong and Cho, Won Ik and Han, Ji Yoon and Park, Jangwon and Song, Chisung and Kim, Junseong and Song, Youngsook and Oh, Taehwan and others},
  booktitle={Thirty-fifth Conference on Neural Information Processing Systems Datasets and Benchmarks Track (Round 2)},
  year={2021}
}

@article{schwartz-etal-1992-photojournalism,
  title={To tell the truth : codes of objectivity in photojournalism},
  author={Dona Schwartz},
  journal={Communicatio},
  year={1992},
  volume={13},
  pages={95-109},
  url={https://api.semanticscholar.org/CorpusID:147250355}
}

@incollection{coleman-etal-2010-framing,
  title={Framing the pictures in our heads: Exploring the framing and agenda-setting effects of visual images},
  author={Coleman, Renita},
  booktitle={Doing news framing analysis},
  pages={249--278},
  year={2010},
  publisher={Routledge}
}

@article{gentzkow-etal-2010-drives,
author = {Gentzkow, Matthew and Shapiro, Jesse M.},
title = {What Drives Media Slant? Evidence From U.S. Daily Newspapers},
journal = {Econometrica},
volume = {78},
number = {1},
pages = {35-71},
doi = {https://doi.org/10.3982/ECTA7195},
url = {https://onlinelibrary.wiley.com/doi/abs/10.3982/ECTA7195},
eprint = {https://onlinelibrary.wiley.com/doi/pdf/10.3982/ECTA7195},
year = {2010}
}

@article{hamborg-etal-2019-automated,
  year={2019},
  doi={10.1007/s00799-018-0261-y},
  title={Automated identification of media bias in news articles : an interdisciplinary literature review},
  number={4},
  volume={20},
  issn={1432-5012},
  journal={International Journal on Digital Libraries},
  pages={391--415},
  author={Hamborg, Felix and Donnay, Karsten and Gipp, Bela}
}

@article{chong-etal-2007-framing,
  title={Framing theory},
  author={Chong, Dennis and Druckman, James N},
  journal={Annu. Rev. Polit. Sci.},
  volume={10},
  number={1},
  pages={103--126},
  year={2007},
  publisher={Annual Reviews}
}

@inproceedings{park-etal-2009-newscube,
  title={NewsCube: delivering multiple aspects of news to mitigate media bias},
  author={Park, Souneil and Kang, Seungwoo and Chung, Sangyoung and Song, Junehwa},
  booktitle={Proceedings of the SIGCHI conference on human factors in computing systems},
  pages={443--452},
  year={2009}
}

@misc{openai-etal-2026-gpt54mini,
  author       = {{OpenAI}},
  title        = {GPT-5.4 mini},
  year         = {2026},
  howpublished = {\url{https://developers.openai.com/api/docs/models/gpt-5.4-mini}},
  note         = {OpenAI API model documentation. Accessed: 2026-05-24}
}

@misc{google-etal-2025-gemini3flash,
  author       = {{Google DeepMind}},
  title        = {Gemini 3 Flash Model Card},
  year         = {2025},
  howpublished = {\url{https://storage.googleapis.com/deepmind-media/Model-Cards/Gemini-3-Flash-Model-Card.pdf}},
  note         = {Published: December 2025. Accessed: 2026-05-24}
}

@misc{google-etal-2026-gemini31flashimage,
  author = {{Google}},
  title = {Gemini 3.1 Flash Image Preview},
  year = {2026},
  howpublished = {\url{https://ai.google.dev/gemini-api/docs/models/gemini-3.1-flash-image}},
  note = {Accessed: 2026-05-25}
}

@misc{google-etal-2026-gemini31pro,
  author       = {{Google}},
  title        = {{Gemini 3.1 Pro Preview}},
  year         = {2026},
  howpublished = {\url{https://ai.google.dev/gemini-api/docs/models/gemini-3.1-pro-preview}},
  note         = {Accessed: 2026-05-26}
}

@misc{anthropic-etal-2026-claude46sonnet,
  author       = {{Anthropic}},
  title        = {Claude Sonnet 4.6 System Card},
  year         = {2026},
  howpublished = {\url{https://www-cdn.anthropic.com/78073f739564e986ff3e28522761a7a0b4484f84.pdf}},
  note         = {Published: February 17, 2026. Accessed: 2026-05-24}
}

@inproceedings{gabielkov-etal-2016-social,
author = {Gabielkov, Maksym and Ramachandran, Arthi and Chaintreau, Augustin and Legout, Arnaud},
title = {Social Clicks: What and Who Gets Read on Twitter?},
year = {2016},
isbn = {9781450342667},
publisher = {Association for Computing Machinery},
address = {New York, NY, USA},
url = {https://doi.org/10.1145/2896377.2901462},
doi = {10.1145/2896377.2901462},
booktitle = {Proceedings of the 2016 ACM SIGMETRICS International Conference on Measurement and Modeling of Computer Science},
pages = {179–192},
numpages = {14},
location = {Antibes Juan-les-Pins, France},
series = {SIGMETRICS '16}
}

@inproceedings{joshi-etal-2020-state,
    title = "The State and Fate of Linguistic Diversity and Inclusion in the {NLP} World",
    author = "Joshi, Pratik  and
      Santy, Sebastin  and
      Budhiraja, Amar  and
      Bali, Kalika  and
      Choudhury, Monojit",
    editor = "Jurafsky, Dan  and
      Chai, Joyce  and
      Schluter, Natalie  and
      Tetreault, Joel",
    booktitle = "Proceedings of the 58th Annual Meeting of the Association for Computational Linguistics",
    month = jul,
    year = "2020",
    address = "Online",
    publisher = "Association for Computational Linguistics",
    url = "https://aclanthology.org/2020.acl-main.560/",
    doi = "10.18653/v1/2020.acl-main.560",
    pages = "6282--6293"
}

\appendix

\setcounter{figure}{0}
\setcounter{table}{0}
\renewcommand{\thefigure}{A\arabic{figure}}
\renewcommand{\thetable}{A\arabic{table}}

\section{Experimental Setups}
\label{app:sec:experimental_setups}

This section provides the experimental details of our study. Each result is reported as the average over five runs with the standard error: trainable models were run using random seeds 42--46, while API-based models were evaluated five times under the same inference configuration because not all backbones support user-specified seeds. Experiments were conducted on three NVIDIA RTX A6000 GPUs (48GB each) with 128GB RAM, using Python 3.10, PyTorch 2.4.1, Transformers 4.57.1, and CUDA 12.1.

We accessed GPT-5.4-mini~\cite{openai-etal-2026-gpt54mini}, Claude-4.6-sonnet~\cite{anthropic-etal-2026-claude46sonnet}, and Gemini-3-flash~\cite{google-etal-2025-gemini3flash} via API, with the thinking level set to low, a maximum of 16,000 output tokens, and a temperature of 1 for all models, since Claude-4.6-sonnet fixes the temperature to 1 when reasoning is enabled. Images were generated with Gemini-3.1-flash-image~\cite{google-etal-2026-gemini31flashimage} under the default configuration (1K resolution), and automatic image quality assessment used Gemini-3.1-Pro~\cite{google-etal-2026-gemini31pro} with default settings.

For textual baselines, we used KLUE-RoBERTa-large~\cite{park-etal-2021-klue} for \textsf{K-News-Stance-MM} and XLM-RoBERTa-large for \textsf{CheeSE} as the backbones for RoBERTa, CoT Embeddings, and PT-HCL. For LKI-BART, we used KoBART-base-v2 and BART-qg-German for \textsf{K-News-Stance-MM} and \textsf{CheeSE}, respectively. We set the learning rate to $3\times10^{-5}$, used a batch size of 32 for CoT Embeddings and 16 for the other baselines, froze the bottom seven layers, and used GPT-4o-mini as the LLM for CoT Embeddings and LKI-BART.

For visual baselines, we used ResNet-50, ViT-B/16, and SwinV2-Base with a batch size of 32, a linear scheduler with a warmup ratio of 0.1, and early stopping with a patience of 3. The learning rate was set to $1\times10^{-4}$ for ResNet-50 and $5\times10^{-5}$ for ViT-B/16 and SwinV2-Base, using the default image preprocessing provided with each checkpoint. For multimodal baselines, we used KLUE-RoBERTa-large and ViT-B/16 as the text and visual encoders, respectively, for RoBERTa+ViT, TMPT, and T-MAD, while a Korean CLIP variant was used for the CLIP baseline. We used a learning rate of $2\times10^{-5}$ and a batch size of 16 for these baselines.

For EAIG4SD~\citep{zhang-etal-2025-exploring-artificial}, we followed the original setup (Stable Diffusion 3 Medium, 28 denoising steps, and a guidance scale of 7), replacing unsupported components with Korean-capable alternatives: Qwen2.5-VL-7B-Instruct for the intermediate stance and sentiment prediction used in prompt construction (LoRA fine-tuned with rank 16, alpha 32, learning rate $1\times10^{-4}$, and 3 epochs) and a Korean CLIP variant for image--text similarity, with images selected via PageRank (damping factor 0.85) over the similarity graph.

For the open-weight experiments (Appendix~\ref{app:sec:openweight}), InternVL3-14B-Instruct and Gemma3-12B-Instruct were evaluated zero-shot with the same prompts used for the proprietary models, using a temperature of 1 with a maximum of 1,024 output tokens for annotation and 16 for stance prediction. Stable Diffusion 3.5 Large generated one image per article with 28 denoising steps and a guidance scale of 4.5.

All fine-tuned models were trained with the AdamW optimizer and a weight decay of 0.01. Unless otherwise specified, hyperparameters were selected using validation subsets held out from the training split, and other method-specific settings followed the original studies.

The model IDs and parameter sizes used in the experiments are provided below.
\begin{itemize}
    \item GPT-5.4-mini: \texttt{gpt-5.4-mini-2026-03-17} (Parameter size: unknown)
    \item Claude-4.6-sonnet: \texttt{claude-sonnet-4-6} (Parameter size: unknown)
    \item Gemini-3-flash: \texttt{gemini-3-flash-preview} (Parameter size: unknown)
    \item Gemini-3.1-flash-image: \texttt{gemini-3.1-flash-image-preview} (Parameter size: unknown)
    \item GPT-4o-mini: \texttt{gpt-4o-mini} (Parameter size: unknown)
    \item Qwen2.5-VL-7B-Instruct: \url{https://huggingface.co/Qwen/Qwen2.5-VL-7B-Instruct} (Parameter size: 7B)
    \item InternVL3-14B-Instruct: \url{https://huggingface.co/OpenGVLab/InternVL3-14B-Instruct} (Parameter size: 15B)
    \item Gemma3-12B-Instruct: \url{https://huggingface.co/google/gemma-3-12b-it} (Parameter size: 12B)
    \item KLUE-RoBERTa-large: \url{https://huggingface.co/klue/roberta-large} (Parameter size: 337M)
    \item XLM-RoBERTa-large: \url{https://huggingface.co/FacebookAI/xlm-roberta-large} (Parameter size: 561M)
    \item BART-qg-German: \url{https://huggingface.co/su157/bart-qg-german} (Parameter size: 139M)
    \item KoBART-base-v2: \url{https://huggingface.co/gogamza/kobart-base-v2} (Parameter size: 124M)
    \item ResNet-50: \url{https://huggingface.co/microsoft/resnet-50} (Parameter size: 26M)
    \item ViT-base-patch16-224: \url{https://huggingface.co/google/vit-base-patch16-224} (Parameter size: 86M)
    \item SwinV2-Base:
    \url{https://huggingface.co/microsoft/swinv2-base-patch4-window12-192-22k} (Parameter size: 88M)
    \item KoCLIP: \url{https://huggingface.co/Bingsu/clip-vit-large-patch14-ko} (Parameter size: 428M)
    \item Stable-diffusion-3-medium: \url{https://huggingface.co/stabilityai/stable-diffusion-3-medium-diffusers} (Parameter size: 2B)
    \item Stable-diffusion-3.5-large: \url{https://huggingface.co/stabilityai/stable-diffusion-3.5-large} (Parameter size: 8B)
    \item Gemini-3.1-pro: \texttt{gemini-3.1-pro-preview} (Parameter size: unknown)
\end{itemize}


\begin{table}[t]
\centering
\subcaptionbox{\textsf{K-News-Stance-MM}}{
\resizebox{\columnwidth}{!}{%
\begin{tabular}{lcccc}
\toprule
\textbf{Split} & \textbf{Total} & \textbf{Supportive} & \textbf{Neutral} & \textbf{Oppositional} \\
\midrule
Train & 909 & 279 & 310 & 320 \\
Test  & 907 & 289 & 305 & 313 \\
\bottomrule
\end{tabular}%
}
}

\vspace{0.8em}

\subcaptionbox{\textsf{CheeSE}}{
\resizebox{\columnwidth}{!}{%
\begin{tabular}{lcccc}
\toprule
\textbf{Split} & \textbf{Total} & \textbf{In Favor} & \textbf{Discussing} & \textbf{Against} \\
\midrule
Train & 1000 & 399 & 439 & 162 \\
Test  & 762  & 303 & 335 & 124 \\
\bottomrule
\end{tabular}%
}
}

\caption{Dataset sizes and label distributions across the data splits.}
\label{tab:dataset_statistics}
\end{table}

\section{Dataset Details}
\label{app:sec:dataset}

Table~\ref{tab:dataset_statistics} presents the dataset sizes and label distributions across the data splits. Below we provide detailed information about each dataset.

\textsf{K-News-Stance-MM} is a multimodal extension of the existing article-level stance detection dataset~\cite{lee-etal-2025-journalism}. The original dataset consists of 2,000 news articles in Korean, with 999 articles for training and 1,001 for testing. Each article's stance toward a social issue (e.g., ``The passage of the Yellow Envelope Act in a National Assembly standing committee'') is labeled as \emph{supportive}, \emph{neutral}, or \emph{oppositional}. We extended the dataset by crawling news images from the original webpages of each article, selecting the lead image (i.e., the first image displayed) when multiple were available. We successfully collected images for 1,816 articles (90.8\%), comprising 909 training samples and 907 testing samples. Following the original dataset, we preserved the issue-level train/test split to prevent issue-level leakage across the splits. To our knowledge, this is the first dataset to provide article-level stance labels together with news images, and it serves as the primary testbed for this study. An example instance is provided in Table~\ref{tab:dataset_example_en} with an English translation. For its broader accessibility, we provide translations into four widely spoken languages: English, Chinese, Indonesian, and Arabic through the GitHub repository. Gemini-3-flash was used for translation, while the stance labels and data splits were kept unchanged. A preliminary comparison of \mymethod with a text-only baseline is provided in Appendix~\ref{app:sec:morelanguages}.

\textsf{CheeSE} is a news stance detection dataset consisting of German news articles~\cite{mascarell-etal-2021-stance}. The original dataset contains 3,693 news articles with article-level stance annotations with respect to debate questions (e.g., ``Are abortions morally acceptable?''). We excluded 503 \textit{unklar} (unclear) samples and 1,428 \textit{Kein Bezug} (unrelated) samples, resulting in 1,762 samples. We split the resulting dataset into 800/200/762 training, validation, and test samples, respectively, while preserving label distributions across the splits. We use this text-only dataset to evaluate the proposed method in a setting without original news images.

\section{Method Details}
\label{app:sec:method}

This section provides details on the baseline and proposed methods. Experimental settings for all methods are summarized in Appendix~\ref{app:sec:experimental_setups}.

\subsection{Baseline Methods}

\paragraph{Textual Baselines}
For RoBERTa, the input sequence is constructed by concatenating the target issue, headline, and article text as
\texttt{[CLS] issue [SEP] headline [SEP] article [SEP]}, and the \texttt{[CLS]} representation is used for stance classification.
CoT Embeddings~\cite{gatto-etal-2023-chain} augments this input with an LLM-generated rationale explaining the article's stance toward the target issue.
LKI-BART~\cite{zhang-etal-2024-llm} first generates stance-relevant background knowledge from the target issue and article and incorporates it into its generation-based stance prediction framework.
PT-HCL~\cite{liang-etal-2022-zero} uses the same textual input format and jointly optimizes stance classification with hierarchical contrastive objectives. The zero-shot LVLM baseline in this category uses the prompt shown in Figure~\ref{fig:text_only_prompt}.

\paragraph{Visual Baselines}
ResNet, ViT, and SwinT predict stance using only the original publisher image associated with each article. The zero-shot LVLM baseline uses the prompt shown in Figure~\ref{fig:image_only_prompt}.

\paragraph{Multimodal Baselines}
For RoBERTa+ViT, RoBERTa encodes the target issue, headline, and article, ViT encodes the original publisher image, and the resulting \texttt{[CLS]} representations are concatenated for stance prediction. For CLIP, text and image embeddings from the pretrained encoders are concatenated and passed through a classification layer. TMPT~\cite{liang-etal-2024-multi} augments both inputs with target-aware prompts before multimodal fusion, and T-MAD~\cite{zhang-etal-2025-mad} uses the separately encoded target representation to derive target-aligned multimodal features. The zero-shot LVLM baseline uses the prompt shown in Figure~\ref{fig:multimodal_prompt}, which is identical to the Stage~3 prompt of \mymethod except that the original news image is provided.

\paragraph{EAIG4SD}
EAIG4SD~\cite{zhang-etal-2025-exploring-artificial} generates multiple stance-aware candidate images from prompts constructed using the article content, target issue, predicted stance, and sentiment, and selects the final image via PageRank using text--image similarity, target consistency, and stance consistency signals.
Since we use EAIG4SD only as an image-generation baseline, we apply its image-generation and selection procedures and feed the selected image to the LVLM used in Stage~3 of \mymethod, with the same prompt (Figure~\ref{fig:multimodal_prompt}), for final stance prediction.

\subsection{\mymethod}
\label{app:sec:method:mymethod}

\begin{table}[t]
\centering
\resizebox{\columnwidth}{!}{%
\begin{tabular}{@{} l l l @{}}
\toprule
\textbf{Level} & \textbf{Feature} & \textbf{Type}\\
\midrule
\textbf{Ideological}
  & -
  & Open-ended \\
\midrule
\textbf{Connotative}
  & -
  & Open-ended \\

\midrule

\multirow{6}{*}{\textbf{Stylistic-Semiotic}}
  & Style
  & Categorical (photo, illustration) \\
  
  & Composition
  & Categorical (centered, split, asymmetric, crowded) \\
  
  & Angle
  & Categorical (low, eye-level, high) \\

  & Distance
  & Categorical (close-up, medium, long) \\

  & Saturation
  & Categorical (saturated, neutral, desaturated) \\

  & Luminosity
  & Categorical (bright, neutral, dark) \\
\midrule

\multirow{2}{*}{\textbf{Denotative}}
  & Inclusion
  & Open-ended \\
  & Exclusion
  & Open-ended \\
\bottomrule
\end{tabular}%
}
\caption{Visual framing schema used in Stage 1.}
\label{tab:visual_framing_features}
\end{table}

\paragraph{Visual Framing Features}
We describe the literature grounding each feature in our schema (Table~\ref{tab:visual_framing_features}). The four-level structure of the schema follows Rodriguez and Dimitrova's (\citeyear{rodriguez-etal-2011-levels}) model of visual framing, in which lower levels capture concrete visual elements and higher levels capture their interpretive meanings. At the denotative level, the \emph{inclusion} and \emph{exclusion} features operationalize the framing functions of selection, emphasis, and omission~\cite{rodriguez-etal-2011-levels, messaris-etal-2001-role, entman-etal-1993-framing}: they specify which actors, objects, and scenes are foregrounded in or omitted from the image. At the stylistic-semiotic level, the \emph{composition} and \emph{angle} features derive from the grammar of visual design~\cite{kress-etal-1996-reading}, particularly the spatial organization of visual elements and the interpersonal meanings of viewing position; the \emph{distance} feature additionally reflects proxemic and social-distance cues~\cite{hall-etal-1966-hidden, kress-etal-1996-reading}. The \emph{saturation} and \emph{luminosity} features correspond to color intensity and brightness as cues of visual modality and expressive tone~\cite{kress-etal-1996-reading}. The \emph{style} feature is a generation-oriented operationalization of representational modality~\cite{messaris-etal-2001-role, kress-etal-1996-reading}. The connotative level captures symbolic or associative meaning beyond literal depiction~\cite{rodriguez-etal-2011-levels, barthes-etal-1977-rhetoric}, and the ideological level concerns the perspectives and interests served by the image~\cite{rodriguez-etal-2011-levels}; both are annotated in Stage~1 but not rendered in Stage~2, providing the interpretive context for the lower-level features.

\paragraph{Prompts and Examples}
We describe the prompts used in each stage of \mymethod using a running example. This manuscript presents English translations of the prompts, while the original Korean prompts and examples are made publicly available in our GitHub repository.

In Stage~1, the LLM receives a news article as input together with the prompt shown in Figure~\ref{fig:visual_framing_system_prompt}, and produces the visual framing specification shown in Figure~\ref{fig:visual_framing_output_en} (original Korean output in Figure~\ref{fig:visual_framing_output_ko}). In Stage~2, the eight features from the stylistic-semiotic and denotative levels of this specification are inserted into a template-based prompt: each of the six stylistic-semiotic features---style, composition, angle, distance, saturation, and luminosity---fills its corresponding slot, and the \texttt{include\_subjects} and \texttt{exclude\_subjects} lists from the denotative level are inserted into the subject slots, yielding the T2I prompt shown in Figure~\ref{fig:text-to-image_example_en}. This prompt is then used to generate one image per article (Figure~\ref{fig:text-to-image_result}), with the API call retried if necessary until a valid image is returned. In Stage~3, the prompt in Figure~\ref{fig:multimodal_prompt} provides the target issue, news headline, article, and generated image to the LVLM, which returns a single-word stance label. 

\section{Supplementary Results}
\subsection{Performance with Publisher Images}
\label{app:sec:analysis}

\begin{table}[h]
\centering
\begin{tabular}{cccc}
\hline
\textbf{Image} & \textbf{Text} & \textbf{ACC} & \textbf{mF1} \\
\hline
\multicolumn{4}{c}{Gemini-3-flash} \\
\hline
\checkmark &            & 0.458  & 0.453 \\
           & \checkmark & 0.713 & 0.72 \\
\checkmark & \checkmark & \textbf{0.718} & \textbf{0.725} \\
\hline
\multicolumn{4}{c}{Claude-4.6-sonnet} \\
\hline
\checkmark &            & 0.437 & 0.434 \\
           & \checkmark & 0.637 & 0.63 \\
\checkmark & \checkmark & \textbf{0.668} & \textbf{0.675} \\
\hline
\multicolumn{4}{c}{GPT-5.4-mini} \\
\hline
\checkmark &            & 0.383 & 0.354 \\
           & \checkmark & 0.649 & 0.656 \\
\checkmark & \checkmark & \textbf{0.656} & \textbf{0.662} \\
\hline
\end{tabular}
\caption{Performance with publisher-provided images.}

\label{tab:preliminary_study}
\end{table}

We conducted a supplementary analysis of the effects of publisher-provided news images and examined whether their inclusion improved overall detection performance. On the test split of \textsf{K-News-Stance-MM}, we compared the zero-shot performance of the three proprietary LVLMs used in the main experiments under three input settings: (1) image only, (2) text only, and (3) text and image. These modality-controlled experiments allow us to assess the contribution of publisher images to stance detection performance.

Table~\ref{tab:preliminary_study} presents the results. When only images were used, all models performed substantially worse than in the text-only setting. GPT-5.4-mini showed the lowest image-only performance, with an accuracy of 0.383 and a macro F1 of 0.354. This finding highlights the primary role of news text in article-level stance prediction. In contrast, all models achieved their best performance when both text and image modalities were incorporated. While the improvement was most pronounced for Claude-4.6-sonnet, only modest gains were observed for Gemini-3-flash and GPT-5.4-mini. These findings suggest that publisher images can provide useful complementary signals for stance detection, while also highlighting their limited added value, potentially because stance-relevant cues in conventional news images are often implicit.

\subsection{Accuracy--Cost Trade-offs}
\label{app:sec:tradeoff}

We examined the trade-off between detection performance              and API cost by comparing \mymethod with more efficient alternatives that remain grounded in visual framing. Table~\ref{tab:app:comparison_tradeoff_perf_inf-cost} presents the accuracy and API cost of the three methods, clearly demonstrating the resulting trade-offs.

\begin{table}[t]
\centering
\resizebox{\linewidth}{!}{
\begin{tabular}{lcc}
\toprule
\textbf{Method}  & \textbf{ACC} & \textbf{API cost}\\
\midrule
\mymethod         & 0.746 $\pm$ 0.002  &  \$0.0378   \\
\textsc{\mymethod~(Text)}       & 0.73 $\pm$ 0.003   &  \$0.00392   \\
1-Step Prompting  & 0.688 $\pm$ 0.007  &  \$0.00169   \\
\bottomrule
\end{tabular}
}
\caption{Accuracy--cost trade-offs between \mymethod and its efficient alternatives grounded in visual framing. API costs are averaged per sample.}
\label{tab:app:comparison_tradeoff_perf_inf-cost}
\end{table}

\begin{table}[t]
\centering
\resizebox{\columnwidth}{!}{%
\begin{tabular}{lcc}
\toprule
\textbf{Configuration} & \textbf{ACC} & \textbf{mF1} \\
\midrule
\mymethod & \textbf{0.746 $\pm$ 0.002} & \textbf{0.747 $\pm$ 0.002} \\
\midrule
w/o compositional & 0.738   $\pm$ 0.004 & 0.74  $\pm$ 0.004  \\
w/o exclusion   & 0.736  $\pm$ 0.002 & 0.738  $\pm$ 0.003  \\
w/o style       & 0.733  $\pm$ 0.002 & 0.736  $\pm$ 0.002  \\
w/o tone       & 0.731   $\pm$ 0.002 & 0.733  $\pm$ 0.002  \\
\bottomrule
\end{tabular}%
}
\caption{Performance differences after ablating visual framing features in Stage 2.}
\label{tab:sub_feature_ablation}
\end{table}

\subsection{Visual Framing Feature Selection}
\label{app:sec:feature_ablation_result}

We conducted a feature-level ablation study to support the selection of eight features in Stage 2 for \mymethod. For the features at the stylistic-semiotic level, we grouped composition, angle, and distance into \emph{compositional} features, and saturation and luminosity into \emph{tone} features, based on their conceptual relatedness~\cite{kress-etal-1996-reading}. At the denotative level, we ablated only the exclusion feature because the inclusion feature specifies the content to be rendered. Table~\ref{tab:sub_feature_ablation} reports the results, where ``w/o $X$'' indicates \mymethod without $X$ in Stage 2. For macro F1, removing the tone features yielded the largest drop of 0.014, followed by the style feature with a drop of 0.011. Despite differences in the number and granularity of the feature groups, the results suggest that tone and style are among the most effective cues for conveying stance, likely because T2I models can render them straightforwardly and LVLMs can readily perceive them. 

\begin{table}[t]
\centering
\small
\resizebox{\columnwidth}{!}{%
\begin{tabular}{l c c}
\toprule
\textbf{Method} & \textbf{ACC} & \textbf{mF1} \\
\midrule

\multicolumn{3}{c}{English} \\
\midrule
\mymethod & \textbf{0.738 $\pm$ 0.002} & \textbf{0.74 $\pm$ 0.002} \\
Text      & 0.697 $\pm$ 0.003 & 0.704 $\pm$ 0.003 \\

\midrule
\multicolumn{3}{c}{Chinese} \\
\midrule
\mymethod & \textbf{0.735 $\pm$ 0.001} & \textbf{0.736 $\pm$ 0.001} \\
Text      & 0.7 $\pm$ 0.002 & 0.707 $\pm$ 0.002 \\

\midrule
\multicolumn{3}{c}{Indonesian} \\
\midrule
\mymethod & \textbf{0.725 $\pm$ 0.003} & \textbf{0.728 $\pm$ 0.003} \\
Text      & 0.68 $\pm$ 0.004 & 0.688 $\pm$ 0.004 \\

\midrule
\multicolumn{3}{c}{Arabic} \\
\midrule
\mymethod & \textbf{0.722 $\pm$ 0.003} & \textbf{0.724 $\pm$ 0.003} \\
Text      & 0.673 $\pm$ 0.001 & 0.68 $\pm$ 0.001 \\

\bottomrule
\end{tabular}%
}
\caption{Article-level stance detection performance on translated versions of \textsf{K-News-Stance-MM}.}
\label{tab:multilingual_supplementary}
\end{table}

\subsection{Multilingual Evaluation}
\label{app:sec:morelanguages}

We conduct a preliminary experiment to evaluate the effectiveness of \mymethod on multilingual versions of \textsf{K-News-Stance-MM} in English, Chinese, Indonesian, and Arabic. As shown in Table~\ref{tab:multilingual_supplementary}, \mymethod achieved higher accuracy and macro F1 scores than the text-only setting across all four languages, suggesting that the proposed method is effective across languages. Performance was highest in English and Chinese, followed by Indonesian and Arabic; however, the original Korean version achieved the best overall performance, with an accuracy of 0.746 and a macro F1 score of 0.747. The LLM-translated versions may lose some stance cues from the original Korean text because of their implicit nature. Future studies could further evaluate the multilingual robustness of \mymethod by constructing new resources that capture language- and culture-specific expressions of stance.

\begin{figure}[t]
\centering
\begin{subfigure}[t]{\columnwidth}
  \centering
  \includegraphics[width=0.8\columnwidth]{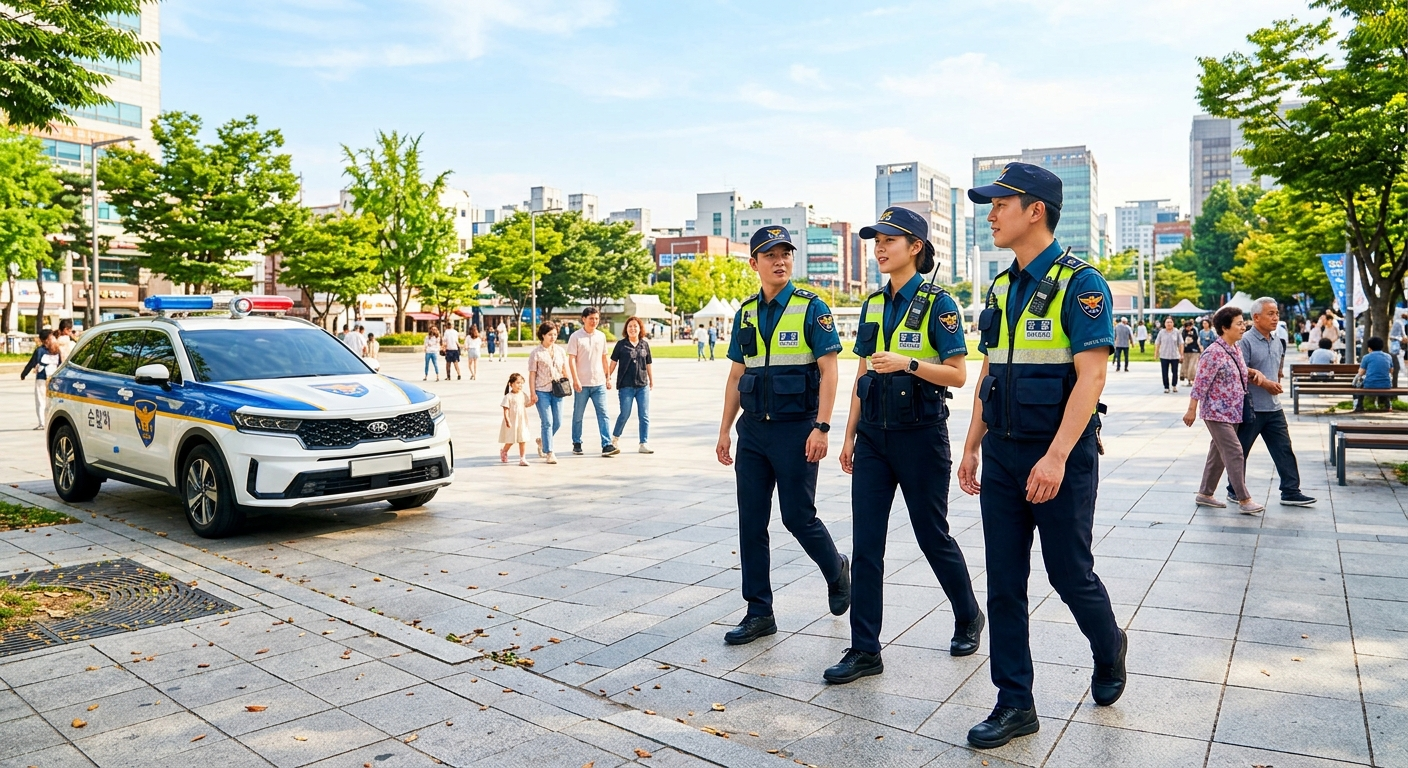}
    \caption{Misleading case}
  \label{fig:error_case_mis}
\end{subfigure}
\hfill
\begin{subfigure}[t]{\columnwidth}
  \centering
  \includegraphics[width=0.8\columnwidth]{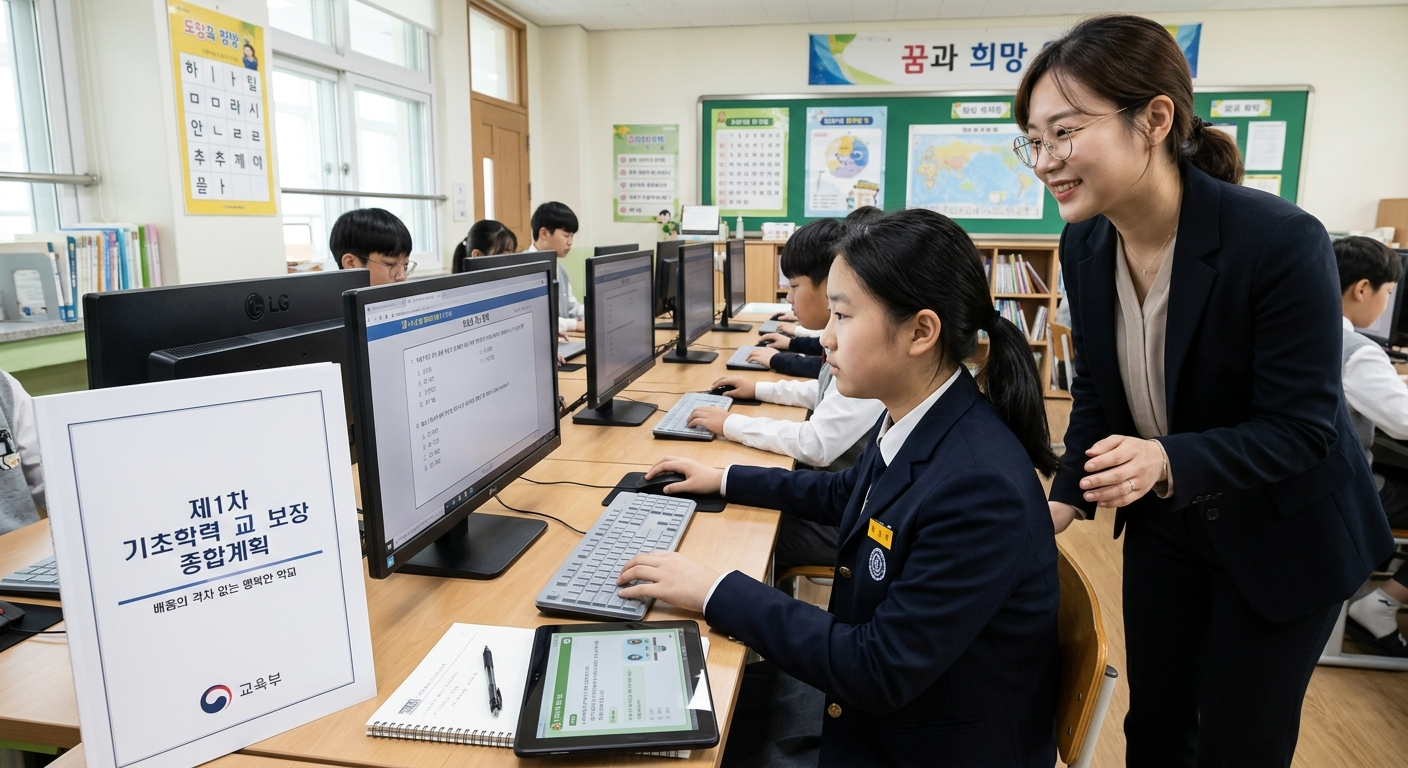}
    \caption{Beneficial case}
  \label{fig:error_case_bene}
\end{subfigure}
\caption{Representative cases where images generated by \mymethod changed stance predictions.}
\label{fig:error_cases}
\end{figure}

\begin{table*}[t]
\centering
\resizebox{\linewidth}{!}{
\begin{tabular}{lllcc}
\toprule
\textbf{Stage 1} & \textbf{Stage 2} & \textbf{Stage 3} & \textbf{ACC} & \textbf{mF1} \\
\midrule
Gemini-3-flash & Gemini-3.1-flash-image & Gemini-3-flash & \textbf{0.746 $\pm$ 0.002} & \textbf{0.747 $\pm$ 0.002} \\
Gemini-3-flash & Gemini-3.1-flash-image & InternVL3-14B-Instruct & 0.619 $\pm$ 0.001 & 0.622 $\pm$ 0.001 \\
Gemini-3-flash & Stable Diffusion 3.5 Large & Gemini-3-flash & 0.708 $\pm$ 0.004 & 0.712 $\pm$ 0.005 \\
InternVL3-14B-Instruct & Gemini-3.1-flash-image & Gemini-3-flash & 0.703 $\pm$ 0.002 & 0.706 $\pm$ 0.002 \\
InternVL3-14B-Instruct & Stable Diffusion 3.5 Large & InternVL3-14B-Instruct & 0.616 $\pm$ 0.002 & 0.619 $\pm$ 0.002 \\
\bottomrule
\end{tabular}
}
\caption{Stage-level model substitution results with open-weight models on \textsf{K-News-Stance-MM}. The first row shows the proprietary-model configuration used in the main experiments. Each subsequent row replaces the model at a single stage with an open-weight model, while the final row uses open-weight models across all stages.}
\label{tab:app:openweight_stage}
\end{table*}

\subsection{Error Case Analysis}
\label{app:sec:error}

We manually analyzed cases where images generated by \mymethod changed the stance prediction, with representative examples shown in Figure~\ref{fig:error_cases}.
First, among 55 cases where the textual LVLM baseline was correct but \mymethod was wrong, 51 had gold-neutral labels, and most errors shifted toward supportive rather than oppositional. In the example in Figure~\ref{fig:error_case_mis}, the article neutrally reports on reintroducing conscripted police, but the generated image depicts officers patrolling a bright street, and its positive tone shifted the prediction from neutral to supportive.

Second, among 57 cases where \mymethod was correct but \mymethod~(\textsc{Text}) was wrong, 48 had gold-neutral labels, with errors again skewed toward supportive. Specification terms such as \textit{bright} or \textit{dark} may act as lexical shortcuts when provided directly as text: in Figure~\ref{fig:error_case_bene}, the specification containing \textit{bright} led \mymethod~(\textsc{Text}) to a supportive misclassification, whereas rendering the same specification as a photographic image yielded a correct prediction. This suggests that perceptual cues are less susceptible to such lexical shortcuts than their textual counterparts.

\subsection{Applicability to Open-Weight Models}
\label{app:sec:openweight}
We conducted two supplementary experiments on \textsf{K-News-Stance-MM} to examine whether \mymethod remains effective when implemented with open-weight models. Since \mymethod is a modular framework, each stage can be instantiated with alternative models beyond the proprietary ones used in the main experiments. We used two open-weight LVLMs, InternVL3-14B-Instruct and Gemma3-12B-Instruct, and an open-weight T2I model, Stable Diffusion 3.5 Large. Detailed experimental settings are provided in Appendix~\ref{app:sec:experimental_setups}.

\paragraph{Open-Weight Backbone Comparison}
To verify that the performance gains of \mymethod are not limited to the tested proprietary backbones, we compared \mymethod with the corresponding textual, visual, and multimodal baselines, using each open-weight LVLM as the Stage~3 detector. Table~\ref{tab:app:openweight_backbone} shows that \mymethod outperformed all corresponding baselines under both backbones, indicating that its effectiveness extends beyond proprietary models. However, the absolute performance remained below the proprietary configuration, which may be partly attributable to the smaller sizes of the open-weight models available under our computational constraints.

\paragraph{Stage-Level Ablation}
To identify which stage is most sensitive to model substitution, we replaced the model at each stage of \mymethod with its open-weight counterpart, as listed in Table~\ref{tab:app:openweight_stage}. Substitution reduced performance at all three stages, with the largest drop observed when the Stage~3 LVLM was replaced (0.127 in accuracy and 0.125 in macro F1). The fully open-weight variant performed comparably to the Stage~3-only substitution, suggesting that Stage~3 is the primary bottleneck: effective stance detection depends particularly on the LVLM's language understanding, visual perception, and multimodal reasoning.

\begin{table}[t]
\centering
\resizebox{\columnwidth}{!}{%
\begin{tabular}{l c c}
\toprule
\textbf{Method} & \textbf{ACC} & \textbf{mF1} \\
\midrule
\multicolumn{3}{c}{\mymethod} \\
\midrule
Gemini-3-flash          & \textbf{0.746 $\pm$ 0.002} & \textbf{0.747 $\pm$ 0.002} \\
\midrule
InternVL3-14B-Instruct  & 0.619 $\pm$ 0.001 & 0.622 $\pm$ 0.001 \\
Gemma3-12B-Instruct     & 0.622 $\pm$ 0.001 & 0.604 $\pm$ 0.001 \\
\midrule
\multicolumn{3}{c}{Multimodal} \\
\midrule
InternVL3-14B-Instruct  & 0.605 $\pm$ 0.001 & 0.608 $\pm$ 0.001 \\
Gemma3-12B-Instruct     & 0.594 $\pm$ 0.001 & 0.586 $\pm$ 0.001 \\
\midrule
\multicolumn{3}{c}{Textual} \\
\midrule
InternVL3-14B-Instruct  & 0.592 $\pm$ 0.001 & 0.588 $\pm$ 0.001 \\
Gemma3-12B-Instruct     & 0.586 $\pm$ 0.002 & 0.565 $\pm$ 0.002 \\
\midrule
\multicolumn{3}{c}{Visual} \\
\midrule
Gemma3-12B-Instruct     & 0.355 $\pm$ 0.001 & 0.298 $\pm$ 0.001 \\
InternVL3-14B-Instruct  & 0.354 $\pm$ 0.002 & 0.282 $\pm$ 0.002 \\
\bottomrule
\end{tabular}%
}
\caption{Stance detection performance on \textsf{K-News-Stance-MM} with open-weight LVLMs for Stage 3, compared against the corresponding baselines under each backbone. The first row shows the proprietary-model configuration used in the main experiments.}
\label{tab:app:openweight_backbone}
\end{table}

\subsection{Generated Image Examples}
Table~\ref{tab:app:image_comparison} compares three image sources---original publisher photographs, na\"ively generated images (\emph{Direct T2I} in Table~\ref{tab:image_gen_comparison}), and \mymethod-generated images---for three Korean articles with different stances toward the same target issue.

\section{User Study Details}
This section provides additional details on the user study described in Section~\ref{sec:case_study}. Participants were balanced by gender and age: 100 female and 100 male, with 20 of each gender in each of five age groups (20--29, 30--39, 40--49, 50--59, and 60+). English translations of the nine articles are shown in Table~\ref{tab:case_study_article_sets_en}, the condition images are compared in Table~\ref{tab:case_study_images}, and the instructions and survey interface are shown in Figures~\ref{fig:case_study_instruction} and~\ref{fig:case_study_interface}. The original Korean articles are available in our GitHub repository.

\begin{figure}[h]
\begin{tcolorbox}[
  title={\textbf{Prompt -- Textual Stance Detection}},
  colbacktitle=gray!20,
  coltitle=black,
  colback=white,
  colframe=gray!50,
  fontupper=\small,
  boxrule=0.5pt
]
Based on the given article text, analyze the article text's stance toward the target issue.\\
 
The response should be in the form of a single word: \texttt{`supportive'}, \texttt{`neutral'}, or \texttt{`oppositional'}.\\
 
Target Issue: \textit{\textcolor{blue}{\{issue\}}}\\
News headline: \textit{\textcolor{blue}{\{headline\}}}\\
News article: \textit{\textcolor{blue}{\{article\}}}
\end{tcolorbox}
\centering
\vspace*{-4mm}
\caption{Textual prompt used for LVLM-based stance detection. Blue italic text highlights the input.}
\label{fig:text_only_prompt}
\end{figure}

\begin{figure}[h]
\begin{tcolorbox}[
  title={\textbf{Prompt -- Visual Stance Detection}},
  colbacktitle=gray!20,
  coltitle=black,
  colback=white,
  colframe=gray!50,
  fontupper=\small,
  boxrule=0.5pt
]
Based on the given image accompanying the article, infer the article text's stance toward the target issue.\\
 
The response should be in the form of a single word: \texttt{`supportive'}, \texttt{`neutral'}, or \texttt{`oppositional'}.\\
 
Target Issue: \textit{\textcolor{blue}{\{issue\}}}\\
\end{tcolorbox}
\centering
\vspace*{-1mm}
\caption{Visual prompt used for LVLM-based stance detection. Blue italic text highlights the input.}
\label{fig:image_only_prompt}
\end{figure}

\begin{figure}[h]
\begin{tcolorbox}[
  title={\textbf{Prompt -- Multimodal Stance Detection (Stage 3)}},
  colbacktitle=gray!20,
  coltitle=black,
  colback=white,
  colframe=gray!50,
  fontupper=\small,
  boxrule=0.5pt
]
Based on the given article text and an image accompanying this article, analyze the article text's stance toward the target issue.\\
 
The response should be in the form of a single word: \texttt{`supportive'}, \texttt{`neutral'}, or \texttt{`oppositional'}.\\
 
Target Issue: \textit{\textcolor{blue}{\{issue\}}}\\
News headline: \textit{\textcolor{blue}{\{headline\}}}\\
News article: \textit{\textcolor{blue}{\{article\}}}
\end{tcolorbox}
\centering
\vspace*{-4mm}
\caption{Multimodal prompt used for LVLM-based stance detection. Blue italic text highlights the input.}
\label{fig:multimodal_prompt}
\end{figure}

\begin{figure}[t]
\begin{tcolorbox}[
  title={\textbf{Prompt -- Stance-aware Image Generation (Stage 2)}},
  colbacktitle=gray!20,
  coltitle=black,
  colback=white,
  colframe=gray!50,
  fontupper=\small,
  boxrule=0.5pt
]
An \textit{\textcolor{blue}{illustration}}-style image.\\
Include the following: \textit{\textcolor{blue}{silhouettes of empty provincial cities, a huge Seoul-centered black hole, a ballot box with election-campaign slogans, a chaotically mixed map of Gyeonggi administrative districts, place names Gimpo, Guri, Hanam, Goyang, Bucheon, and Gwangmyeong}}.\\
Composition is \textit{\textcolor{blue}{asymmetric}}, \textit{\textcolor{blue}{high}} angle and \textit{\textcolor{blue}{long}} shot. Saturation is \textit{\textcolor{blue}{desaturated}}, luminosity is \textit{\textcolor{blue}{dark}}.\\
Exclude the following: \textit{\textcolor{blue}{Rep. Cho Kyung-tae, Lee Jae-myung, brightly smiling citizens, a glamorous Seoul nightscape}}.\\
Generate only one image at a time.
\end{tcolorbox}
\centering
\vspace*{-4mm}
\caption{The English-translated text-to-image prompt used for the T2I model in Stage~2, shown with an illustrative input. Blue italic text highlights placeholders filled using the LLM-based annotations from Stage~1; the remaining text constitutes the fixed template.}
\label{fig:text-to-image_example_en}
\end{figure}

\begin{figure}[t]
\centering
\includegraphics[width=\columnwidth]{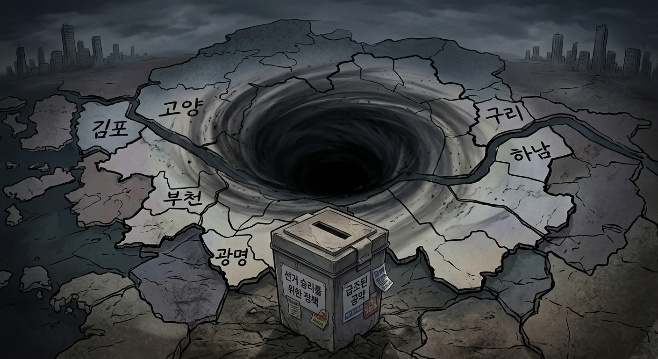}
\vspace*{-2mm}
\caption{Image generated by the T2I model in Stage~2 from the prompt in Figure~\ref{fig:text-to-image_example_en}.}
\label{fig:text-to-image_result}
\end{figure}

\begin{table*}[t]
\centering
\small
\begin{tabularx}{\textwidth}{lX}
\toprule
\textbf{Target Issue} & Democratic Party-Led Revision of the Grain Management Act Put Directly to Plenary Session for Vote \\
\midrule
\textbf{Headline} & Lee's `Bill No.~1' Grain Act likely to become Yoon's `Veto No.~1' \\
\midrule
\textbf{Article} & The revision of the Grain Management Act, which mandates the government to purchase excess rice production, passed the plenary session of the National Assembly on the 23rd, led by the Democratic Party of Korea and the Justice Party. President Yoon Suk-yeol is reportedly reviewing the option of exercising his veto power over the revision. As a result, the Grain Management Act, which Democratic Party leader Lee Jae-myung had designated as his `Bill No.~1,' is increasingly likely to become the subject of President Yoon's first-ever veto. If President Yoon does exercise his veto, it would be the first such action in approximately seven years, since former President Park Geun-hye vetoed the National Assembly Act revision centered on `standing hearings' in May 2016. The revision was approved at the plenary session with 169 votes in favor, 90 against, and 7 abstentions out of 266 members present. The core provision requires the government to purchase the entire excess production when rice output exceeds demand by 3--5\% or when rice prices fall by 5--8\% compared to the previous year. The Democratic Party pushed the Grain Management Act forward, emphasizing rice price stabilization, farmer protection, and food sovereignty. However, the government and the People Power Party opposed it on grounds of rice oversupply, fiscal burden, and weakened agricultural competitiveness, but were unable to overcome their numerical disadvantage. The government expressed deep regret over the passage of the revision. Minister of Agriculture, Food and Rural Affairs Chung Hwang-keun stated at an emergency briefing at the Government Complex Seoul that ``the side effects are all too obvious'' and declared the revision unacceptable. With the opposition's forced passage of the revision through the National Assembly, concerns about the bill's side effects are growing. Following the passage, the government is expected to purchase an additional average of approximately 200,000 tons of rice per year, with the estimated cost reaching around 1 trillion won. As demand for rice declines due to changing dietary habits, critics argue that the mandatorily purchased rice will pile up in government warehouses and ultimately be sold at a loss for purposes such as producing makgeolli. There are also significant concerns that the expansion of mandatory government purchases could lead to increased rice production and a chronic decline in rice prices. President Yoon is reportedly leaning toward exercising his veto over the revision. A senior presidential office official stated, ``Once the revision is sent to the government next week, there will be a period of public consultation and deliberation.'' \\
\midrule
\textbf{Image} & \url{https://imgnews.pstatic.net/image/005/2023/03/24/2023032321490368287_1679575743_0924293423_20230324041205219.jpg?type=w860} \\
\midrule
\textbf{Stance} & Oppositional \\
\bottomrule
\end{tabularx}
\caption{English-translated example article from \textsf{K-News-Stance-MM}. Publisher-provided original images are omitted to respect copyright; their URLs are provided instead.}
\label{tab:dataset_example_en}
\end{table*}

\begin{table*}[t]
\centering
\small
\begin{tabular}{@{} c | m{0.28\textwidth} m{0.28\textwidth} m{0.28\textwidth} @{}}
\toprule
\multicolumn{4}{@{}l}{\textbf{Issue:} Court Recognizes Same-Sex Partners as Legal Dependents for Health Insurance} \\
\multicolumn{4}{@{}l}{\textbf{이슈:} 법원 동성 부부 배우자도 건강보험 피부양자 인정} \\
\midrule
\textbf{Stance} & \centering \textbf{Original} & \centering \textbf{Na\"ive} & \centering \textbf{\mymethod} \tabularnewline
\midrule

\multicolumn{4}{@{}l}{\textbf{Headline:} Court Recognizes Same-Sex Partner's Health Insurance\ldots A Step Forward for Minority Rights} \\
\multicolumn{4}{@{}l}{\textbf{제목:} 법원, 동성반려자 건보 인정\ldots 소수자 인권 진일보} \\
\midrule
Sup. &
\centering \small{[See: \url{https://imgnews.pstatic.net/image/469/2023/02/22/0000724756_001_20230222061125823.jpg?type=w860}]} &
\centering \includegraphics[width=0.28\textwidth]{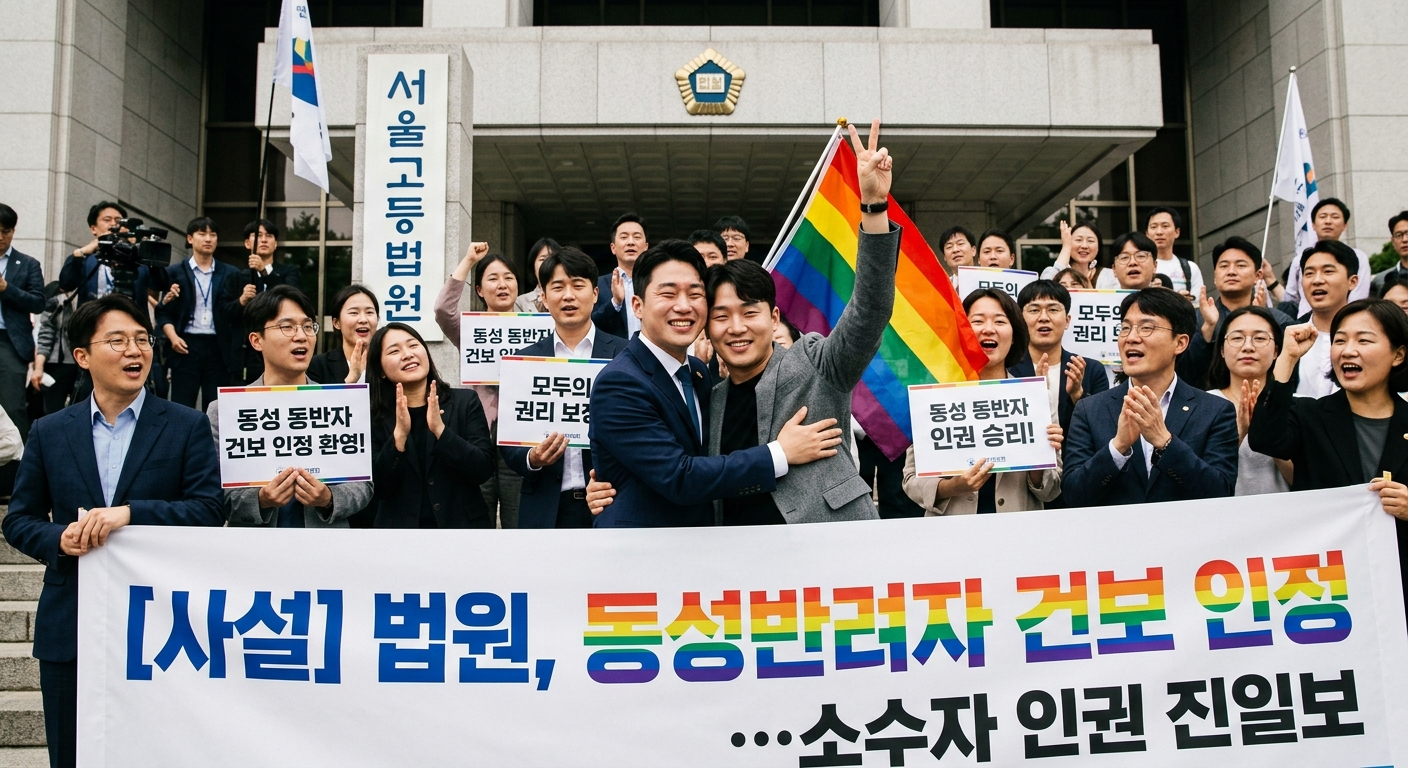} &
\centering \includegraphics[width=0.28\textwidth]{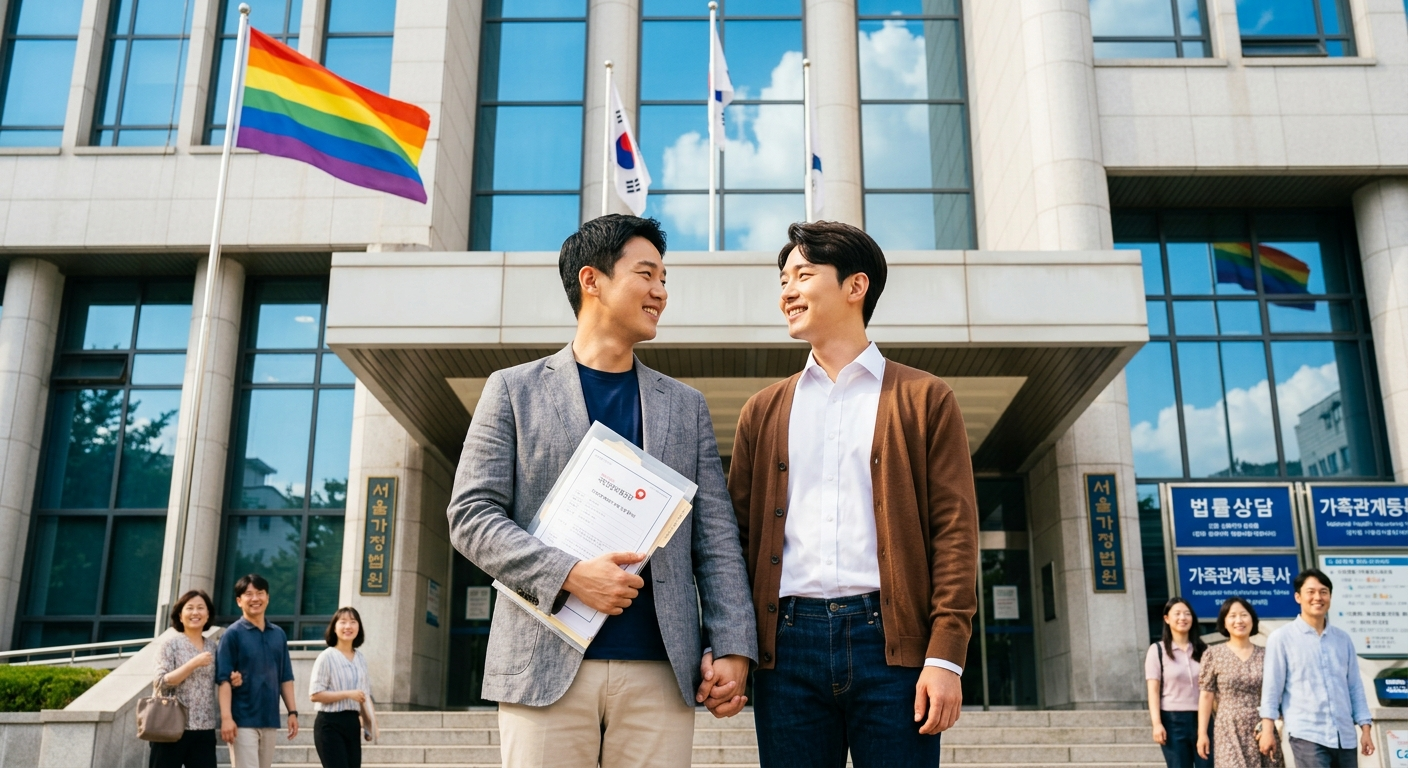} \tabularnewline[6pt]
\midrule

\multicolumn{4}{@{}l}{\parbox{\textwidth}{\textbf{Headline:} First Recognition of Same-Sex Partner as Health Insurance Dependent\ldots Court Rules ``No Discrimination \\Based on Sexual Orientation''}} \\ \addlinespace[0.3em]
\multicolumn{4}{@{}l}{\textbf{제목:} 동성커플 건보 피부양자 첫 인정\ldots 법원 ``성적지향 이유로 차별 안돼''} \\
\midrule
Neu. &
\centering \small{[See: \url{https://imgnews.pstatic.net/image/020/2023/02/22/0003481184_001_20230222031302451.jpg?type=w860}]} &
\centering \includegraphics[width=0.28\textwidth]{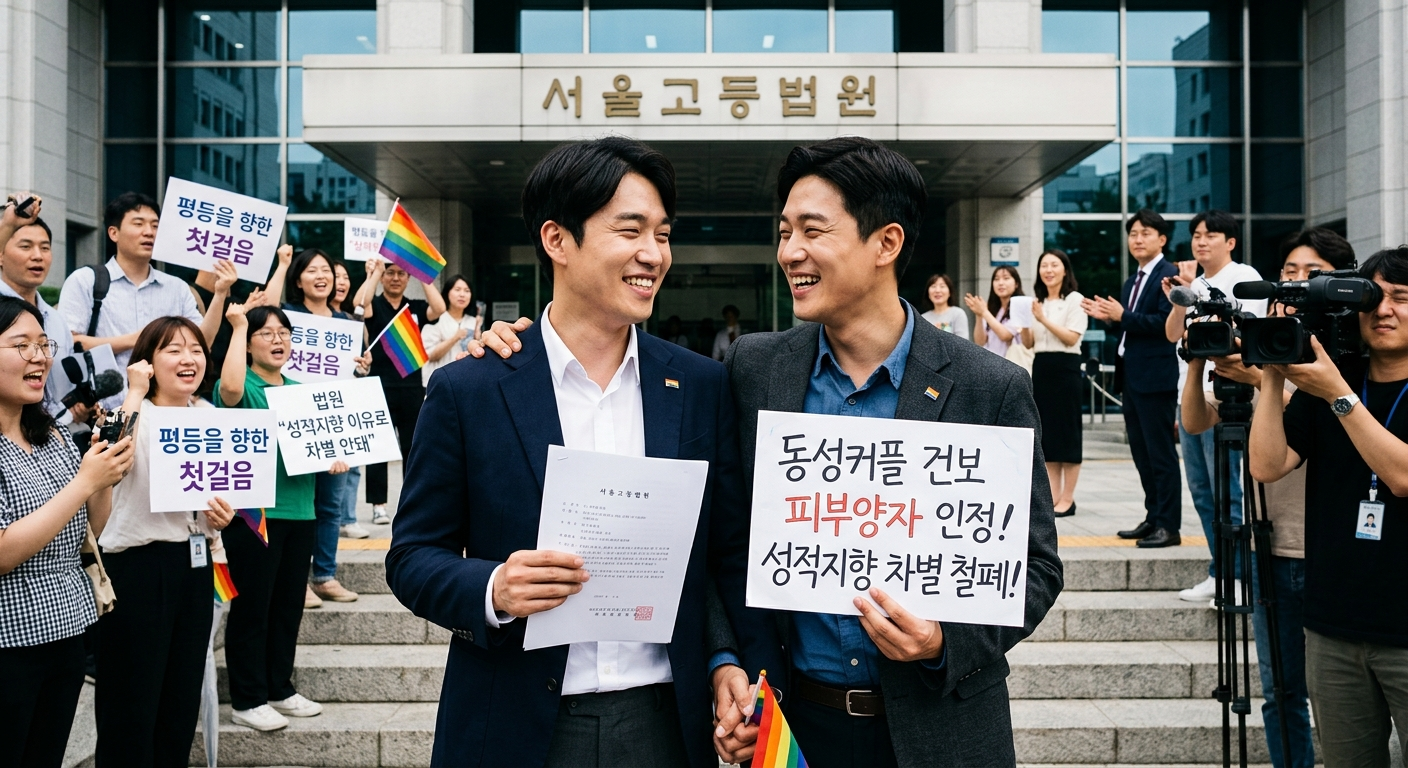} &
\centering \includegraphics[width=0.28\textwidth]{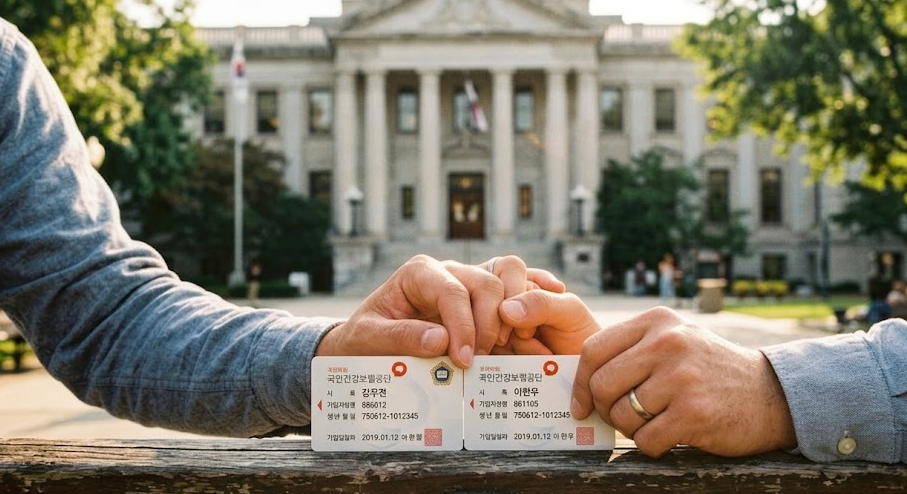} \tabularnewline[6pt]
\midrule

\multicolumn{4}{@{}l}{\textbf{Headline:} Ruling Recognizing Same-Sex Couple's Insurance Eligibility — Supreme Court Should Rectify} \\
\multicolumn{4}{@{}l}{\textbf{제목:} `동성커플 건보 자격' 인정 판결, 대법원이 바로잡으라} \\
\midrule
Opp. &
\centering \small{[See: \url{https://imgnews.pstatic.net/image/005/2023/02/22/2023022118390482487_1676972344_0924288416_20230222040306049.jpg?type=w860}]} &
\centering \includegraphics[width=0.28\textwidth]{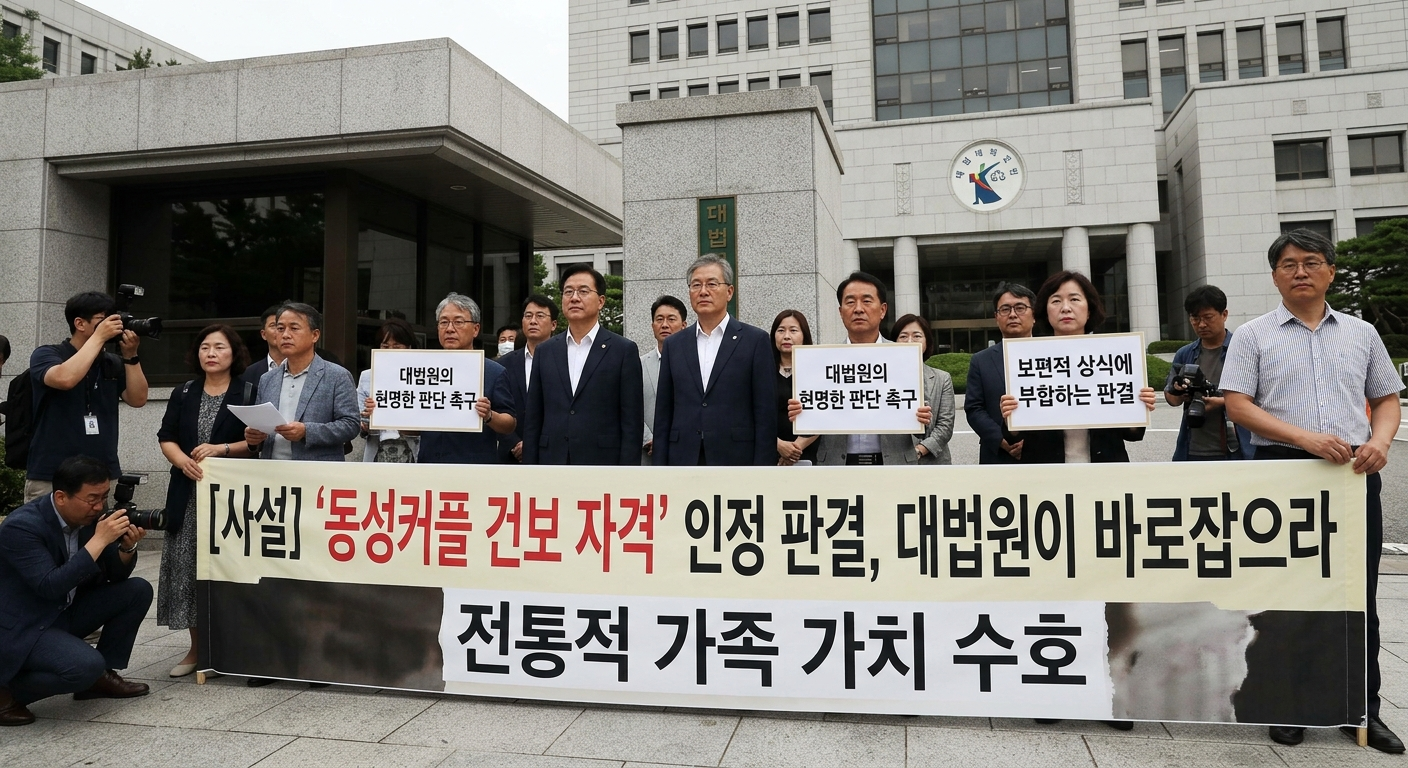} &
\centering \includegraphics[width=0.28\textwidth]{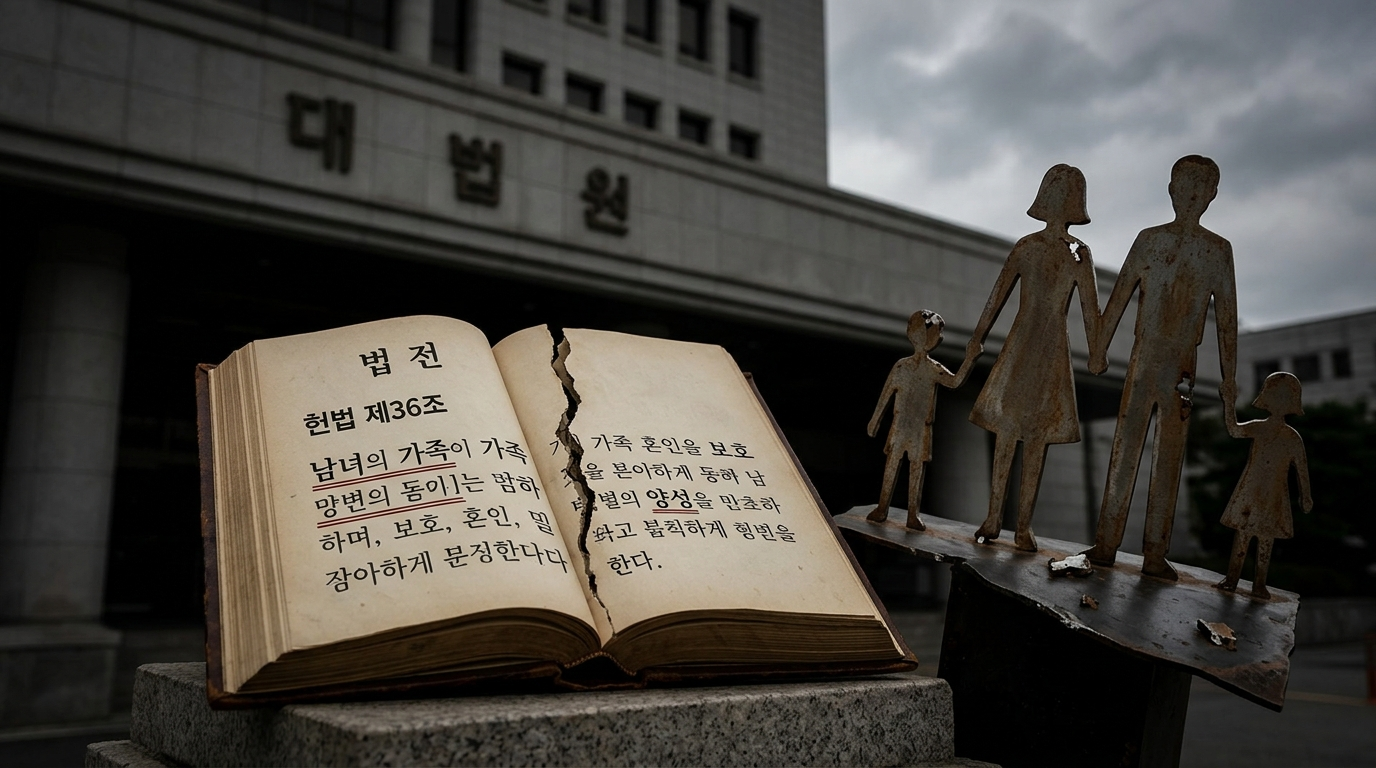} \tabularnewline
\bottomrule
\end{tabular}
\caption{Examples from three image sources: publisher-provided original images, images generated with a straightforward prompt, and \mymethod-generated images for three articles with different stances on the same issue. Original images are omitted to respect copyright; their URLs are provided instead.}\label{tab:app:image_comparison}
\end{table*}

\begin{figure*}[t]
\begin{tcolorbox}[
  title={\textbf{Prompt -- Visual Framing Annotation (Stage 1)}},
  colbacktitle=gray!20,
  coltitle=black,
  colback=white,
  colframe=gray!50,
  fontupper=\small,
  boxrule=0.5pt,
  halign=left
]
\textless role\textgreater You are a specialized assistant for generating image prompts based on Visual Framing.\textless /role\textgreater\\

\textless instructions\textgreater Design an image that visually communicates the stance a given news article takes toward a specified target issue. The stance will be one of the following: supportive, neutral, oppositional. To effectively convey the stance, annotate the Visual Framing spec according to the instructions below. The image as a whole should construct a single coherent interpretation that aligns with and effectively communicates the article's interpretation toward the issue.\\
 
\textbf{\# Visual Framing Levels}\\
\textbf{\#\# 1. Ideological level: Analyze the core axis of conflict surrounding the issue, then determine — according to the article's stance} — whose perspective and voice the image will support, whose perspective and voice it will marginalize or silence, or whether the image will instead maintain distance from the conflict axis and present the issue from a neutral/balanced viewpoint. State this in one sentence.\\
(whose interests does the image serve? whose voices are silenced?)\\
 
\textbf{\#\# 2. Connotative level: Specify the cultural or symbolic associations the image should evoke, along with the concrete visual elements that will evoke them, using the format ``A $\rightarrow$ B''.}\\
- A: the concept, value, emotion, or idea to be conveyed.\\
- B: a concrete, visually representable object that would evoke A within the socio-cultural context of the country in which the article was written.\\
 
\textbf{\#\# 3. Stylistic-Semiotic level: Determine the stylistic conventions and technical transformations involved in conveying the stance.}\\
- style: [photo, illustration] signals factual versus interpretive presentation.\\
- composition: [centered, split, asymmetric, crowded] structures the spatial arrangement of visual elements.\\
- angle: [low, eye-level, high] conveys power relations through viewpoint.\\
- distance: [close-up, medium, long] controls social distance from the subject.\\
- saturation: [saturated, neutral, desaturated] sets image tone through color intensity.\\
- luminosity: [bright, neutral, dark] sets image tone.\\
 
\textbf{\#\# 4. Denotative level: At the surface level of the image, decide what should be intentionally included and what should be intentionally excluded. For real individuals or groups, include the exact names mentioned in the article.}\\
- include\_subjects: specific subjects intentionally included.\\
- exclude\_subjects: specific subjects intentionally excluded.
\textless /instructions\textgreater\\
 
\textless constraints\textgreater\\
- All outputs must be written in Korean.\\
- Every B in the Connotative level must also appear in Denotative subjects.\\
- Keep the decision space of each level separate.\\
\textless /constraints\textgreater\\

\textless output\_format\textgreater Respond only with a JSON object. No preface or explanation.\\
\{\\
\hspace*{4mm}``ideological'': ``'',\\
\hspace*{4mm}``connotative'': [``A $\rightarrow$ B'', ...],\\
\hspace*{4mm}``stylistic\_semiotic'': \{\\
\hspace*{8mm}``style'': ``photo $|$ illustration'', ``composition'': ``centered $|$ split $|$ asymmetric $|$ crowded'', ``angle'': ``low $|$ \hspace*{8mm} eye-level $|$ high'', ``distance'': ``close-up $|$ medium $|$ long'', ``saturation'': ``saturated $|$ neutral $|$ desaturated'', \hspace*{8mm}``luminosity'': ``bright $|$ neutral $|$ dark''\\
\hspace*{4mm}\},\\
\hspace*{4mm}``denotative'': \{``include\_subjects'': [``'', ...], ``exclude\_subjects'': [``'', ...]\}\\
\}
\textless /output\_format\textgreater

\textless context\textgreater\\
- Target Issue: \textit{\textcolor{blue}{People Power Party Launches Special Committee for Gimpo-Seoul Integration}}\\
- News Headline: \textit{\textcolor{blue}{All of them to Seoul? ``Five Gyeonggi cities besides Gimpo also considered for incorporation''}}\\
- News Article (excerpt): \textit{\textcolor{blue}{Cho Kyung-tae: ``Guri, Hanam, Goyang, Bucheon, and Gwangmyeong under review \ldots at least five, and hopefully more, should join.'' The People Power Party's ``Mega Seoul'' plan, including Gimpo's incorporation into Seoul, is reportedly also considering five other cities in Gyeonggi Province, raising concerns over confusion across the metropolitan area. The People Power Party officially declared that it would pursue Seoul incorporation not only for Gimpo but also for cities across the Seoul living sphere. Rep. Cho Kyung-tae, chair of the party's Special Committee for Improving Resident Convenience in the Metropolitan Area (tentative name), told Channel A on the 2nd, ``Beyond Gimpo, I hope at least five places, including Guri, Hanam, Goyang, Bucheon, and Gwangmyeong, will join Seoul.'' \ldots}}\\
\textless /context\textgreater
\end{tcolorbox}
\centering
\vspace*{-4mm}
\caption{English-translated system and user prompts used in Stage~1 for visual framing annotation in \mymethod, shown with an illustrative input. Blue italic text highlights the input.}
\label{fig:visual_framing_system_prompt}
\end{figure*}

\begin{figure*}[t]
\begin{tcolorbox}[
  title={\textbf{Stage~1 Output -- Visual Framing Specifications (Korean)}},
  colbacktitle=gray!20,
  coltitle=black,
  colback=white,
  colframe=gray!50,
  fontupper=\small,
  boxrule=0.5pt
]
\begin{Verbatim}[
  breaklines=true,
  breaksymbolleft={}
]
{
  "ideological": "수도권 집중화와 선거용 선심성 정책이 초래할 국가적 불균형과 혼란을 우려하는 관점을 옹호하며, 무분별한 서울 확장의 부정적 파급력을 부각한다.",
  "connotative": [
    "수도권 집중으로 인한 지역 소멸의 위기감 → 텅 빈 지방 도시들의 실루엣",
    "계획 없는 무분별한 서울 확장 시도 → 거대한 서울 중심의 블랙홀",
    "선거 승리만을 목적으로 한 급조된 정책 → 선거용 홍보 문구가 적힌 투표함",
    "경기도 주요 도시들의 행정적 혼란 → 어지럽게 뒤섞인 경기도 행정 구역 지도"
  ],
  "stylistic_semiotic": {
    "style": "illustration", "composition": "asymmetric", "angle": "high", "distance": "long", "saturation": "desaturated", "luminosity": "dark"
  },
  "denotative": {
    "include_subjects": [
      "텅 빈 지방 도시들의 실루엣", "거대한 서울 중심의 블랙홀", "선거용 홍보 문구가 적힌 투표함", "어지럽게 뒤섞인 경기도 행정 구역 지도", "김포, 구리, 하남, 고양, 부천, 광명 지명"
    ],
    "exclude_subjects": [
      "조경태 의원", "이재명 대표", "밝게 웃는 시민들", "화려한 서울의 야경"
    ]
  }
}
\end{Verbatim}
\end{tcolorbox}
\centering
\vspace*{-4mm}
\caption{Original Korean-language output of Stage~1 visual framing annotation in \mymethod.}
\label{fig:visual_framing_output_ko}
\end{figure*}

\begin{figure*}[t]
\begin{tcolorbox}[
  title={\textbf{Stage~1 Output -- Visual Framing Specifications (English)}},
  colbacktitle=gray!20,
  coltitle=black,
  colback=white,
  colframe=gray!50,
  fontupper=\small,
  boxrule=0.5pt
]
\begin{Verbatim}[
  breaklines=true,
  breaksymbolleft={}
]
{
  "ideological": "Advocates a perspective concerned about the national imbalance and confusion that may result from metropolitan concentration and election-oriented populist policies, while highlighting the negative ripple effects of indiscriminate Seoul expansion.",
  "connotative": [
    "A sense of crisis over regional extinction caused by metropolitan concentration → silhouettes of empty provincial cities",
    "An unplanned attempt at indiscriminate Seoul expansion → a huge Seoul-centered black hole",
    "A hastily assembled policy aimed only at electoral victory → a ballot box with election-campaign slogans",
    "Administrative confusion among major Gyeonggi cities → a chaotically mixed map of Gyeonggi administrative districts"
  ],
  "stylistic_semiotic": {
    "style": "illustration", "composition": "asymmetric", "angle": "high", "distance": "long", "saturation": "desaturated", "luminosity": "dark"
  },
  "denotative": {
    "include_subjects": [
      "silhouettes of empty provincial cities", "a huge Seoul-centered black hole", "a ballot box with election-campaign slogans", "a chaotically mixed map of Gyeonggi administrative districts", "place names Gimpo, Guri, Hanam, Goyang, Bucheon, and Gwangmyeong"
    ],
    "exclude_subjects": [
      "Rep. Cho Kyung-tae", "Lee Jae-myung", "brightly smiling citizens", "a glamorous Seoul nightscape"
    ]
  }
}
\end{Verbatim}
\end{tcolorbox}
\centering
\vspace*{-4mm}
\caption{English-translated output of Stage~1 visual framing annotation in \mymethod.}
\label{fig:visual_framing_output_en}
\end{figure*}

\begin{table*}[h]
  \centering
  \small
  \begin{tabularx}{\textwidth}{@{} c c X @{}}
    \toprule
    \textbf{ID} & \textbf{Stance} & \textbf{Headline \& Lead} \\
    \midrule
    \multicolumn{3}{@{}l}{\textbf{Issue 1.} Military to Resume Full-Scale Propaganda Broadcasts Following Repeated Trash Balloon Incidents} \\
    \midrule
    A1 & Sup. &
      \textbf{JCS to ``fully implement propaganda broadcasts to the North''\ldots\ countering N.\ Korean trash balloons}
      \newline In response to North Korea's trash balloon launches, the military authorities have played the card of fully implementing propaganda broadcasts to the North. A hardline tit-for-tat standoff between the two Koreas through psychological warfare appears to be deepening. \\
    A2 & Neu. &
      \textbf{N.\ Korea launches 9th round of trash balloons\ldots\ Military counters with ``full-scale propaganda broadcasts to the North''}
      \newline After North Korea once again released trash balloons toward the South on the morning of the 21st, the military authorities responded with the ``full-scale implementation of propaganda broadcasts to the North,'' and military tensions between the two Koreas are escalating. \\
    A3 & Opp. &
      \textbf{How will the North respond to the full expansion of propaganda broadcasts?\ldots\ Border tensions mount}
      \newline Using domestic civic groups' anti-North leaflet drops as a pretext, North Korea continues to launch trash balloons, and our military authorities have repeatedly countered with propaganda broadcasts to the North. \\
    \midrule
    \multicolumn{3}{@{}l}{\textbf{Issue 2.} Impeachment Motion Against the BAI Chairman Faces Vote Tomorrow} \\
    \midrule
    B1 & Sup. &
      \textbf{BAI, unable to question Kim Keon-hee about ``21Gram,'' protests: ``We can't be expected to torture it out of them''}
      \newline The Board of Audit and Inspection (BAI), after failing to identify who recommended ``21Gram''---the contractor awarded the presidential residence renovation in Hannam-dong, Seoul---protested that ``it isn't something we can uncover by torturing people.'' \\
    B2 & Neu. &
      \textbf{Impeachment motions against the BAI Chairman and prosecutors reported to the plenary session\ldots\ Budget bill put on hold}
      \newline Speaker Woo urged the ruling and opposition parties to reach a budget agreement by the 10th. The impeachment motions against the BAI Chairman and prosecutors will be voted on on the 4th, creating a vacuum in the chain of command. The year-end political situation is becoming increasingly unpredictable. \\
    B3 & Opp. &
      \textbf{BAI: ``There's no Plan B at this stage\ldots\ We trust the impeachment will be withdrawn''}
      \newline As the impeachment motion against BAI Chairman Choe Jae-hae, filed by the Democratic Party of Korea, was reported to the National Assembly's plenary session on the 2nd, the Board of Audit and Inspection stated that it ``is not considering a Plan B at this stage'' in preparation for the aftermath of the impeachment. \\
    \midrule
    \multicolumn{3}{@{}l}{\textbf{Issue 3.} President Yoon Nominates Vice Justice Minister Shim Woo-jung for Prosecutor General} \\
    \midrule
    C1 & Sup. &
      \textbf{Prosecutor General nominee Shim Woo-jung, a ``planning specialist'': ``I will do my utmost to earn the public's trust''}
      \newline On the 11th, President Yoon Suk-yeol nominated Vice Justice Minister Shim Woo-jung (53, Judicial Research and Training Institute Class 26) as the next Prosecutor General candidate. \\
    C2 & Neu. &
      \textbf{Prosecutor General nominee on the ``Kim Keon-hee handbag'' allegations: ``Law and principle are what matter''}
      \newline Prosecutor General nominee Shim Woo-jung took a principled stance on the ``luxury handbag'' allegations involving First Lady Kim Keon-hee, former head of Covana Contents, stating that ``it is important to uphold the law and principle.'' \\
    C3 & Opp. &
      \textbf{Emphasis on ``tighter control of the prosecution''\ldots\ Yongsan's ``safe choice''}
      \newline On the 11th, President Yoon Suk-yeol nominated Vice Justice Minister Shim Woo-jung as the next head of the prosecution. Commentators view this as a ``safe choice'' by Yoon---who has been unable to shake off various legal risks---made with an eye to the relationship between the presidential office in Yongsan and the prosecution, and to the stability of the prosecutorial organization. \\
    \bottomrule
  \end{tabularx}
  \caption{English translation of articles used in the user study. Each cell shows the headline (bold) and lead of the corresponding article. The original Korean articles are available in our GitHub repository.}
  \label{tab:case_study_article_sets_en}
\end{table*}

\begin{table*}[t]
\centering
\small
\begin{tabular}{@{} c c m{0.24\textwidth} m{0.24\textwidth} m{0.24\textwidth} @{}}
\toprule
\textbf{ID} & \textbf{Stance} & \centering \textbf{Original} & \centering \textbf{Na\"ive} & \centering \textbf{Proposed} \tabularnewline
\midrule
A1 & Sup. &
\centering \small{[See: \url{https://imgnews.pstatic.net/image/022/2024/07/21/20240721505503_20240721142308421.jpg?type=w860}]} &
\centering \includegraphics[width=0.18\textwidth]{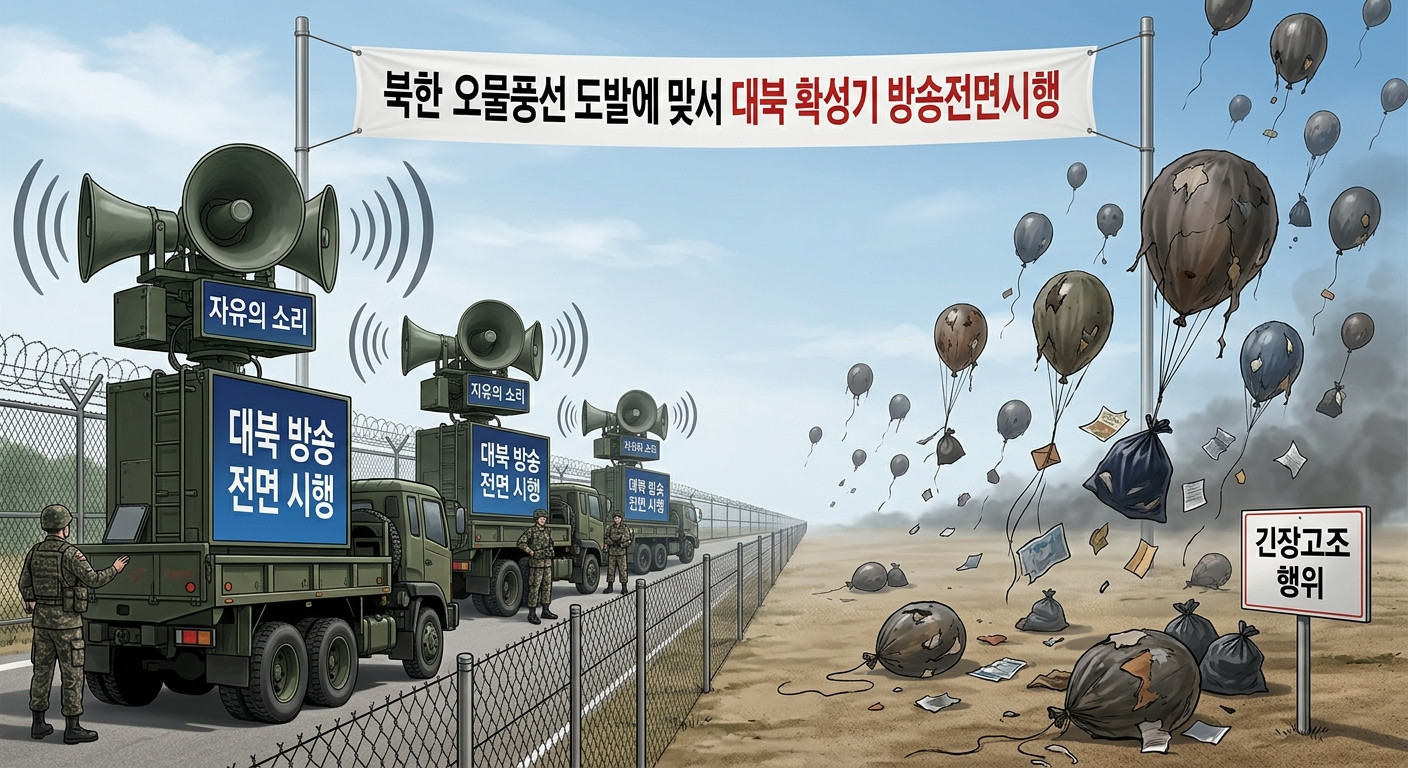} &
\centering \includegraphics[width=0.18\textwidth]{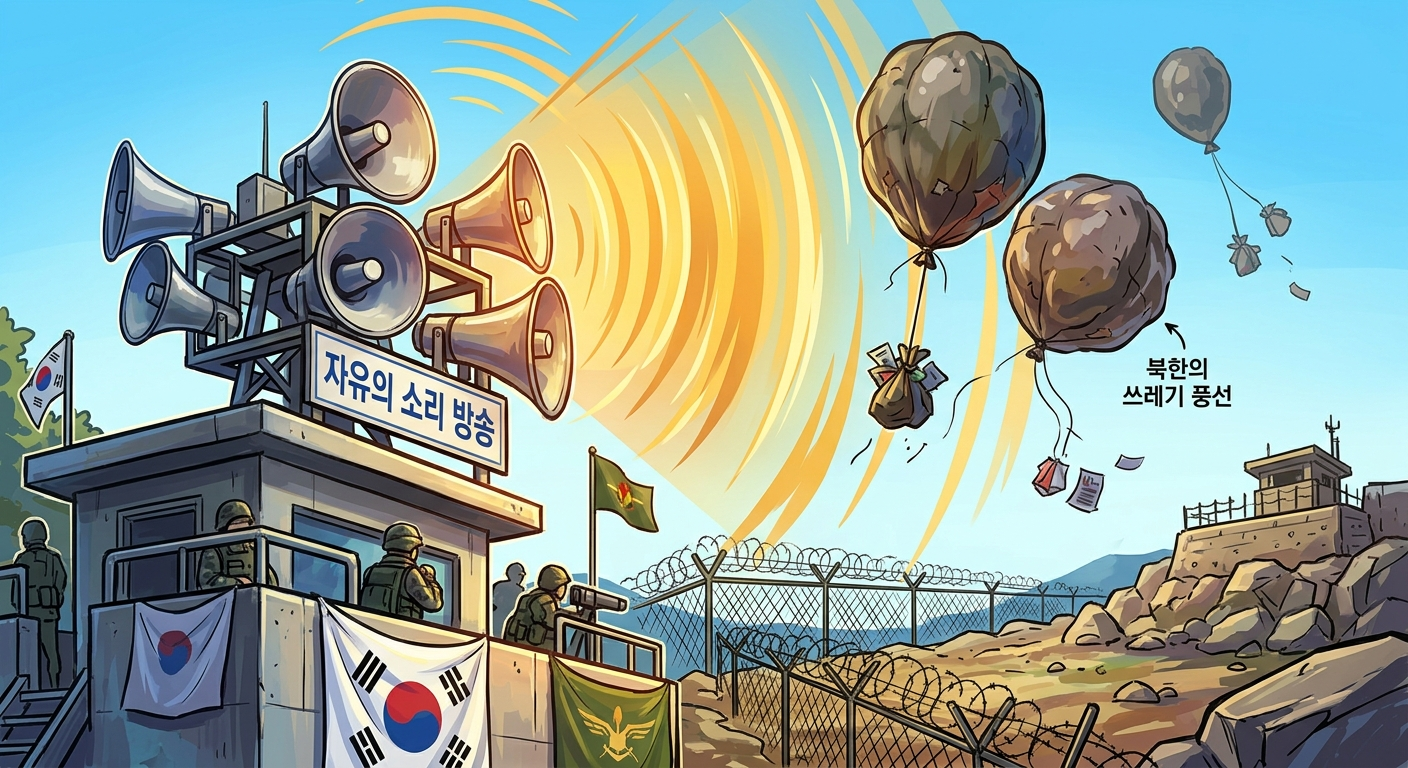} \tabularnewline
A2 & Neu. &
\centering \small{[See: \url{https://imgnews.pstatic.net/image/009/2024/07/21/0005337843_001_20240721130306464.jpg?type=w860}]} &
\centering \includegraphics[width=0.18\textwidth]{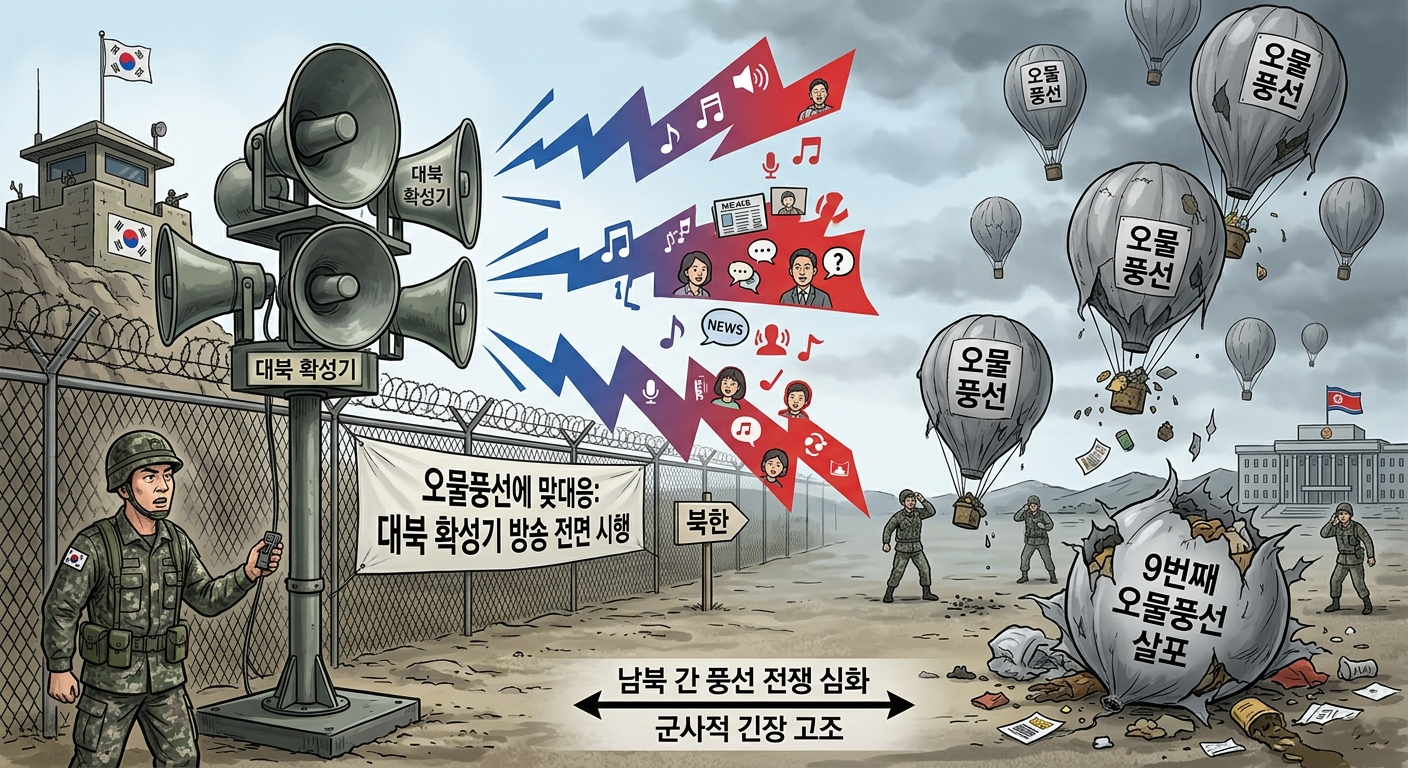} &
\centering \includegraphics[width=0.18\textwidth]{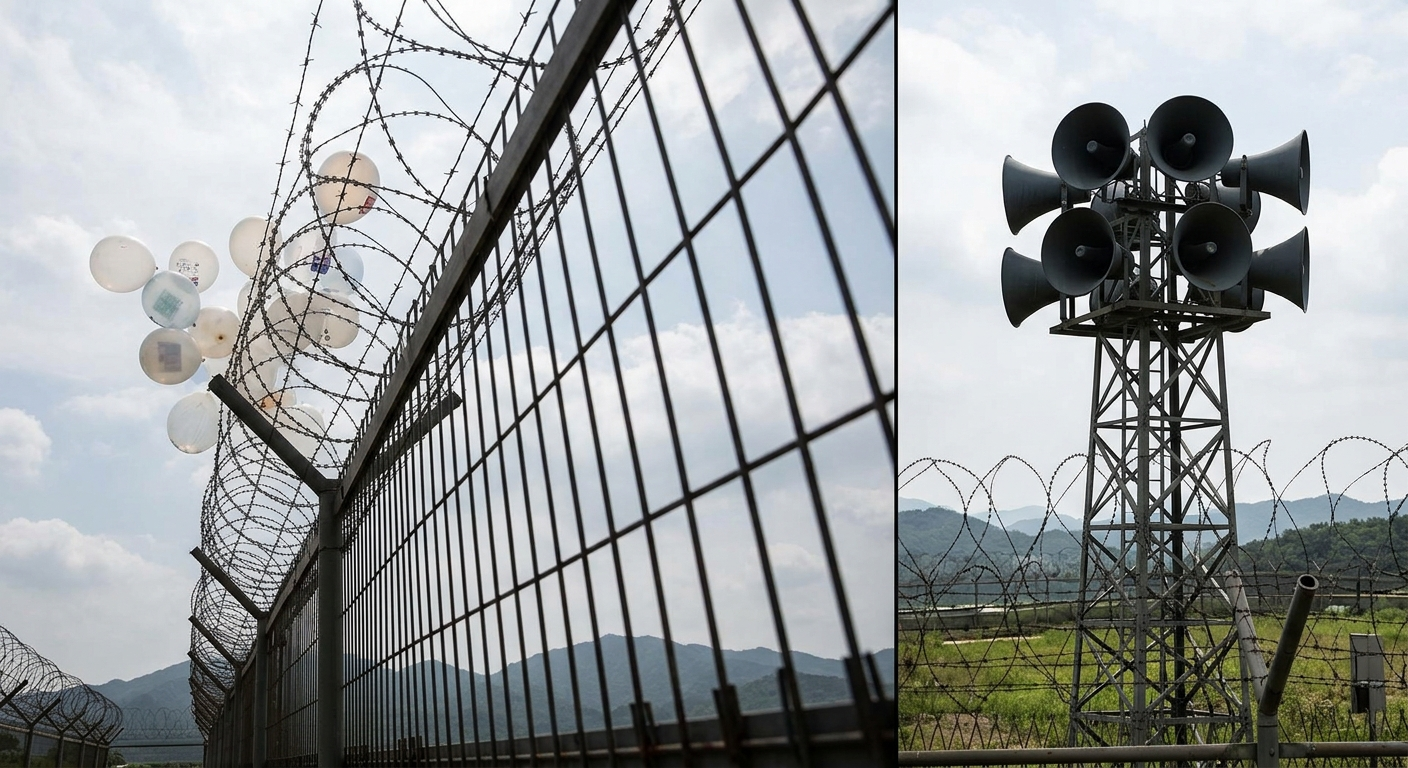} \tabularnewline
A3 & Opp. &
\centering \small{[See: \url{https://imgnews.pstatic.net/image/079/2024/07/21/0003918478_001_20240721161010430.jpg?type=w860}]} &
\centering \includegraphics[width=0.18\textwidth]{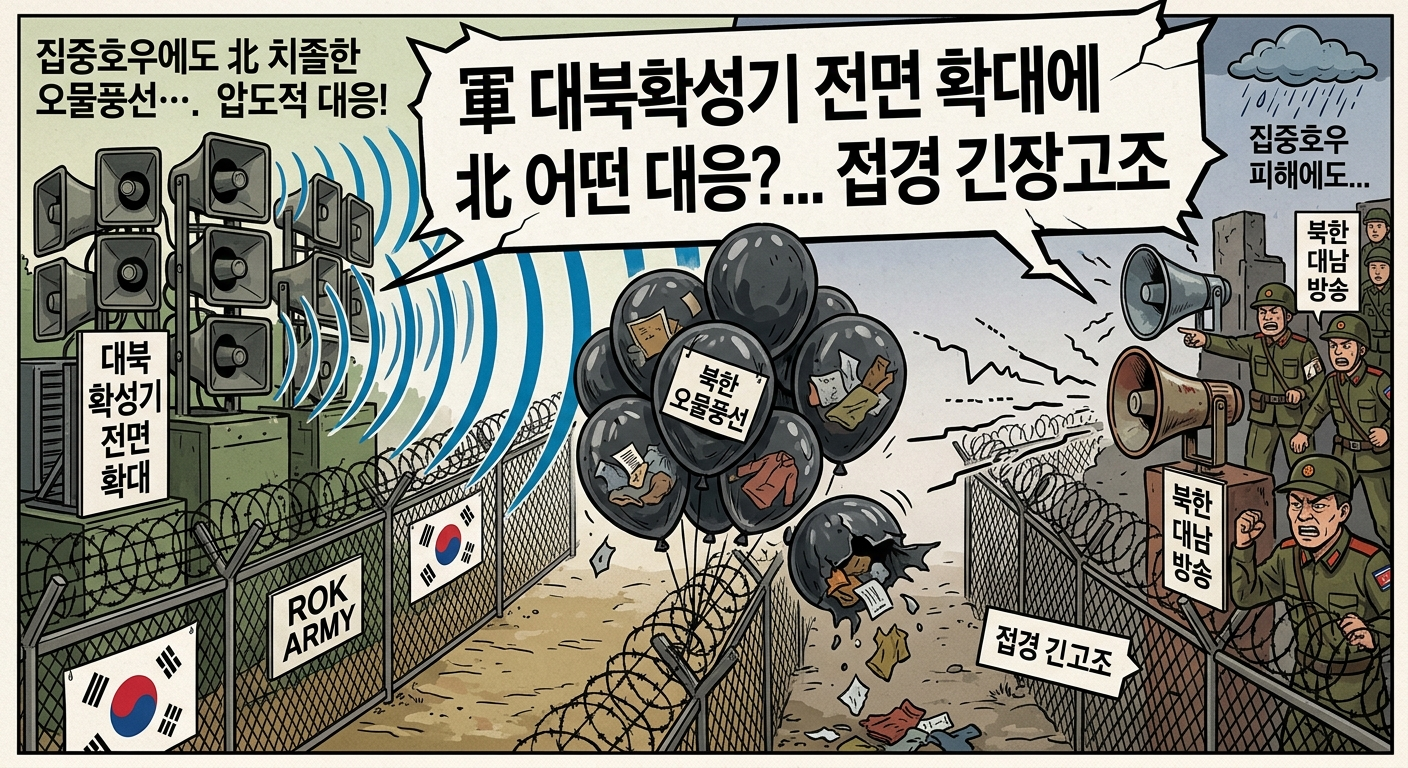} &
\centering \includegraphics[width=0.18\textwidth]{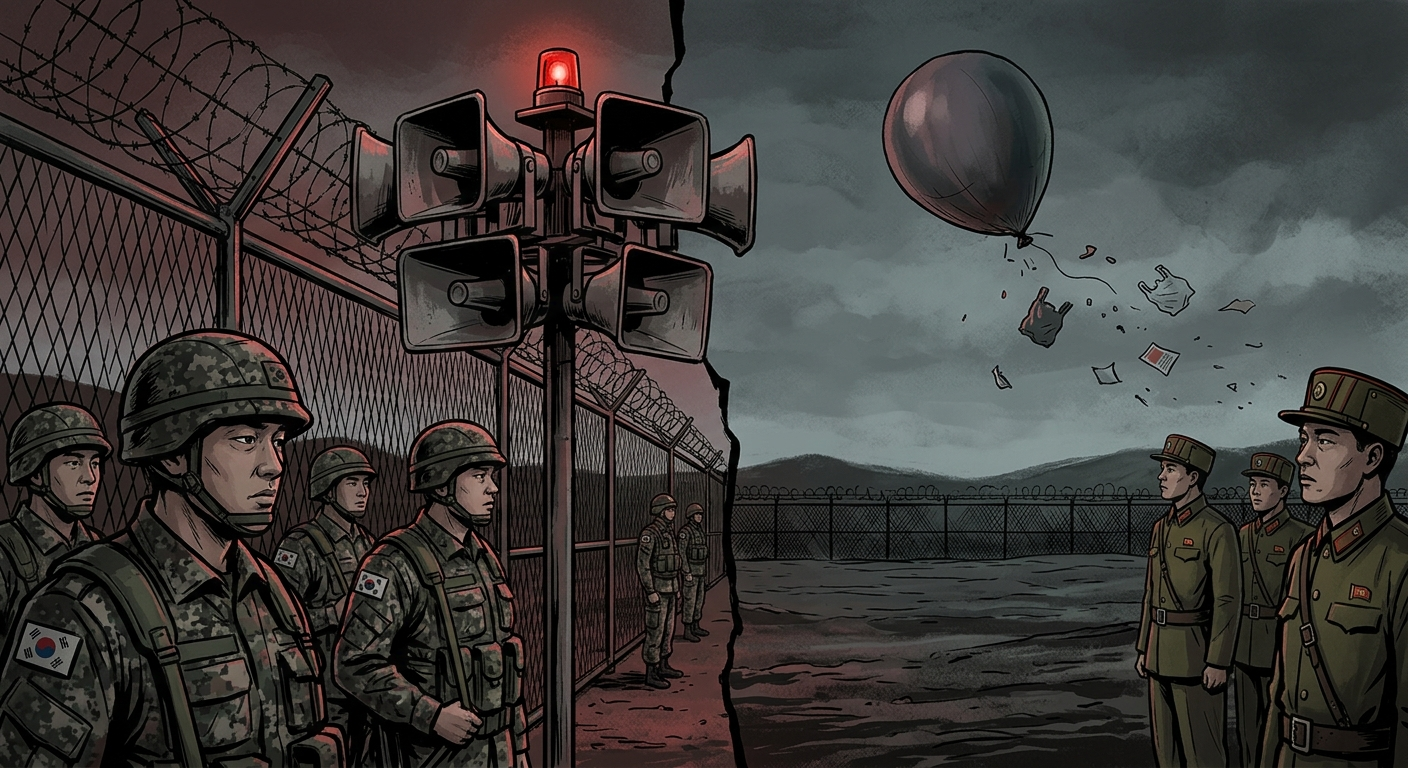} \tabularnewline
\midrule
B1 & Sup. &
\centering \small{[See: \url{https://imgnews.pstatic.net/image/028/2024/12/02/0002719024_001_20241203153828650.jpg?type=w860}]} &
\centering \includegraphics[width=0.18\textwidth]{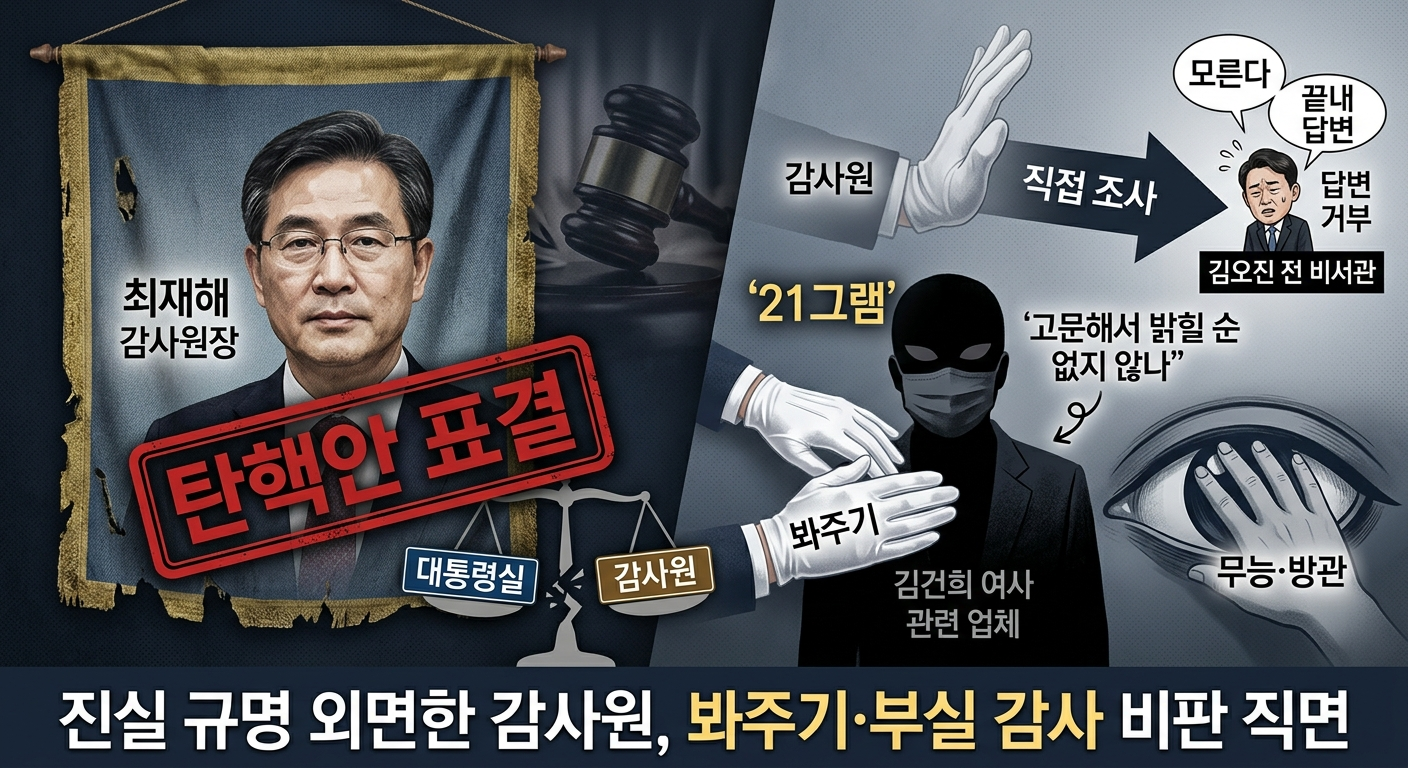} &
\centering \includegraphics[width=0.18\textwidth]{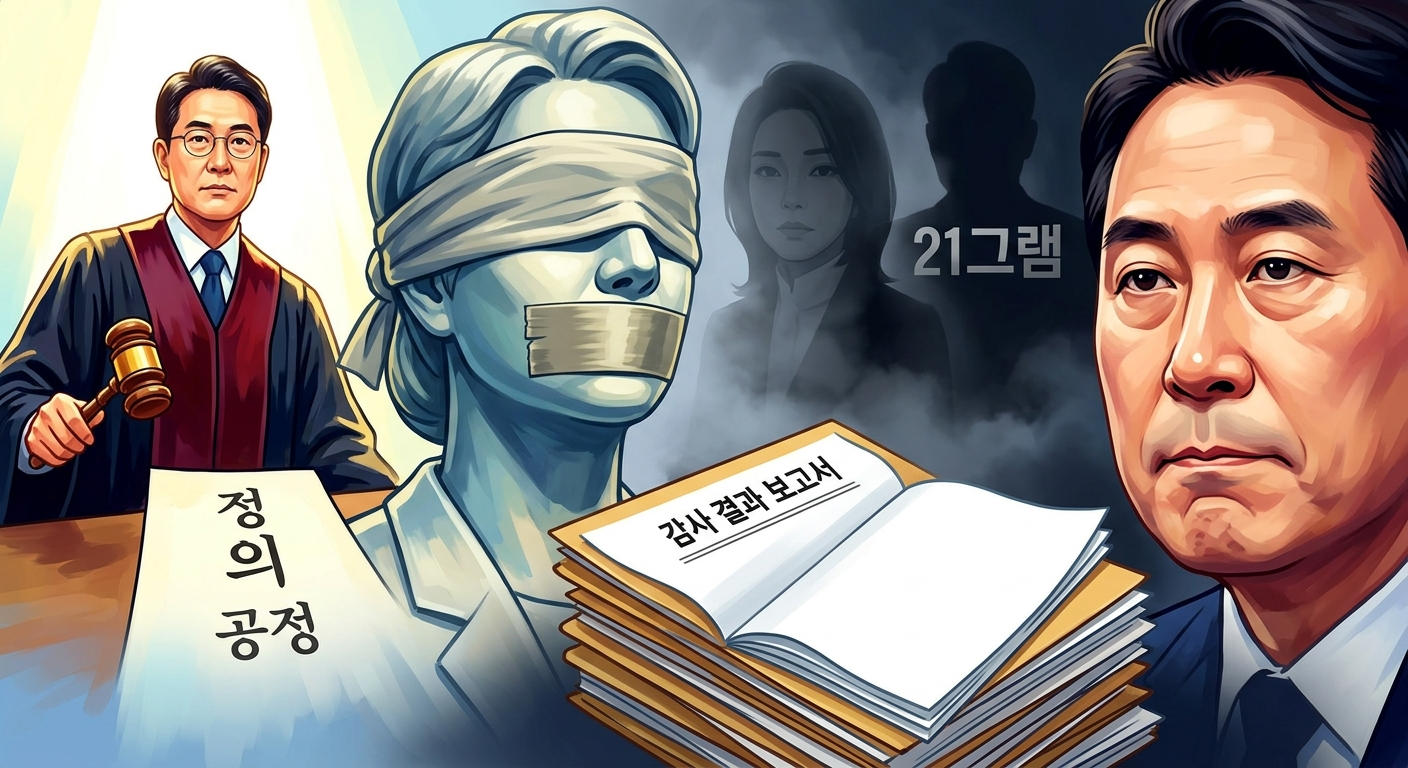} \tabularnewline
B2 & Neu. &
\centering \small{[See: \url{https://imgnews.pstatic.net/image/656/2024/12/02/0000113206_001_20241202185211320.jpg?type=w860}]} &
\centering \includegraphics[width=0.18\textwidth]{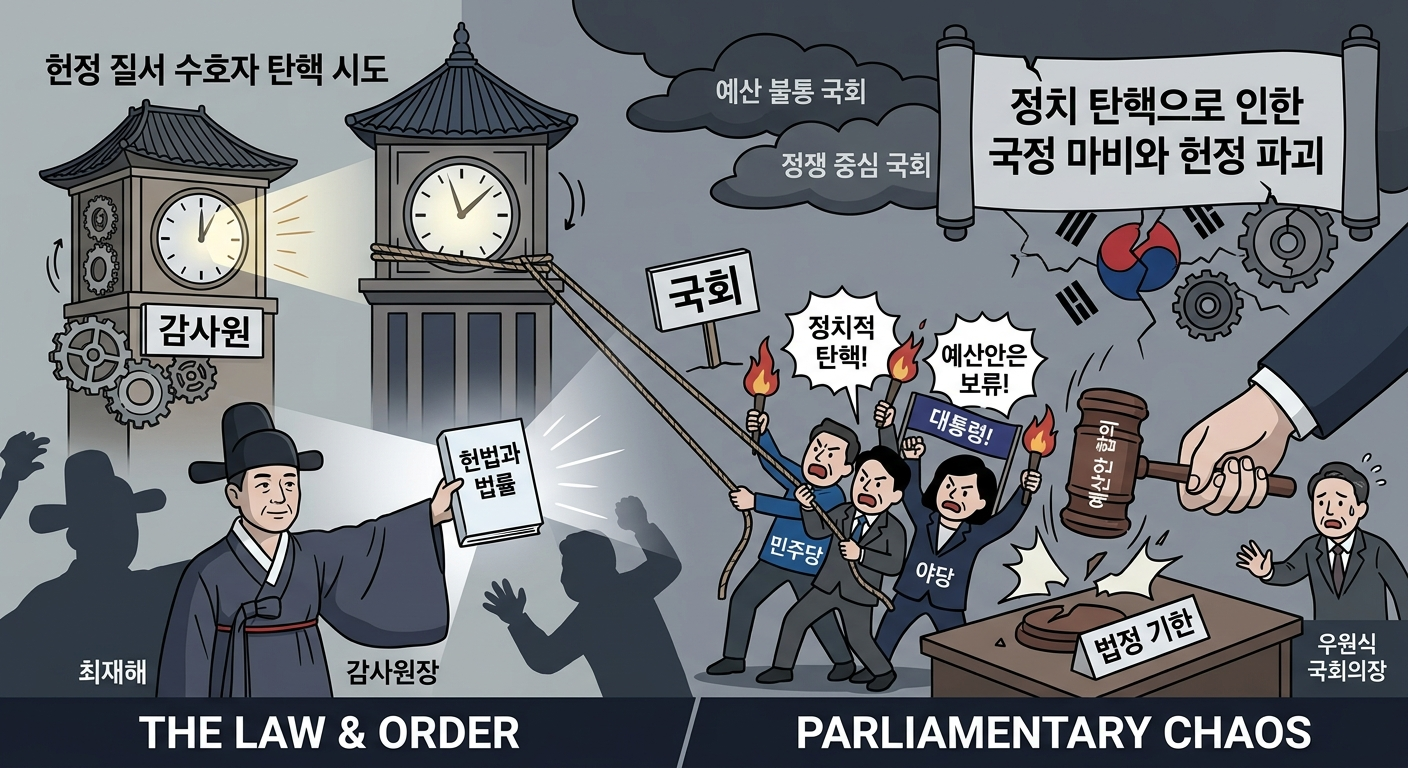} &
\centering \includegraphics[width=0.18\textwidth]{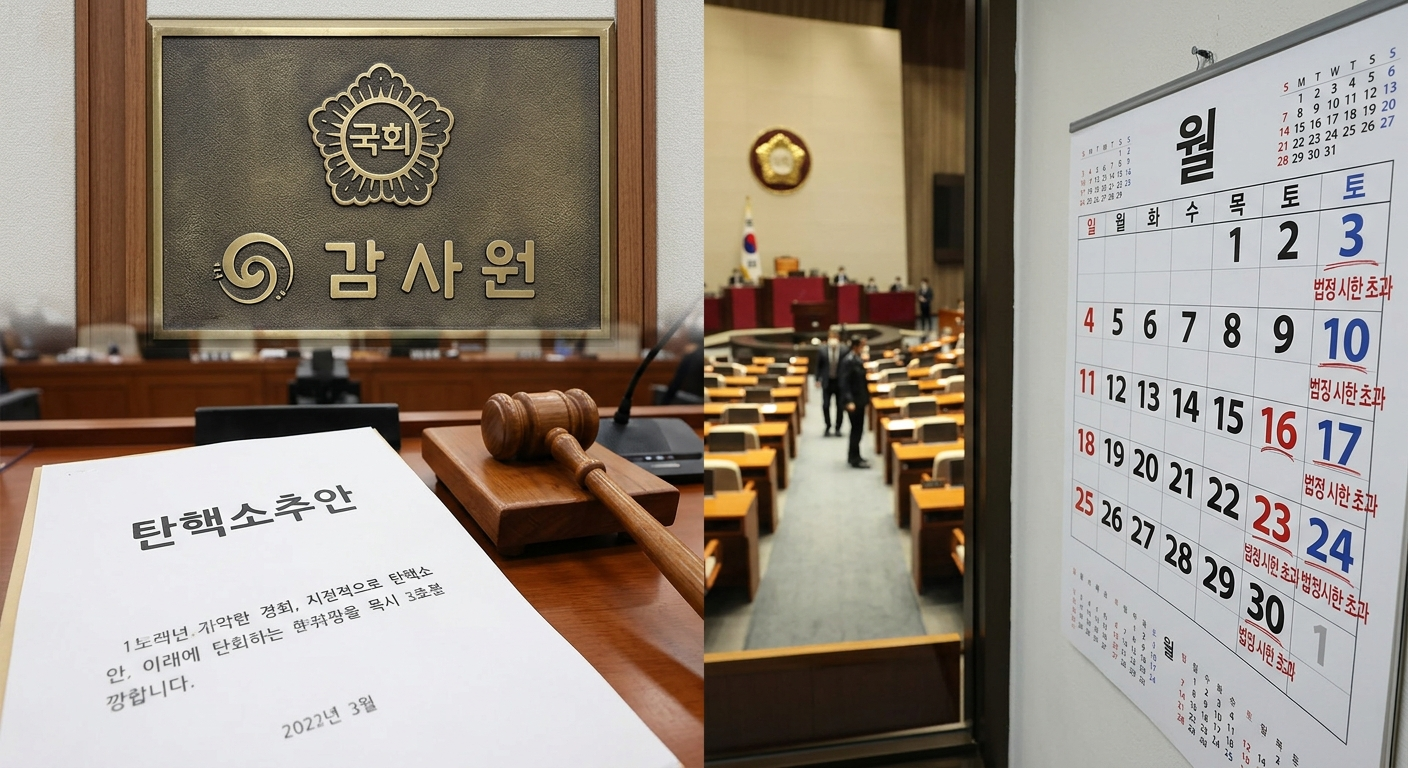} \tabularnewline
B3 & Opp. &
\centering \small{[See: \url{https://imgnews.pstatic.net/image/079/2024/12/02/0003965096_001_20241202160219972.jpg?type=w860}]} &
\centering \includegraphics[width=0.18\textwidth]{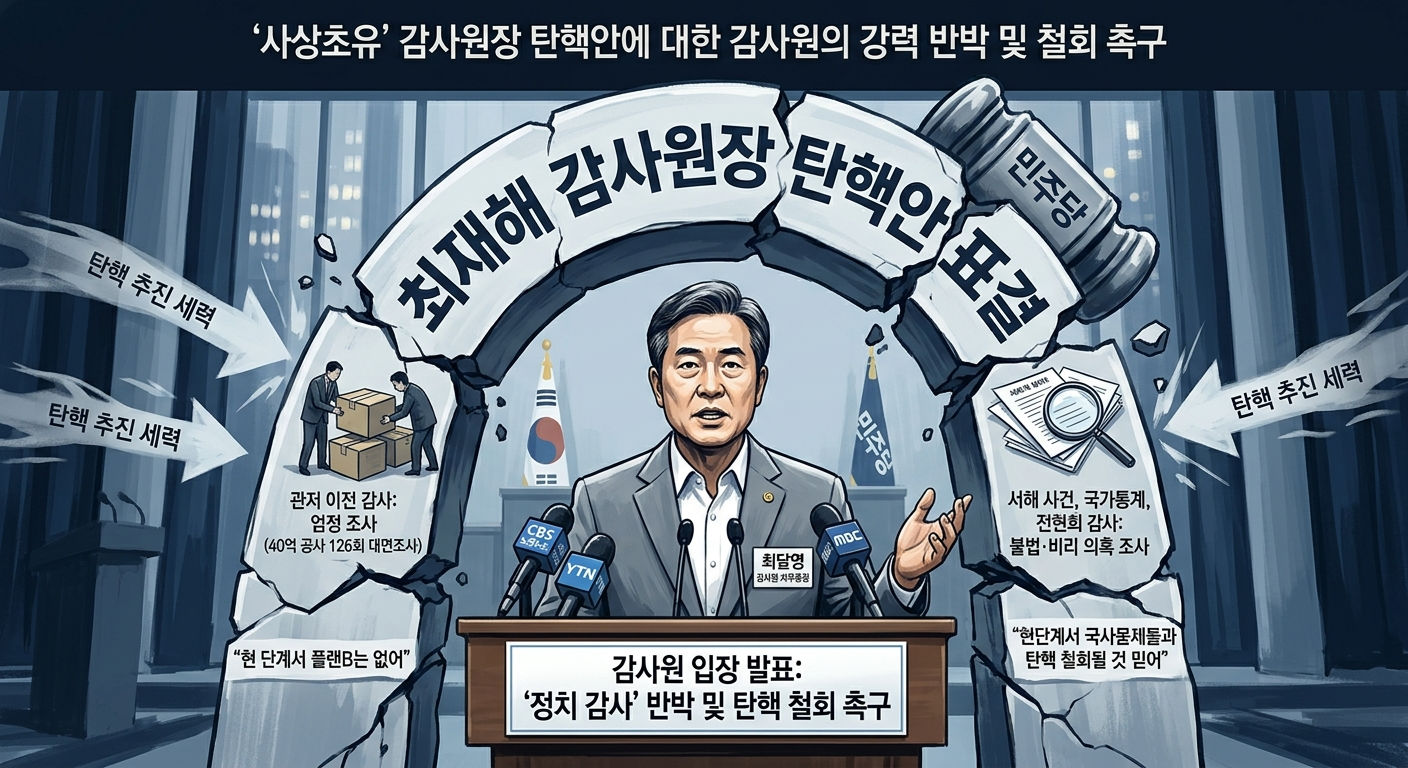} &
\centering \includegraphics[width=0.18\textwidth]{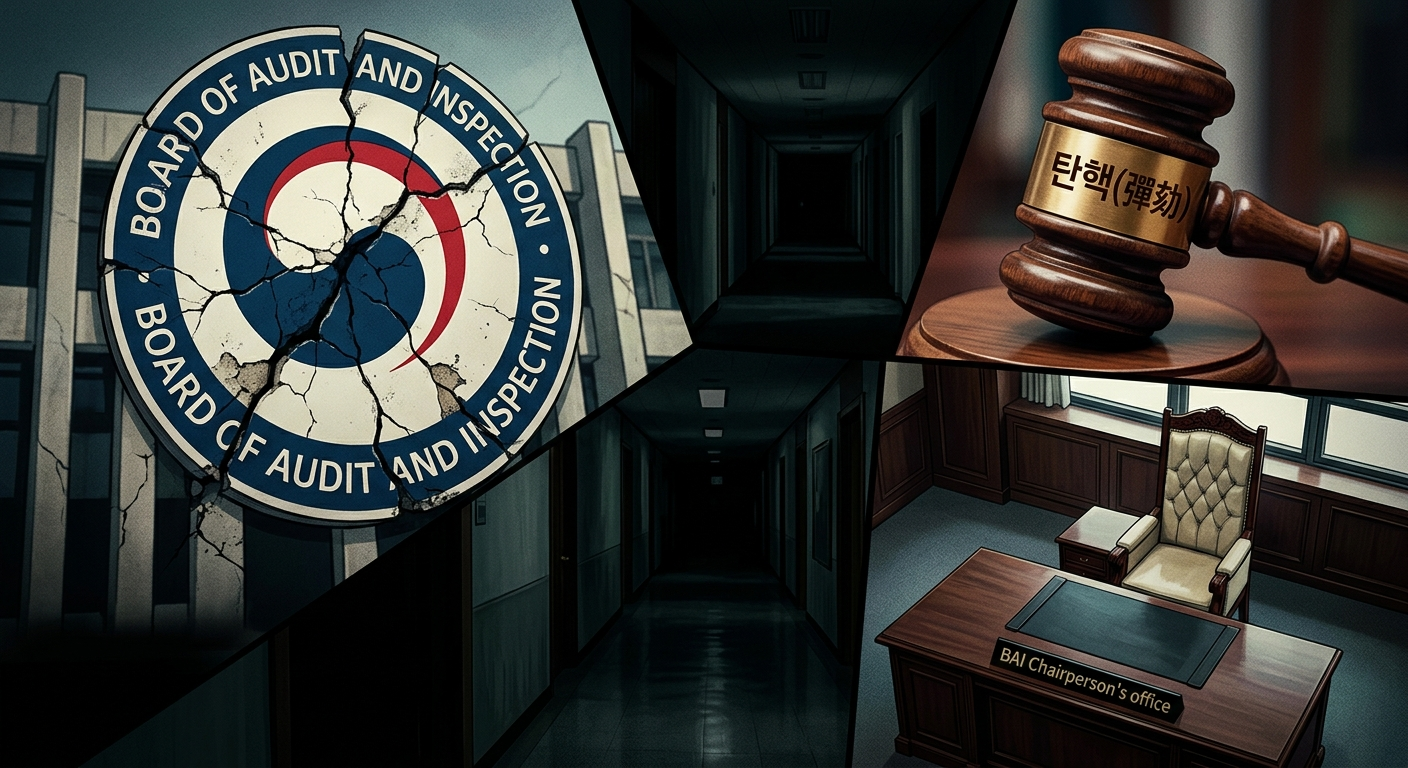} \tabularnewline
\midrule
C1 & Sup. &
\centering \small{[See: \url{https://imgnews.pstatic.net/image/020/2024/08/11/0003581252_001_20240811204909147.jpg?type=w860}]} &
\centering \includegraphics[width=0.18\textwidth]{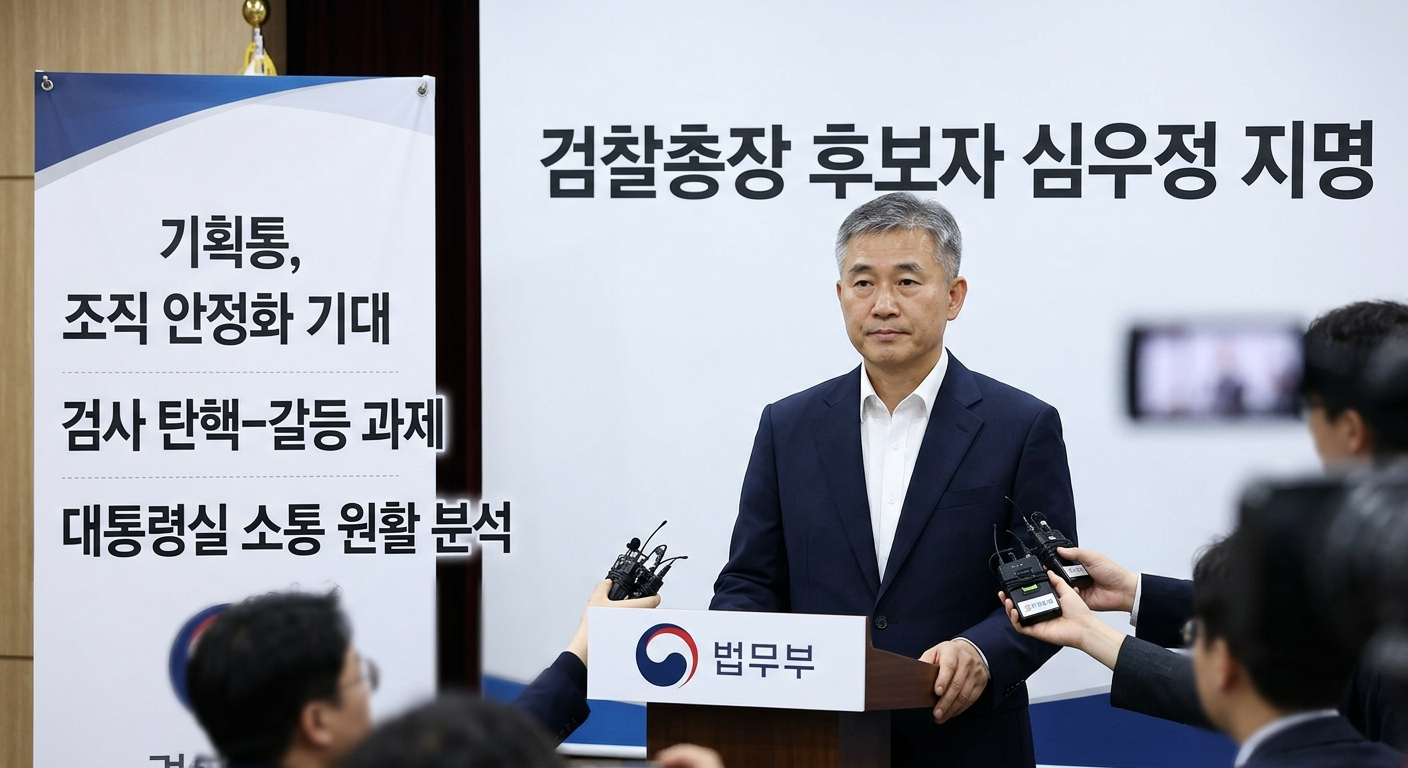} &
\centering \includegraphics[width=0.18\textwidth]{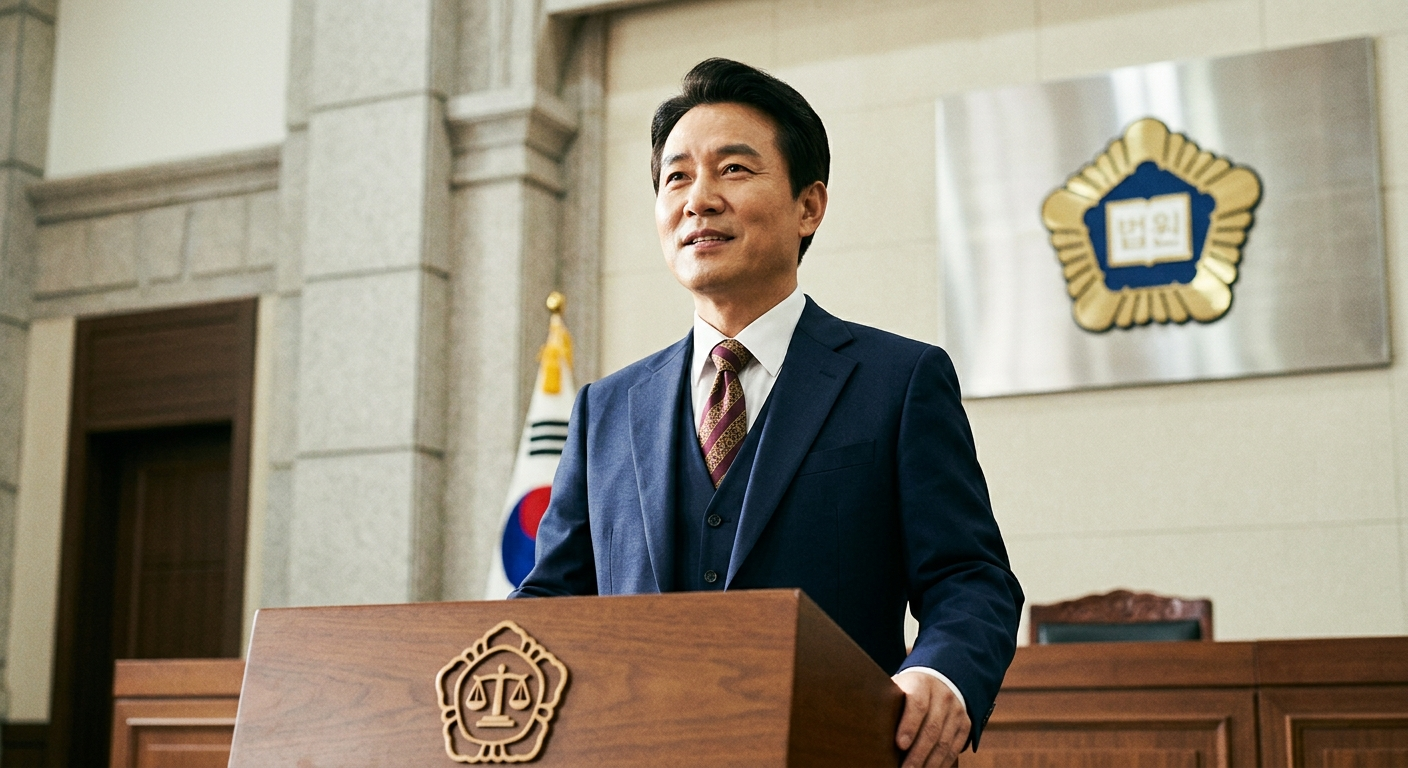} \tabularnewline
C2 & Neu. &
\centering \small{[See: \url{https://imgnews.pstatic.net/image/002/2024/08/11/0002345370_001_20240811180508603.jpg?type=w860}]} &
\centering \includegraphics[width=0.18\textwidth]{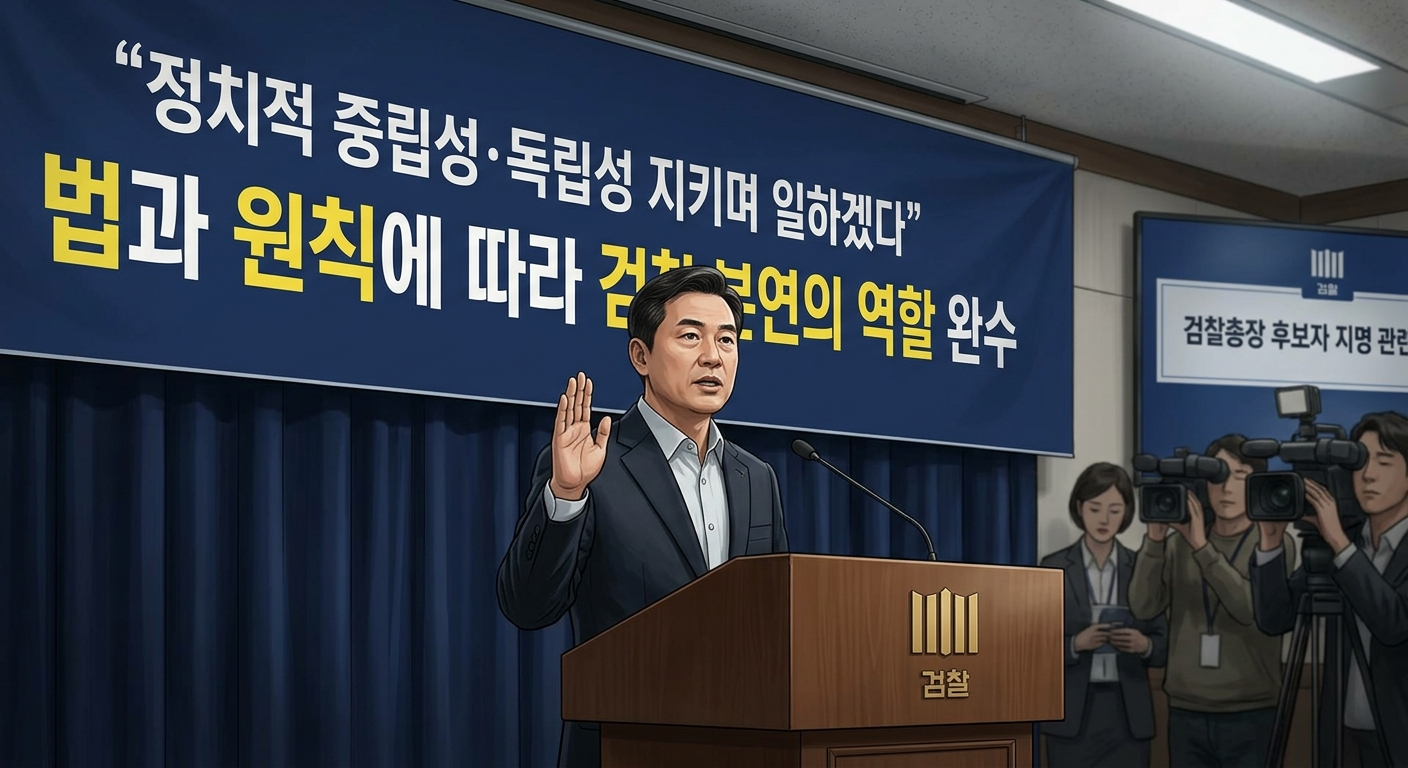} &
\centering \includegraphics[width=0.18\textwidth]{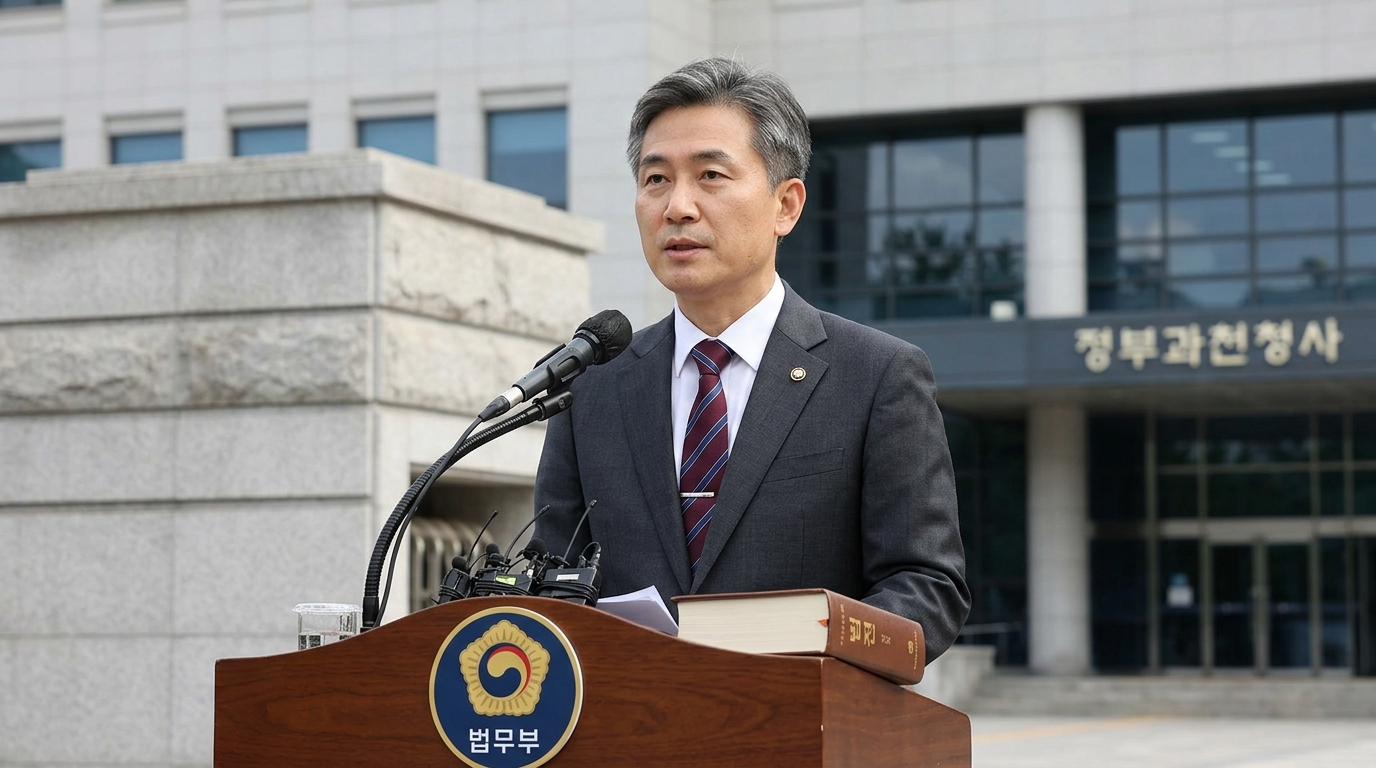} \tabularnewline
C3 & Opp. &
\centering \small{[See: \url{https://imgnews.pstatic.net/image/032/2024/08/11/0003314304_001_20250514095121338.jpg?type=w860}]} &
\centering \includegraphics[width=0.18\textwidth]{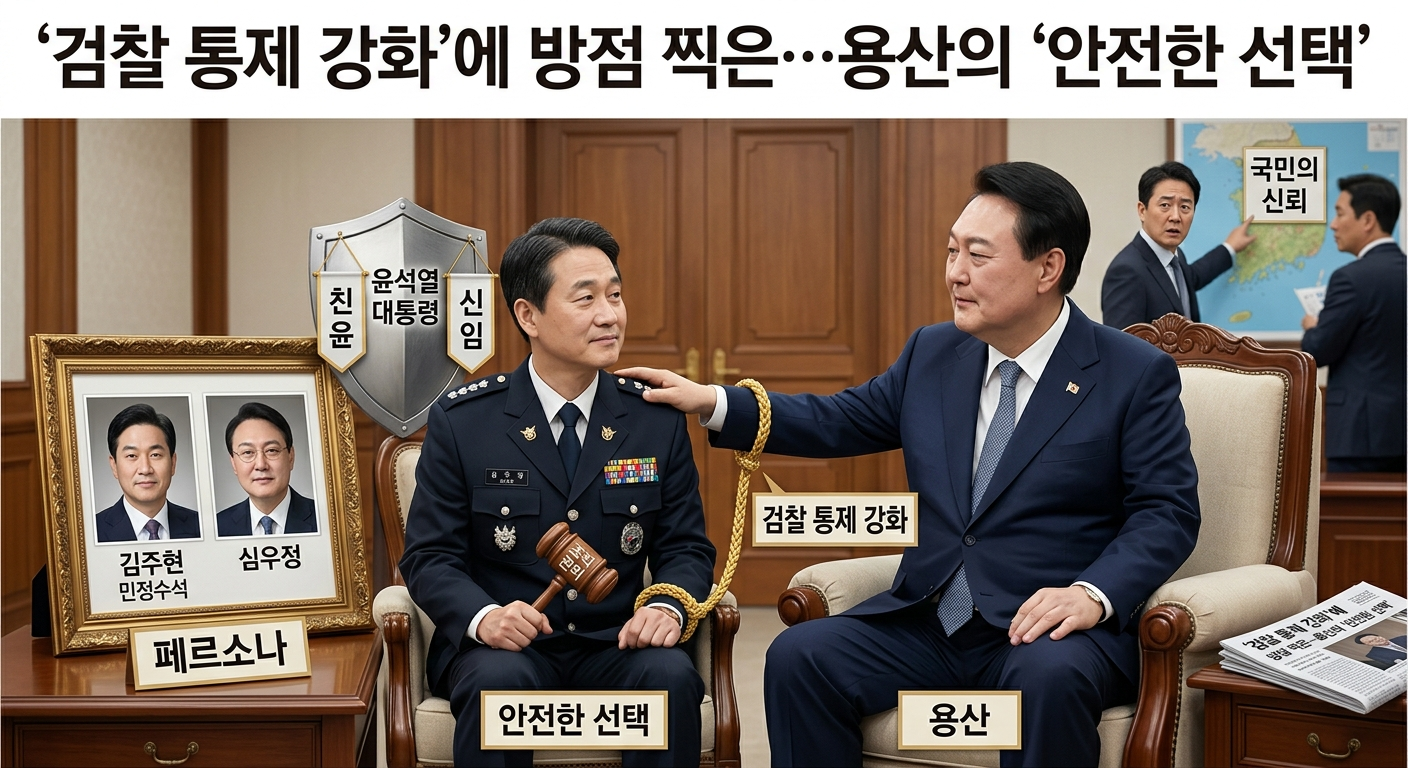} &
\centering \includegraphics[width=0.18\textwidth]{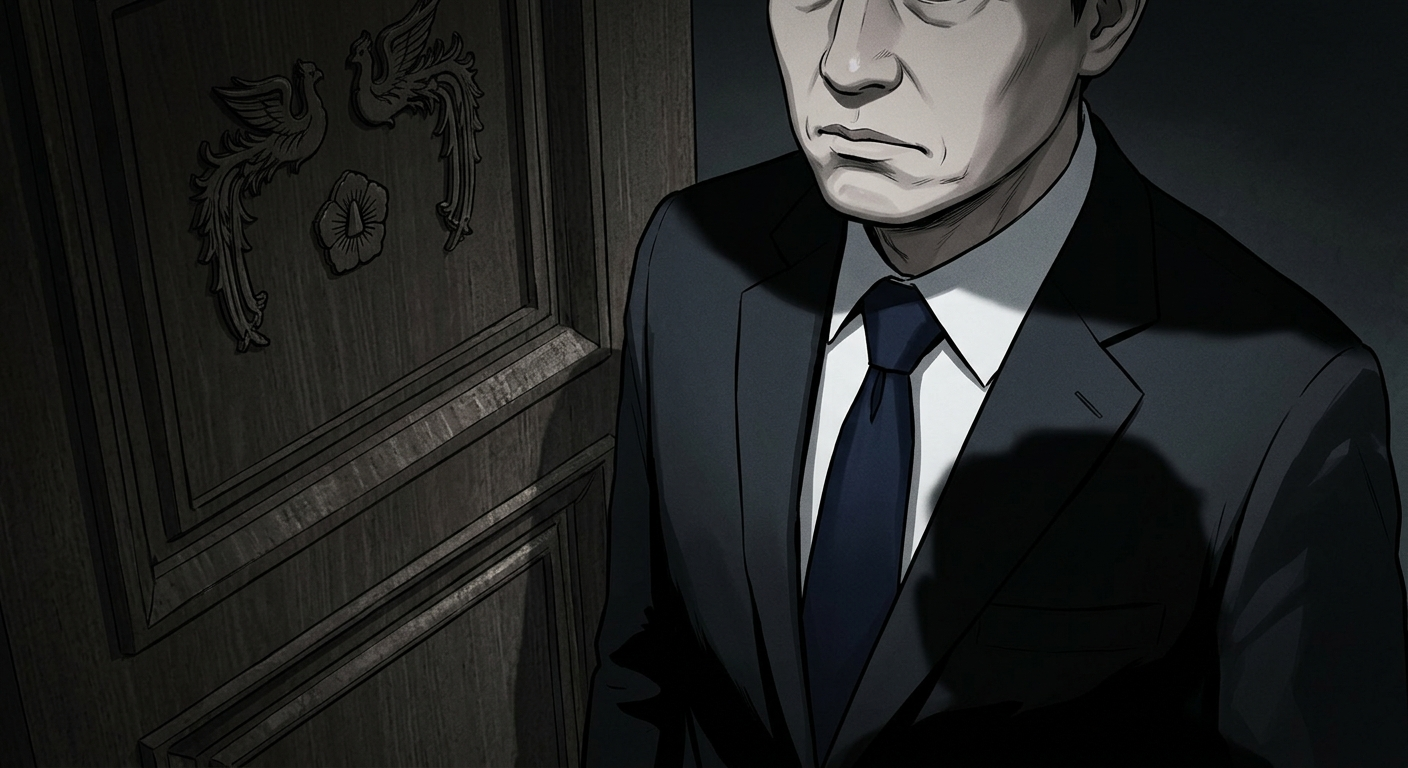} \tabularnewline
\bottomrule
\end{tabular}
\caption{Images for the nine articles used in the user study (Table~\ref{tab:case_study_article_sets_en}). Original images are omitted to respect copyright; their URLs are provided instead.}
\label{tab:case_study_images}
\end{table*}

\begin{figure*}[t]
\centering
\begin{subfigure}[t]{0.49\textwidth}
  \centering
  \includegraphics[width=\textwidth,height=0.88\textheight,keepaspectratio]{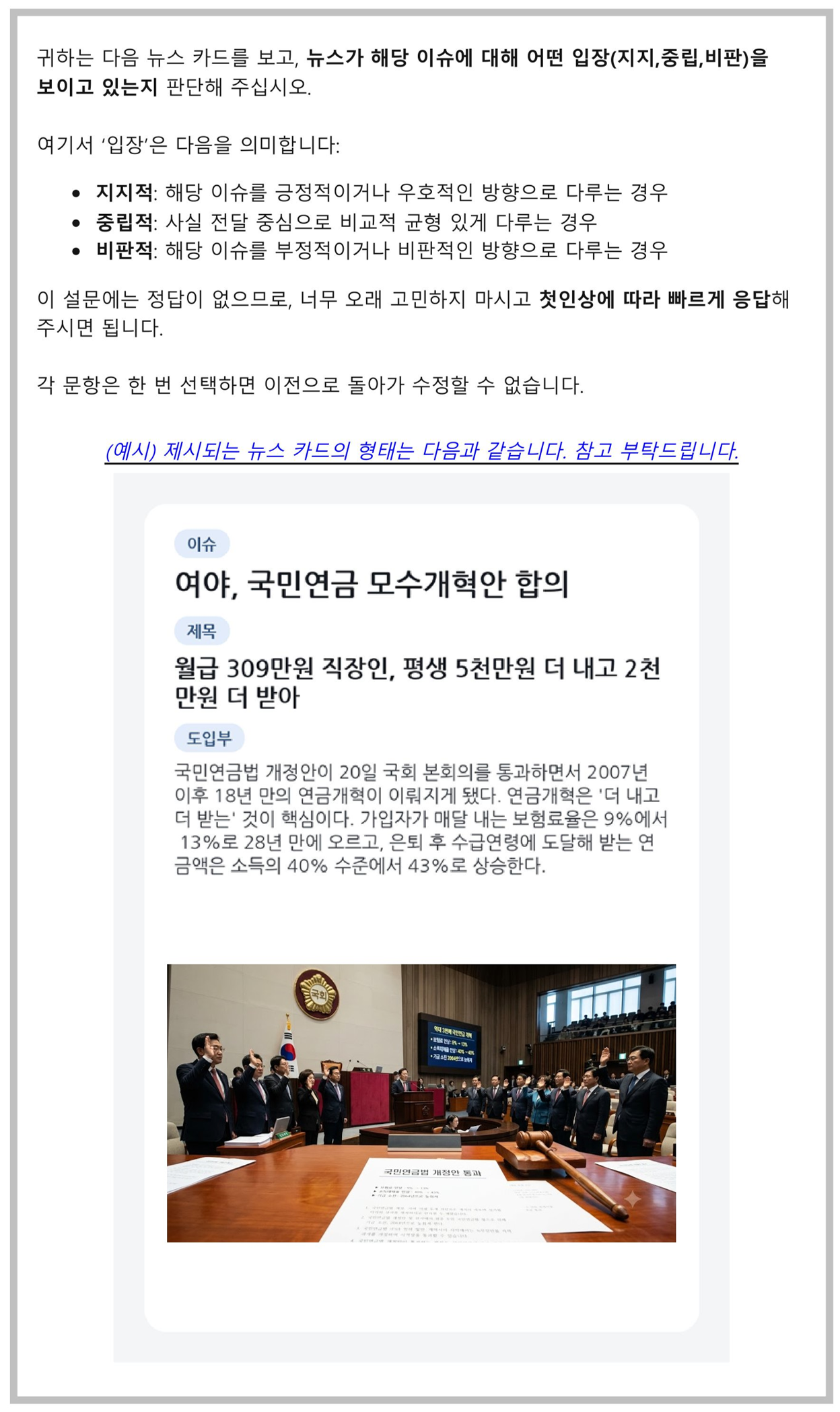}
  \caption{Instructions}
  \label{fig:case_study_instruction}
\end{subfigure}\hfill
\begin{subfigure}[t]{0.49\textwidth}
  \centering
  \includegraphics[width=\textwidth,height=0.88\textheight,keepaspectratio]{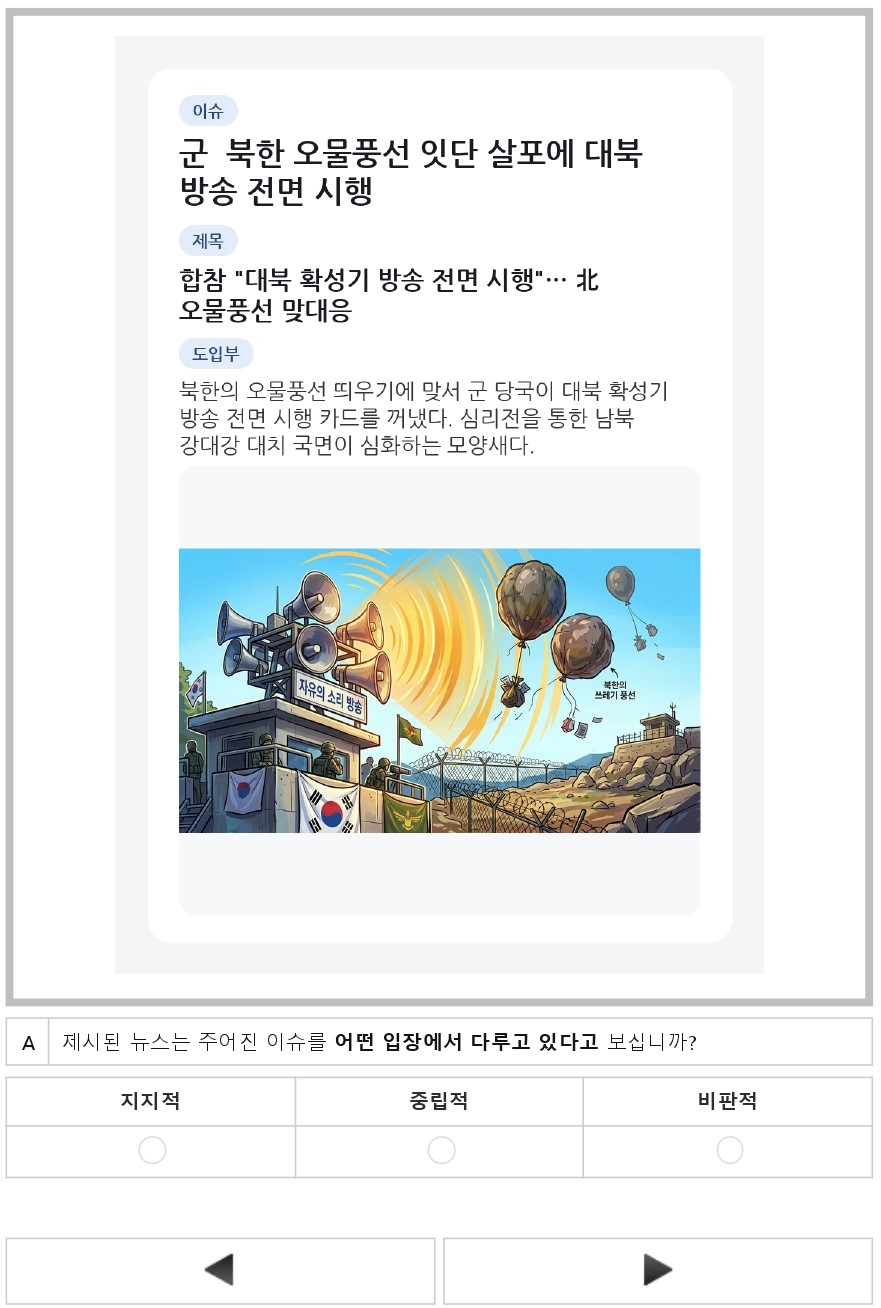}
  \caption{User interface}
  \label{fig:case_study_interface}
\end{subfigure}
\caption{Instructions and user interface used in the user study.}
\label{fig:case_study_materials}
\end{figure*}

\end{document}